\documentclass{aplreport}

\usepackage{trace}
\usepackage[utf8]{inputenc}
\usepackage{parskip}
\usepackage{amsmath,amssymb}
\usepackage{hyperref}
\usepackage{xurl}
\usepackage{soul}
\usepackage[paper=letterpaper,margin=1in]{geometry}
\usepackage{graphicx}
\usepackage{caption}
\usepackage{subcaption}
\usepackage{comment}
\usepackage[shortlabels]{enumitem}
\usepackage{pdfpages}

\newcommand{\smpl}{\mathbf{d}}
\newcommand{\smplideal}{\mathbf{\tilde{d}}}
\newcommand{\smplfeats}{\mathbf{d}^x}
\newcommand{\smpllbls}{\mathbf{d}^y}
\newcommand{\smplposs}{\mathbf{\tilde{d}}^p}
\newcommand{\dataset}{\mathcal{D}}
\newcommand{\datasetideal}{\tilde{\mathcal{D}}}
\newcommand{\singlefeat}{d^x_j}

\hypersetup{
    colorlinks=true,
}

\begin{document}

\title{CaTE Data Curation for Trustworthy AI}
\author{Mary Versa Clemens-Sewall \\
Christopher Cervantes \\
Emma Rafkin \\
J. Neil Otte, Ph.D. \\
Tom Magelinski, Ph.D. \\
Libby Lewis \\
Michelle Liu \\
Dana Udwin, Ph.D. \\
Monique Kirkman-Bey}
\maketitle


\hypersetup{linkcolor=black}
\chapter*{Executive Summary}
This report provides practical guidance to teams designing or developing AI-enabled systems for how to promote trustworthiness during the data curation phase of development. In this report, the authors first define data, the data curation phase, and trustworthiness. We then describe a series of steps that the development team, especially data scientists, can take to build a trustworthy AI-enabled system. We enumerate the sequence of core steps and trace parallel paths where alternatives exist. The descriptions of these steps include strengths, weaknesses, preconditions, outcomes, and relevant open-source software tool implementations. In total, this report is a synthesis of data curation tools and approaches from relevant academic literature, and our goal is to equip readers with a diverse yet coherent set of practices for improving AI trustworthiness.

Data curation is the phase in the AI development lifecycle after data have been acquired for development but before a machine learning model has been trained on that data. Typical actions in the data curation phase include exploring the data, manipulating it to prepare for training, and dividing (splitting) the data into portions that can be used to train and validate the model. 

These actions involve decisions about the meaning of the data and its fitness as training data for the AI-enabled system. In this report, we explain how to make these decisions in a way that promotes trustworthiness. We contend that developers can contribute to the ``trustworthiness" of an AI-enabled system by building the system to do its task as well as possible. For the data curation phase, when data is the primary focus, we formalize this notion into an \emph{actionable definition of trustworthiness} (Section~\ref{intro/scope}):
\begin{quote}
    \textit{A trustworthy AI-enabled system must be optimized for performance on the true distribution of inputs it will encounter in a deployed environment.}
\end{quote}

\begin{figure}[ht]
  \centering
  \includegraphics[width=\linewidth]{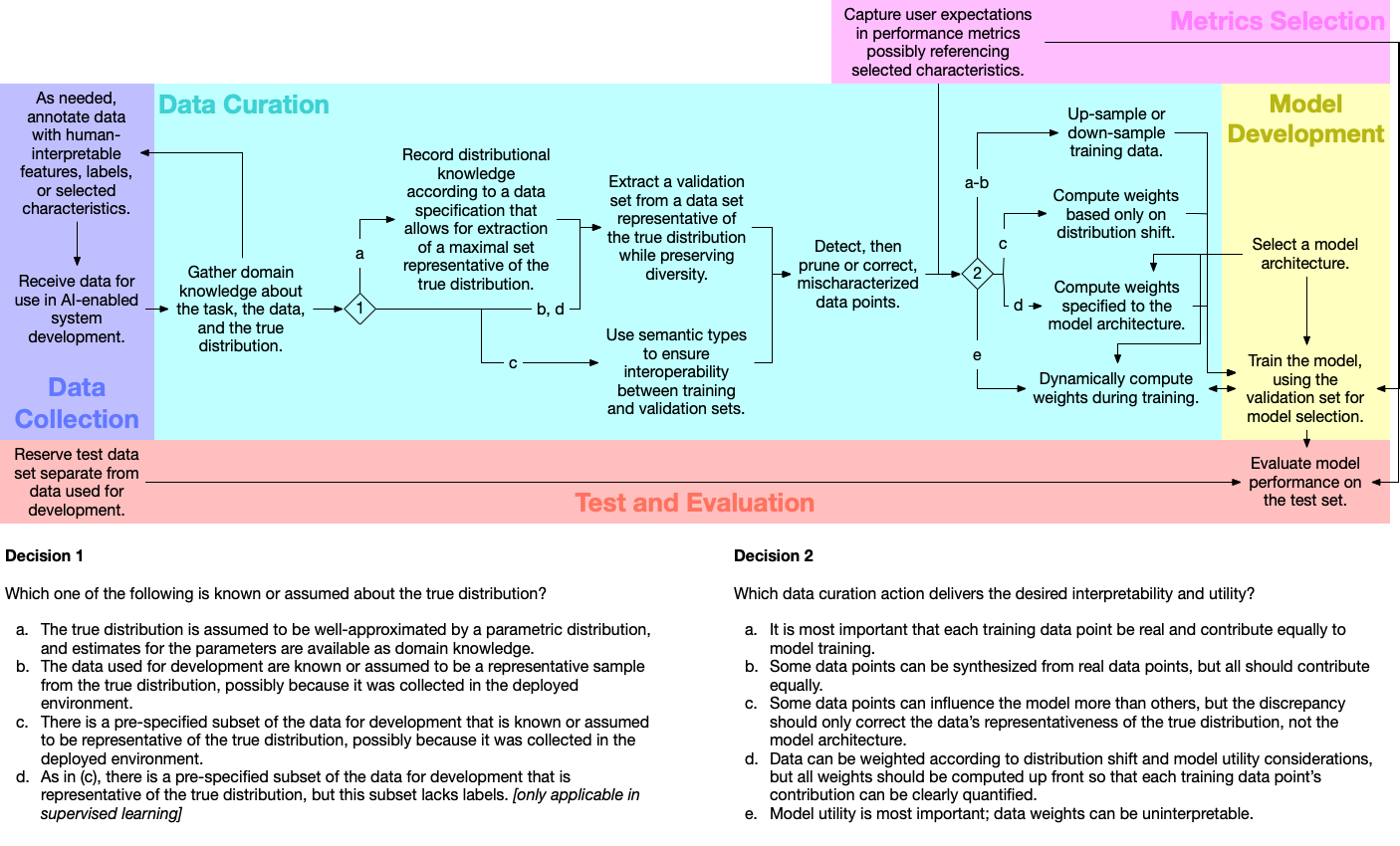}
  \caption{Data curation workflow composed of actions in the data curation phase and neighboring phases, dependencies between actions, and decision points for parallel paths.}
  \label{figs/curation_in_context}
\end{figure}

We organize the actions of the data curation phase to enable trustworthiness into a sequence of steps that are described and illustrated in Figure~\ref{figs/curation_in_context}. Chapters~\ref{knowledge}, \ref{sec:splitting}, \ref{sec:resample}, \ref{sec:reweight}, \ref{sec:semantic_types}, and \ref{mischar} are each devoted to one or more data curation actions. 

In the AI development lifecycle (Figure~\ref{figs/curation_in_context}), the data curation phase neighbors data collection, metrics selection, model development, and test and evaluation, where dependencies between steps and phases are indicated with arrows. We identified two major data curation decision points focused on (i) what is known or assumed about the true distribution and (ii) what actions best reflect the priorities of interpretability and utility., These decision points are summarized in Figure~\ref{figs/curation_in_context} and explained in Chapter~\ref{statistical_tools}. A decision tree in that chapter (Figure~\ref{figs/decision_tree}) guides project managers and data scientists through appropriate data curation actions based on Decisions 1 and 2.

In order to make informed actions and decisions, data scientists need domain knowledge about the task and the data. To facilitate the elicitation of domain knowledge, we introduce a set of 12 questions in Chapter~\ref{knowledge} to help guide conversations about domain knowledge, data characteristics, and the desired outcome of the data curation actions. These pieces of information are necessary for ensuring and assessing the trustworthiness of the AI-enabled system, and we trace the accumulation of this information across the data curation phase in Chapter~\ref{sec:information_system} and propose building a system to hold the information and enable an assessment of trustworthiness.

Chapter~\ref{sec:pretrained} addresses the topic of pretrained models, including many large language models, that are increasingly a part of AI-enabled systems. Rather than training all machine learning models from scratch, many development teams combine pretrained general-purpose models with additional modules they train from task-specific data. The chapter surveys emerging techniques that such developers can use to promote trustworthiness, especially when a pretrained model is used in the data curation phase for deriving new features of data points from those originally present in the dataset.

In addition to synthesizing relevant literature, we also experimented with open-source data curation tools for trustworthy AI. Our findings are described in the ``practitioner's perspective'' sections at the ends of Chapters~\ref{statistical_tools}, \ref{sec:splitting}, \ref{sec:resample}, \ref{sec:reweight}, \ref{mischar}, and \ref{sec:pretrained}. Our experiments were organized into two use cases with national security relevance: 
\begin{itemize}
    \item A computer vision model to help triage battlefield casualties, whose training dataset is a task-irrelevant dataset of drone footage and whose validation dataset is a smaller, task-relevant dataset.
    \item A natural language processing model to detect the mentions of chemicals in recent academic literature, whose training dataset is from many years ago and is weighted to better mimic today's journal articles.
\end{itemize}
Code used to perform the experiments is provided in the appendices.

Taken altogether, the contents of the report answer the question, ``what can be done in the data curation phase to promote AI trustworthiness?'' The data curation phase, however, is one part of a larger AI development lifecycle, and analogous questions could be asked about other phases in future work, including metrics selection, model training, and evaluation. We conclude this report with a discussion of those potential research directions.

\newpage

\tableofcontents

\chapter{Introduction}

\label{intro}

Data curation is the process of preparing data to derive the parameters of an artificial intelligence (AI)-enabled system by 
consuming one dataset and creating another. 

Parameters are commonly learned from data through machine learning (ML): an ML model architecture is selected, and curated data is used to train and validate the ML model. Data curation comprises a significant phase in the AI product lifecycle, as data scientists often spend less time conducting analyses than preparing data \cite{muller2019data}. We contend that data curation is not just a preparatory step for building an AI-enabled system but also an opportunity to promote trustworthiness in the eventual system. 

\section{Who this report is for and when to use it}
\label{intro/scope}

This report is for people building, designing, or envisioning systems enabled by AI, and it focuses on the data curation phase of the AI product lifecycle. 

The data curation phase is an early and requisite phase in the development of any AI-enabled system that incorporates an ML model. Data curation comes after all data used in developing the AI-enabled system has been obtained and, if necessary, systematically annotated. Data curation concludes when the AI-enabled system's ML model is fully trained. For our purposes, data curation may include preliminary training and evaluation of initial models because preliminary results may inform further curation decisions.

During curation, data scientists consume data that is available for AI development (which we will simply refer to as “data") into training and validation datasets. This step of the AI product lifecycle is crucial because learning in the ML context refers to the process of ``improv[ing]... performance on future tasks after making observations about the world" or, more formally, ``from a collection of input-output pairs,\footnote{Though learning can be understood in the context of explicit input-output pairs as in the supervised learning context, this definition can be conceptually broadened (e.g., given inputs and reinforcement signal).} learn[ing] a function that predicts the output for new inputs" \cite{russell2016artificial}. In this framing, the dataset on which an ML model is trained -- the observations about the world -- are the basis on which future predictions will be made. Careful curation, therefore, is essential to ensure ML model performance.

The purpose of this report is to describe and catalog techniques for creating training and validation datasets in a way that can increase the trustworthiness of the AI-enabled system once it is ultimately deployed. We assume that our reader intends to deploy the system once it is built and intends it to be trustworthy.

While trust is complex and trustworthiness multifaceted, we created and focus on the following actionable definition:

\textbf{Actionable Definition of Trustworthiness:} \emph{A trustworthy AI-enabled system must be \underline{optimized} for \underline{performance} on the \underline{true distribution of inputs} it will encounter in a \underline{deployed environment}}

This definition does not encompass all aspects of trustworthiness, but we contend that it is a necessary ingredient for trustworthiness and is actionable during the data curation phase of the AI product lifecycle.

\section{How prioritizing trustworthiness in data curation helps build trusted systems}
\label{intro/trust}

The actionable definition of trustworthiness aims to guide data scientists in the data curation phase, when the AI-enabled system is not yet built, toward design choices that contribute to the likelihood that end users will trust the eventual system, once built. The process of gaining users' trust is complex, and many factors are out of data scientists' control. Our actionable definition ties data science work into the overall process of building trustworthiness of the model in the deployed environment. We next examine the connection between the definition and the Integrated Model of Trust, introduced by Kelton et al. \cite{kelton}. This model of trust, which builds on earlier work from Mayer et al. \cite{mayer_1995_integrative}, highlights components that are commonly cited throughout the literature on when and why people trust. The Integrated Model of Trust enumerates four components of trustworthiness: competence, positive intentions, ethics, and predictability. 

In the view of Kelton et al., the trustee is competent if they “possesses the knowledge, expertise, or skill necessary to fulfill the needs of the trustor.” In the setting of AI, the trustee is the AI-enabled system, and the trustor is the user in the deployed environment. Relative the actionable definition, we consider optimization to relate closely to trustee competence. The model architecture may not yet have been selected as of the data curation phase, and the eventual model has not yet been fully trained, so data scientists cannot yet guarantee that the eventual model will have enough “skill” (i.e., according to some performance metric) to meet the end user’s need. However, an optimized model should have the best “skill” of any model of the same architecture.

Kelton et al. define positive intentions as “the trustee’s feelings toward the trustor.” Realistically, data scientists and model developers have very little influence over the end users' feelings about AI-enabled systems. User perception is complex, including model performance, the way results are presented, the overall design of the system, and the experiences users bring to their interactions. If developers, data scientists, and end users collaborate during system deployment, it may be possible for end users to establish trust in their collaboration and transfer that trust to the AI-enabled system. In such contexts, it would be important for developers and data scientists to communicate how and in what contexts they considered the ``deployed environment" in developing the system to meet user needs.

Third, Kelton et al. list ethics, to which they attribute the qualities of fairness and honesty relating to trust. While not explicitly part of our definition of trustworthiness, these are both integral to the methods we catalog in the following sections. We push for an honest portrayal of the model's likely performance in the deployed environment by highlighting the need for -- and providing methods for (imperfectly) deriving -- a validation dataset that represents the “true distribution of inputs” as much as possible. Our best effort for honesty, as developers, is to report model performance on such a validation set, as well as held-out processed and unprocessed testing sets. As for fairness, many of the techniques compiled in this report were designed to achieve fairness across different model inputs. “Performance” in the actionable definition can be one of the fairness metrics developed and used in the literature. We hope that fairness to different inputs is perceived by users as evidence of the model’s fairness and, hence, demonstrates a clear consideration of ethical development.

For Kelton et al., predictable trustees are those whose behavior conforms to expectations or, synonymously, reliable trustees. The extent to which the model is predictable depends on the model architecture and whether there is enough data to train it well. Nevertheless, data curation assists by trying to retain as much of the data as is likely to be helpful for model training. Where justified, data curation involves synthesizing new data similar to existing data. By placing emphasis on the “true distribution of inputs,” data scientists imply that they want the model to perform best on the sorts of inputs that the AI-enabled system is likely to encounter once deployed.

\section{What performance metrics capture}
\label{sec:performance}

The measurement of an AI-enabled system's performance -- what good looks like -- is an essential component of enabling the system's trustworthiness. Performance, in the context of the actionable definition of trustworthiness, is only meaningful relative to the deployed environment; a model may be performant on some benchmark data, but is only trustworthy when measured against the true distribution of inputs in the deployed environment and the larger context in which the system will be used. 

Data scientists must gain specific insights about the deployed environment to (a) understand the true distribution (so that performance can be optimized) and (b) appropriately define performance. To do so, data scientists need both of the following.
\begin{enumerate}
    \item Knowledge of the true distribution by way of either approximate distributional parameters or a representative sample from the true distribution, which likely originate from the deployed environment
    \item Knowledge of the expectations that the deployed system's users will have about the AI-enabled system, encapsulated in performance metrics that can be used to judge how well a candidate model meets user expectations (i.e., model validation)
\end{enumerate}

Choosing the appropriate methods for assessing system performance, or metric selection, is a nuanced and challenging task that requires knowledge of the task and the deployed environment. For example, one common user expectation is that an AI-enabled system will perform similarly across the true distribution. That is, the system will be no more correct on certain kinds of inputs than others. By contrast, one common performance metric is accuracy over a validation, or held-out, dataset: the proportion of predictions that match the annotations (i.e., how often was the model correct). Accuracy does not provide information on how the model’s performance varies across the validation set and, by extension, across the true distribution. Even if the validation set is carefully chosen or constructed to be representative of the deployed environment, accuracy fails to convey variation in performance across the distribution. In this scenario, a different performance metric is needed to capture variable model performance.

When variable performance is a concern, one option is to measure performance on smaller regions in the space of possible inputs. Measurements on individual regions are combined to form a single performance metric -- for instance, by taking the average across all regions or the worst of all regions’ accuracies. Recently, many such techniques have been developed in the field of group fairness for ML. As an added benefit, many of these techniques do not assume that the deployed AI-enabled system can determine which region a given input is from. We will refer to the regions of the space of possible inputs as values of a selected characteristic. For each data point, the value of the selected characteristic specifies the region in which the data point is located. Fairness literature often focuses on applications in which the selected characteristic is a protected or sensitive attribute of each data point, such as race or gender for data points representing humans. The techniques are applicable broadly, though, to any selected characteristic across which reliable performance is key.

While performance metric selection is a critical task for fostering trustworthy AI-enabled systems, it is beyond the scope of this report. Metric selection is not directly related to the data curation phase (i.e., it is taken as a given that a system will be evaluated against some defined criteria separate from curation activities). For data scientists seeking to evaluate AI-enabled systems, many performance metrics are collected, described, and implemented in scikit-learn \cite{sklearn-metrics}, and fairness metrics specifically are implemented in AI Fairness 360 \cite{aif360-oct-2018}.\footnote{AI Fairness 360: \url{https://github.com/Trusted-AI/AIF360}} 

\section{Why knowledge of the true distribution of inputs in the deployed environment is needed}
\label{intro/truedistro}

The actionable definition of trustworthiness assumes the existence of a true distribution of inputs in the deployed environment: a conceptual grouping of all inputs the AI-enabled system may encounter, framed in the language of statistics (i.e., a data point in the deployed environment is drawn from the true distribution). It is not assumed that the data available for development are a representative sample from the true distribution, nor is it necessarily the case that the true distribution can be fully known and parameterized. In many cases, data are repurposed or were originally created by convenience sampling. While sometimes care and effort can mitigate discrepancies between the development data and true distribution, this is not always the case. For example, a forecasting model (predicting future events) is necessarily trained on historical data, and there could be substantive distributional differences between data from the past, the present, and the future.

We contend that the demonstrable trustworthiness of an AI-enabled system is constrained by the extent to which the true distribution is known. Most data curation techniques expect this knowledge to be in the form of a representative sample of data from the true distribution, which we refer to as the validation dataset. The validation set could have been extracted from the larger pool of data available for development, or it could have been created independently from the rest of the data by sampling directly from the deployed environment. Since data curation comes after the data has been obtained, we assume that the contents of the validation set have been obtained but, possibly, have not been extracted from the data as a whole. 
In Section~\ref{knowledge/common}, we offer guidance for eliciting subject matter expertise that can be used in extracting the validation set from a larger dataset that is not representative of the true distribution yet contains enough breadth to encompass common inputs in the deployed environment. Chapter~\ref{sec:splitting} offers techniques for splitting the representative subset while maintaining diversity in a small validation set, and Chapter~\ref{sec:semantic_types} describes semantic types, which can help when using the elicited subject matter expertise to query the data for a representative subset of the true distribution. 

The degree to which trustworthiness (by our actionable definition) can be measured, demonstrated, and achieved is limited by the extent to which the validation set is representative of the true distribution. There may be many reasons that knowledge of the true distribution is limited, including for military applications or information security. The performance of an AI-enabled system, however, can only be assured and optimized relative to how much is known about the true distribution. 

The U.S. Department of Defense has adopted ethical principles for AI \cite{dodaiethicalprinciples}, including the “reliable” principle, with which our actionable definition fits well: “the Department's Al capabilities will have explicit, well-defined uses, and the safety, security, and effectiveness of such capabilities will be subject to testing and assurance within those defined uses across their entire life-cycles.” When defining the use of an AI-enabled system, it is most useful to provide the development team with a quantitative description (either in approximate distributional parameters or in a sufficiently numerous set of examples) of the true distribution of inputs in the deployed environment. Without such knowledge of the true distribution, it is not possible to demonstrate that an AI-enabled system is trustworthy. 

In this report, we assume the reader intends for the AI-enabled system being designed or implemented to be trustworthy and reliable, and one limiting factor in that pursuit is the extent to which the true distribution is known and, ultimately, represented in validation data. It might be that data scientists are provided with a validation set that is representative of the true distribution. We describe, in this report, approaches to construct such a validation set when one is not provided.

\section{How to navigate this report}

This report details the data curation phase, from gathering necessary information about the data to choosing among several methods that can account for the expected discrepancy between the data available for model development and the environment into which the AI-enabled system is intended to be deployed. 

Part I focuses on understanding the task, the data, and the deployed environment relative to the data curation phase. Chapter~\ref{data} provides a theoretical framing of data and the entities or phenomena represented thereby. Chapter~\ref{knowledge} equips the data scientist with a set of questions and guidance for collecting crucial information early in the data curation phase. Some of this information may be found in project and data documentation, and the rest comes through bidirectional communication with subject matter experts (SMEs). Unless the whole dataset or a pre-specified subset is known to be representative of the true distribution, one piece of information elicited is a quantitative description of the true distribution. 

Building on this understanding, Part II focuses on actions the data scientist can use to consume development data and create training and validation datasets. One scaffolding to help the data scientist is a decision tree (Chapter~\ref{statistical_tools}), which provides guidance on when and under what conditions certain techniques can be used. Chapter~\ref{sec:splitting} catalogs splitting techniques that aim to choose a validation set that is as representative as possible of what is known about the true distribution. We expect that developers will construct a training set from the remainder of the data, and the subsequent two chapters provide techniques for curating it. Chapter~\ref{sec:resample} details options for winnowing or synthesizing data given some criteria (e.g., down-sampling data points that are too common, up-sampling data points that are not common enough). Chapter~\ref{sec:reweight} describes the process of assigning numerical value on data points and/or their features transparently (e.g., each data point is assigned a weight to represent how important that data point is in approximating the true distribution) or opaquely (e.g., weights on inputs are learned as part of training). 

Part III discusses additional topics that span some of the data curation actions and decisions a data scientist may take in support of trustworthy AI. These include the use and importance of semantic types (Chapter~\ref{sec:semantic_types}) and the need for relevant information about the data and true distribution to be documented in an information system (Chapter~\ref{sec:information_system}). Chapter~\ref{sec:distributional_knowledge} details the challenges and opportunities for detecting and correcting for data that does not accurately reflect the entities or phenomena it purports to. Finally, Chapter~\ref{sec:pretrained} brings the conversation of data curation actions to the domain of pretrained models and resources, including generative large language models (LLMs), and details how the discussion of data and the true distribution remains crucial even when portions of the AI-enabled system are inaccessible to the data scientist.

Most chapters end with a ``practitioner's perspective'' section, in which we report on our firsthand experience using one or more techniques described in the chapter. In contrast to the rest of a chapter, a ``practitioner's perspective'' focuses on a single, open-source tool that we used in a specific way. The purpose of these sections is to provide data scientists and developers with examples of when and how to use open-source data curation tools. 

Appendix~\ref{sec:nlp_tutorial} contains a tutorial describing our use cases for the tools and including the code we used to run the software-based tools.

\part{Understanding Data and Its Context}

\chapter{Consider data as purposeful simplifications of the world}
\label{data}
In the data curation phase of the AI product lifecycle, data creation and annotation is already complete. A data scientist is thus assumed to have (a) access to data, (b) an intent to learn parameters for (training) an ML model, and (c) a desire to ensure the trustworthiness of the system by training the model on a curated version of the dataset. This section defines what is meant by data in this context, and how that definition relates to the work of data scientists generally and the data curation phase specifically. 

\section{Enumerate features and labels of data points}
\label{data/def}

ML is the ``creation of mechanisms that can look at examples and produce generalizations" \cite{goldberg2022neural}. In order to do this, the real world must be encoded in discrete, machine-interpretable values. When data are prepared for machine learning contexts, entities or phenomena of interest are represented as sequences of \emph{features}, or encoded aspects of the entity or phenomenon. A single data point is typically represented as a vector of feature values, where human-interpretable representations typically encode one feature per dimension (e.g., an object's length, the date of a record). More generally, these feature vectors are represented as real numbers ($\smplfeats \in \mathbb{R}^m$).

Neural network architectures typically require inputs to be \emph{embedded} as dense, low-dimensional (length-$m$) vectors. In this paradigm, a network learns statistical correlations between inputs and shares parameters between them, creating a learned representation of each input without value $d^x_i$ having any human-interpretable meaning \cite{goldberg2016primer}. The networks that learn these embeddings can be trained on a domain-specific task, a proxy task, or in an unsupervised manner where there is an expectation that the learned embeddings will be of general purpose. Word embeddings, for example, are typically produced by unsupervised approaches that derive from the distributional hypothesis: that words are similar if they appear in similar contexts \cite{harris54}. 

Traditionally, labeled data points consist of two parts: the features ($\smplfeats$) and label(s) ($d^y \in \mathbb{R}$).\footnote{In many classification problems, real-world data points are annotated $d^y \in \mathbb{N}$, where each class is represented by a natural number (e.g., fishing vessel = 0, passenger ship = 1, military ship = 2, etc.). However, we consider $d^y$ more broadly to encompass datasets curated for other purposes, like regression-based problems.} For example, an image of a ship may be represented as pixel values (features) and annotated with whether the ship is a military or civilian watercraft ($d^y \in \{0, 1\}$). While it is typical for a dataset to have a single label per data point, some data points have multiple labels such that the annotation is a vector ($\smpllbls \in \mathbb{R}^l$, where $l$ is the number of possible labels). 

Though there can be practical distinctions between features and labels (e.g., dataset creation and dataset annotation are performed in distinct phases), a label is simply an encoded aspect of a data point that a model is trained to predict from the other encoded aspects (features) \cite{Hossain2024}. We therefore conceptually combine features and labels into a single representation in which data point $\smpl \in \dataset$ describes all encoded aspects of an entity or phenomenon after dataset creation and annotation. 

Consider a dataset describing ships, such that each data point contains encoded information about a ship's appearance (e.g., pixel values in an image) as well as other aspects (sometimes called metadata) like its age, country of origin, and type. In framing $\smpl$ in this way, data points can be considered without reference to a specific AI-enabled system. For example, one system might seek to predict ship type (military or civilian) from pixel values alone. In this case, $\smplfeats$ would be the pixels, $d^y$ would be the ship type, and all other encoded aspects would be ignored. Another system may predict country of origin from all other data, and a third system might attempt to cluster similar images without regard to the other encoded aspects. All such cases could leverage the same $\smpl$, but the parts of the data point that are chosen to be features or labels vary by task.

In this paradigm, a data point is a real-valued vector the length of the total features and labels available after creation and annotation: $\smpl \in \mathbb{R}^{m + l}$. This generalized $\smpl$ contains all encoded aspects available to the AI-enabled system, and can therefore be understood as the entity from the system's perspective. Features and labels are both encoded aspects of the real-world entity or phenomenon, and are crucial to the AI-enabled system's ability to learn correlations between those aspects.

\section{Recognize limitations of data}
\label{data/simple}

Creating dataset $\dataset$ -- describing real-world entities and phenomena as data points -- is a human-centered process in which data are not collected from an environment but are purposefully created. Choices are made about what to include, what to exclude, and how to reduce the complexities of reality into machine-interpretable representations \cite{muller2019data}. In some contexts, an image may be a useful approximation of an entity. In others, it may be necessary to encode other aspects of an entity into its representation. There may also be tasks in which the data collection context itself may be necessary to encode (e.g., the camera that took the image, the time of day). Simplifying reality into data points -- encoding some aspects as features or labels -- inherently excludes other aspects. 

$\dataset$ is a manifestation of choices, and $\smpl$ therefore does not necessarily describe an entity or phenomenon relative to the goals of the AI-enabled system. Instead, $\smpl$ reflects the aspects of the real world that were chosen during the creation of $\dataset$ along with the choices of how those aspects would be encoded. In the case of human-interpretable features, these choices include what features were chosen and how they were represented. In the case of dense vectors, like word embeddings, these choices include the method and goals of the embedding system (e.g., while word similarity has proven useful in practice, “word similarity is hard to define and is usually very task-dependent" \cite{goldberg2016primer}). Thus, while $\smpl$ is the entity or phenomenon from the perspective of the AI-enabled system, it isn't necessarily the case that $\smpl$ is a good or even sufficient representation for the task. It is merely the data point that was created. 

We therefore consider an ideal data point in an ideal dataset ($\smplideal \in \datasetideal$). This ideal vector, $\smplideal \in \mathbb{R}^{m + l + p}$, contains all possible aspects of the entity or phenomenon of interest, encoded in all possible ways (constrained to real values), such that $p$ refers to the size of the possible feature space ($m << p \leq \infty$). While no dataset can contain such idealized data points, this representation highlights the choices that are made in the construction of $\dataset$. A real entity or phenomenon is only partly represented by $\smplfeats$ and $\smpllbls$; the rest of its representation appears in the idealized $\smplposs$.

This formalism is particularly helpful relative to trustworthiness and fairness in AI, as issues of relative importance, mischaracterization, or distribution shift often occur in the $\smplposs$ space. This may be because relevant aspects are missing (and only appear in $\smplposs$) or because aspects are encoded in $\smplfeats$ as proxy features or labels where a more direct encoding appears in $\smplposs$. 

When classifying images as military or civilian ships, a relevant aspect may be the presence of a mounted gun ($\tilde{d}^p_j \in \{0, 1\}$). In some images, the gun may be visible and thus encoded by proxy in the pixels. In others, the image may be taken from an angle that obscures the gun and thus does not represent $\tilde{d}^p_j$ even by proxy. ML models trained on such images, then, may yield unexpected classification results relative to an important but only partially encoded aspect of the entities of interest.

Data curation for AI trustworthiness requires understanding the true distribution of inputs in the deployed environment. This true distribution, however, is informed by the relationships between $\smplfeats$, $\smpllbls$, and $\smplposs$. The choices made in dataset creation and the potential differences between the environment in which the data was created and the deployed environment require not only an awareness of the distribution of features and labels (e.g., training data is 50/50 military and civilian ship images, but the deployed environment is 70/30) but also an understanding of how the real-world entities or phenomena represented by the deployed environment's data differ from those in the training data, and how those differences might influence input features and labels. 

\subsection{Example data characterization challenges}
\label{data/example}

\begin{figure}[ht]
\centering
\begin{subfigure}{.5\textwidth}
  \centering
  \includegraphics[width=.8\linewidth]{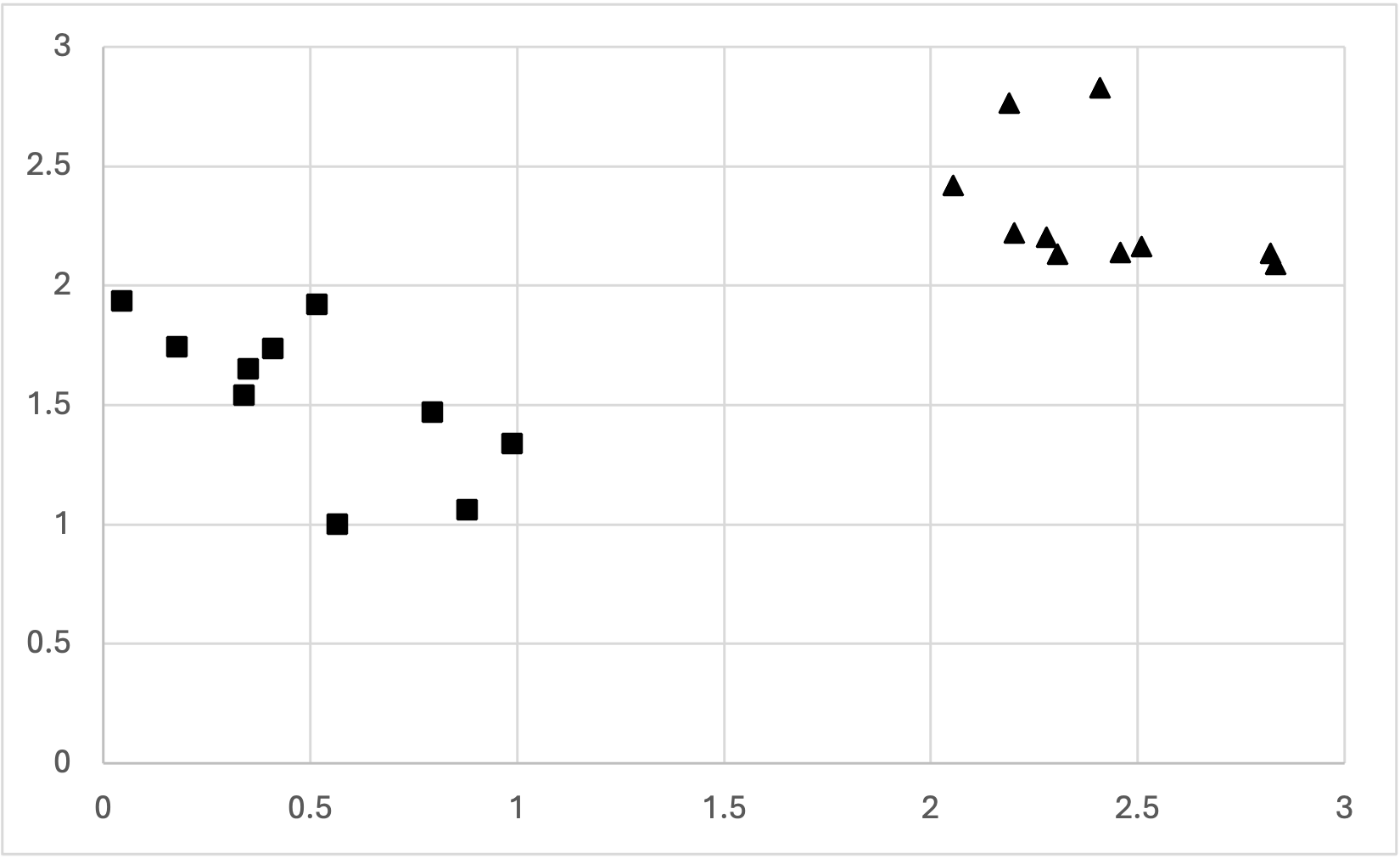}
  \caption{}
\end{subfigure}%
\begin{subfigure}{.5\textwidth}
  \centering
  \includegraphics[width=.8\linewidth]{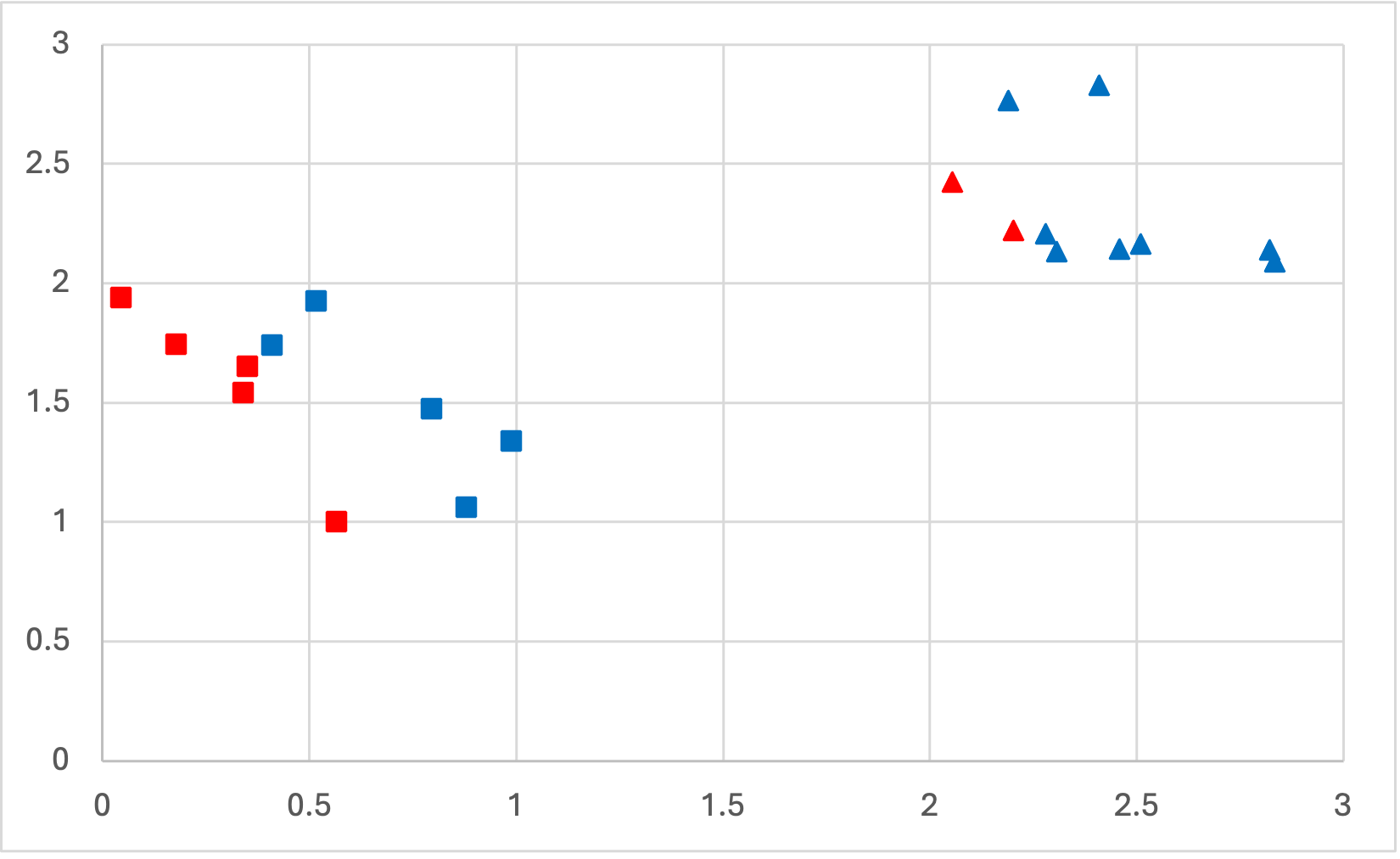}
  \caption{}
\end{subfigure}
\caption{Two-dimensional dataset ($\vert \smplfeats \vert = 2$) shown with labels (square, triangle); (a) depicts the available dataset ($\dataset$); (b) depicts the ideal dataset ($\datasetideal$) in which a relevant but un-encoded aspect is pictured (red, blue)}
\label{figs/2d-dataset}
\end{figure}

Consider a project in which a data scientist has been given a simple labeled dataset, shown in Figure~\ref{figs/2d-dataset} (a). Here, the dataset includes two real-valued features ($\smplfeats$ are coordinates on the plane) and one binary categorical label ($d^y \in \{\textit{square}, \textit{triangle}\}$). Given this dataset, it may be reasonable to consider that new data points close to $(0.5, 1.5)$ should be squares, and those close to $(2.5, 2.5)$ should be triangles. Some models trained on this dataset will learn the statistical relationship between these apparent centroids and the labels. If the inputs in the deployed environment are similar to those in $\dataset$, it may be reasonable to assume that performance in the deployed environment will meet expectations.

As discussed in Section~\ref{data/simple}, dataset $\dataset$ is a purposeful simplification: some aspects of interest were included, while others were excluded. Consider the idealized dataset $\datasetideal$, shown in Figure~\ref{figs/2d-dataset} (b), in which each data point has a color (a new feature to which the data scientist does not have direct access). If the red/blue distinction (or anything else encoded in $\smplposs$) is statistically independent of $\smplfeats$ and $\smpllbls$, they are considered irrelevant to the task. That is, if unrepresented aspects are not correlated with $\smpl$, these aspects can be ignored. With enough data, these unrepresented aspects will be uniformly distributed relative to the represented aspects of $\smpl$.

It is common, however, for aspects in $\smplposs$ to have an unknown dependency with $\smpl$ (e.g., $\singlefeat$ is a proxy for $\tilde{d}^p_k$). In the Figure~\ref{figs/2d-dataset} example, consider a SME or other stakeholder who informs the data scientist that there are unequal numbers of red and blue points in the deployed environment, and that the model should perform equally well on red and blue points (equal opportunity; see Section~\ref{data/fairness}). 

If the color of the data points is independent of other features, there is no need for special data curation relative to color: a model trained on $\dataset$ will perform equally well on red and blue points, as they can be considered independent and identically distributed. If, however, the distribution of red and blue points is correlated with the features or labels -- as in Figure~\ref{figs/2d-dataset} (b) -- data curation may be appropriate. 

Since data curation occurs after dataset creation and annotation, there is not an opportunity to add the red/blue distinction as a new feature. As a result, a system may not be as optimized for performance in the deployed environment as one with direct access to all relevant aspects. Trustworthiness for AI-enabled systems, however, is not a binary. Given the additional domain knowledge about data point color, data curation can still improve the trustworthiness of the system~--~its performance in the deployed environment~--~even without new features. 

In this scenario, the data scientist -- following Chapter~\ref{knowledge} -- might ask the expert about the distribution of red and blue points in the deployed environment and learn that red points are more likely to have low feature values. Sample weighting, data splitting, and mischaracterization correction and detection could thus operate on the expert's distributional knowledge, conceptually (i.e., in the choice or execution of those techniques) or directly (e.g., by creating synthetic features). Even in the context where the red/blue distinction exists only in $\smplposs$, data curation provides methods for operating on what knowledge can be understood from domain experts and data to develop more trustworthy AI-enabled systems.

While the example in Figure~\ref{figs/2d-dataset} is a simple abstraction, it is representative of a common scenario. Consider a dataset of satellite images taken in one geographic region (e.g., North America) but the images in the deployed environment are of another geographic region, or a dataset of technical articles concerning one domain (e.g., protein-protein interactions) when the deployed environment articles belong to a different biology domain. In such cases, a data scientist might understand $\smplfeats$ and $\smpllbls$ and might be assured that the structure and values of inputs in the deployed environment will conform to their understanding (satellite image pixels or technical article tokens, respectively). 

Without accounting for the choices that were involved in the creation of $\dataset$ -- without fully understanding the relevant aspects encoded in $\smplposs$ -- the resultant AI-enabled system may not be optimally trustworthy. The trustworthiness of an AI-enabled system is a function of its performance on the true distribution of inputs in the deployed environment, and the true distribution can only be understood by accounting for the stated ($\smplfeats$, $\smpllbls$) and unstated ($\smplposs$) aspects of the real-world entities or phenomena being modeled. 

\section{Identify selected characteristics of data points}
\label{data/fairness}

We consider the trustworthiness of an AI-enabled system according to the definition in Section~\ref{intro/scope}, which is similar to but distinct from related academic literature at the intersection of trust, AI/ML, and fairness. This literature often references protected or sensitive attributes, typically in relation to systems in which people are represented as data points (e.g., race, gender), with an aim toward developing AI-enabled systems that perform without systemic bias toward or against a given group. 

While this specific goal may be relevant to some mission sets, we contend that the conceptual framework of trust relative to bias and fairness is a special case of the concept of trustworthiness as defined in this report. Rather than focus on legally protected attributes, we conceptually generalize this idea to \emph{selected characteristics}: any aspects of the entity or phenomenon of interest that are to be considered of particular interest (e.g., the color of the data point, whether the ship has a mounted gun) and in need of special handling (e.g., performance must be equal for red and blue points). 

Fairness can therefore be understood in the context of data as defined throughout Section~\ref{data}: there may be some aspects of $\smplideal$ -- in the features directly or by proxy -- that are relevant to the performance of the system on the true distribution in the deployed environment. Given this conceptual generalization, the academic literature focusing on protected attributes nevertheless offers data curation techniques that can be used for any mission-relevant selected characteristics. 

Drawing from this literature, we consider three types of algorithmic fairness \cite{mehrabi2021survey}: 
\begin{itemize}
    \item \emph{Equal Opportunity}: The proportion of true positives should be independent of selected characteristics, given the true label.
    \item \emph{Equalized Odds}: The proportion of true positives and false positives should be independent of selected characteristics, given the true label.
    \item \emph{Demographic Parity}: The likelihood of a given label should not vary on selected characteristics.
\end{itemize}

Consider a dataset where each data point is a ship such that $\smplfeats$ contains encoded aspects of the ship and $d^y$ reflects the country of origin. Further consider the ship type as the selected characteristic ($\textit{ShipType}=\{$military, civilian$\}$).

A classifier satisfying Equal Opportunity, the most specific type, would yield the same rate of correct predictions for military as for civilian ships: $P(\hat{y}^i \vert \textit{ShipType}^a, y^i) = P(\hat{y}^i \vert \textit{ShipType}^b, y^i)$. That is, given the true label, the probability of predicting that true label does not change in the presence of the selected characteristic. 

The next most general type is Equalized Odds, where a satisfying classifier would yield an equal rate of correct and incorrect predictions for military as for civilian ships: $P(\hat{y} \vert \textit{ShipType}^a, y) = P(\hat{y} \vert \textit{ShipType}^b, y)$. That is, given the true label, the probability of predicting any label does not change in the presence of the selected characteristic. 

A classifier satisfying the most general type of fairness considered here, Demographic Parity, would yield an even distribution over labels (e.g., United States of America, People's Republic of China) given a $\textit{ShipType}$: $P(\hat{y} \vert \textit{ShipType}^a) = P(\hat{y} \vert \textit{ShipType}^b)$. That is, the probability of predicting a given label should be completely independent of a selected characteristic, regardless of the true label. 

A useful generalization of this performance consideration is the cost-sensitive metrics where the costs are defined for accuracy per selected characteristics value. For example, it is to identify a military vessel when it is a Chinese ship, etc.

These types of algorithmic fairness are framed relative to outcomes\footnote{As defined here, algorithmic fairness is also framed relative to classification. While this is rhetorically convenient, fairness is not restricted to classification problems. Demographic parity, for example, may be more generally stated as being satisfied when the performance of the AI-enabled system -- regardless of task or performance metric -- does not vary on selected characteristics.}: the probability of $\hat{y}$ given the selected characteristic. It is expected that this framing will be typical of real-world mission sets \footnote{One useful manifestation of this consideration in real-world scenarios is the application of cost-sensitive metrics where the costs are defined for accuracy per selected characteristic value; e.g., military vessel when it is also a US ship}. Rather than a data scientist selecting characteristics and pursuing algorithmic fairness arbitrarily, it is likely that a SME, decision-maker, or other stakeholder has defined some selected characteristic and some notion of fairness an AI-enabled system should satisfy (e.g., equal performance on civilian and military ships). 

Data curation techniques that support fair outcomes are a subset of techniques that support trustworthy AI-enabled systems, because fairness -- as defined here -- reflects an understanding of and accounting for the relationships among $\smplfeats$, $\smpllbls$, and $\smplposs$. In the example scenario, $\textit{ShipType}$ may be encoded directly in $\smplfeats$ or may only be represented in $\smplposs$ (with dependencies on features in $\smplfeats$). Related literature often describes the latter in the context of latent biases -- e.g., ZIP code as a proxy for race -- but this framing is simply an instantiation of the larger concept of stated and unstated aspects in the data. 

Data curation that supports fair outcomes is therefore data curation for accurate outcomes. Selected characteristics are aspects of real-world entities or phenomena that have been chosen as properties that should not impact performance in the deployed environment. A model may learn relationships between $\textit{ShipType}$ and the country of origin, but if a SME has determined that those relationships are inappropriate for the deployed environment -- that $\textit{ShipType}$ should not influence predictions -- then data curation to support algorithmic fairness is also supporting optimized performance on the true distribution of inputs and, thus, trustworthy AI.

\chapter{Leverage subject matter expertise to understand and transform data}
\label{knowledge}

In projects with strong data science components, data scientists primarily focus on developing ML models, analyzing data, finding trends, and managing data \cite{piorkowski2021ai}. These activities are supported by \emph{domain knowledge}, or relevant information about the project's area of interest, which can include an understanding of the project goals, the environment in which capabilities will be deployed, the contexts in which analyses will be interpreted, how data was collected, and what data values mean. Domain knowledge can be contained in documentation or intuited from the data, but it is typically elicited from SMEs: people with extensive knowledge and skills in a certain domain \cite{yarovoy_assessing_2020}.

Domain knowledge is critical for data curation, as datasets are inherently incomplete representations of the world (see Section~\ref{data/simple}). As part of their activities, data scientists typically develop an intuition about the data and the task and they use it to actively shape the data, building on a combination of their learned data science experience, the data, the available documentation, and interactions with SMEs \cite{muller2019data}. This intuition, while critical to data curation and derived from domain knowledge, is often informal and satisficing: data scientists typically learn only what they need to best perform the tasks as they understand them with the data they have available \cite{heger2022understanding}.

Though the development of this intuition is necessary for every project, there is no generally accepted framework for how data scientists incorporate domain knowledge into their projects. As a discipline, data science is sometimes depicted as interacting with reliable standards and practices, but it is inherently heterogeneous: each dataset and task is different, and the domain knowledge collection process requires collaboration between data scientists and SMEs \cite{passi2018trust}. While data scientists can develop domain knowledge on their own, it is typically the case that SME collaboration is necessary to ensuring trustworthy AI.

This collaboration can benefit from careful consideration of domain knowledge: what is needed and how best to collect it. Trustworthy AI, as defined in Section~\ref{intro/scope}, particularly benefits from such consideration, as optimizing performance to the true distribution in the deployed environment requires understanding the task, its background, and the deployed environment (Section~\ref{knowledge/common}); the data as a collection of points, including what values features can take and what labels mean (Section~\ref{knowledge/transformation}); and the dataset as a whole, including the expectations on the true distribution (Section~\ref{knowledge/true}). 

We propose twelve questions to which a data scientist needs to collect answers. They are not intended to be exhaustive but rather to illustrate the breadth of domain knowledge that is necessary for trustworthy AI. Some of the answers may be found by reading project and data documentation, and the remainder can be posed to a SME. Each question is titled for the reader's and data scientist's convenience, using terminology from data science. The questions, meanwhile, are written to be understandable by data scientists and SMEs alike.

The questions are organized into three sections, which mirror the structure of this chapter. The subsequent sections in the chapter expound on the purpose of each question and how a data scientist can elicit its answer.

\paragraph{Understand the task in the deployed environment.} See Section~\ref{knowledge/common}. These questions are a variation on the Heilmeier Catechism \cite{kerrigan_survey_2021, park2021facilitating, heilmeier1992some}.
\begin{enumerate}
\setcounter{enumi}{0}
    \item \textbf{Task:} What is the task? 
    \item \textbf{Current Approach:} How is the task currently done? What are current practices and workflows?
    \item \textbf{Current Limitations:} What limitations in the current approach will be addressed by an AI-enabled system? 
    \item \textbf{Performance Evaluation:} How will the approach be evaluated? What metrics are currently in use?
\end{enumerate}

\paragraph{Decide on data transformations.} See Section~\ref{knowledge/transformation}. These questions were derived, in part, from a close reading of Chicco et al. \cite{chicco2022tips}.
\begin{enumerate}
\setcounter{enumi}{4}
    \item \textbf{Systematic Noise:} What technology was used to measure and record values in the data, and how might that technology systematically introduce noise? Put another way, are any of the features or labels imperfect proxies and, if so, how do they systematically differ from the true features and labels?
    \item \textbf{Missing Values:} What does a missing value indicate (e.g., not measured, not recorded, or a negative result)? Are missing values completely random, or else are some observations more likely to have a missing value?
    \item \textbf{Allowable Values:} What is the allowable range or set of values for each feature? Does a certain value in one feature rule out the possibility of the same observation having an otherwise permissible value in another feature?
    \item \textbf{Domain-Specific Feature Engineering:} Is it possible (i.e., for a trained human analyst or researcher) to determine the label exclusively from the features present in the dataset? If so, what subsets or combinations of features would be most important to making the determination? If not, which important features are missing?
    \item \textbf{Selected Characteristics:} Are there dimensions of diversity in the data, either explicitly encoded in a feature or latent among the aggregate of features, along which reliable performance is key? 
\end{enumerate}

\paragraph{Elicit information about the true distribution.} See Section~\ref{knowledge/true}.
\begin{enumerate}
\setcounter{enumi}{9}
    \item \textbf{Data Coverage:} Is each possible input in the deployed environment similar to at least one data point in the dataset? How frequently is the AI-enabled system, once deployed, expected to encounter an input substantially different from all the available data?
    \item \textbf{Covariate Shift:} Are certain kinds of inputs more or less common in the deployed environment than in the data available for development? How does the true distribution of values for each feature differ from the distribution observed in the data?
    \item \textbf{Label Shift:} Would a feature vector, if observed in the deployed environment, have a different true label (i.e., “right" answer for the AI-enabled system to predict) than the same feature vector as labeled in the data? For instance, in the case of a binary label, are borderline cases more likely to be labeled with a 0 in the data but a 1 in the deployed environment? Are there other such biases in the data that should not be replicated in the behavior of the AI-enabled system?
\end{enumerate}

\section{Build common ground to understand the task in the deployed environment}
\label{knowledge/common}

The trustworthiness of an AI-enabled system requires that performance in the deployed environment meets expectations, typically those formed by performance in the development environment. Data curation helps set these expectations by shaping the training and validation data in the development environment, with the goal of optimizing performance on the true distribution of inputs in the deployed environment. Doing so, however, requires understanding the deployed environment, both explicitly (e.g., what will the deployed data look like?) and implicitly (e.g., what are the processes by which data in the deployed environment differs from the development environment?). In this context, it may seem that domain knowledge centers on the inputs to the AI system, but data is the product of choices about what to include and what to exclude (see Section~\ref{data/simple}). Data curation to support trustworthy AI, therefore, requires a deeper understanding of the problem as a whole.

\subsection{Domain knowledge about the deployed environment}
\label{knowledge/common/deployed}

The domain knowledge pertaining to the background and context of a project can be described in part as the answers to a variation on the Heilmeier Catechism \cite{kerrigan_survey_2021, park2021facilitating, heilmeier1992some}:
\paragraph{Task:} What is the task? 
\paragraph{Current Approach:} How is the task currently done? What are current practices and workflows?
\paragraph{Current Limitations:} What limitations in the current approach will be addressed by an AI-enabled system? 
\paragraph{Performance Evaluation:} How will the approach be evaluated? What metrics are currently in use?

While some projects may presuppose a task (e.g., retrieve relevant documents), successfully answering the first question requires a broader understanding of the task landscape, which may in turn shape the data scientist's understanding of the deployed environment or the chosen performance metrics (e.g., What does “relevant" mean? Do documents need to be ranked by relevance? Is there some top-$k$ for which relevance is most important?). Successfully answering the second and third questions serves a similar purpose. Even if the decision to design and implement an AI-enabled system has already been made, understanding the task prior to such a system or how a new AI-enabled system will fill a gap in existing workflows will help refine the understanding of the deployed environment and how the system will be understood by its users. Performance can therefore be understood relative to the problem that an AI-enabled system was attempting to solve. The fourth question most directly addresses the need to understand performance in the deployed environment, which in turn helps the data scientist ensure the system is optimized against such metrics.

Consider a system designed to classify documents into one of several categories, which was a task previously performed by human experts. Without the appropriate domain knowledge, a data scientist may design a system and evaluate its performance according to accuracy (i.e., the number of documents the system classified correctly). In this scenario, consider that the system has yielded 30\% accuracy on a set of validation documents. Isolated from domain knowledge, this result may appear to be undesirable.

Domain knowledge about the task in the deployed environment, however, may include relevant details like the importance of some categories over others,\footnote{The importance of labels relates to the interpretation of performance in the deployed environment, which is distinct from the prevalence or meaning of labels in the data, which relate to the properties of individual data samples (Section~\ref{knowledge/transformation}).} the accuracy of human experts on the same validation data (which may be worse than 30\%), the rate of agreement between experts classifying the same set of documents, or the contours of the task in practice (e.g., in the deployed environment, experts may first triage relevant and irrelevant documents prior to classification activities). While these pieces of domain knowledge may help with result interpretation or model design (which extend beyond the scope of this report), they also inform data curation for trustworthy AI. Curating data to support optimization for performance in the deployed environment requires understanding that environment and what performance means in its context.

\subsection{Building and maintaining common ground}
\label{knowledge/common/build}

The questions in Section~\ref{knowledge/common/deployed} can help shape collaborations with SMEs to support understanding the deployed environment, but they are not sufficient for incorporating domain knowledge into an AI-enabled system. Data science projects are typically situated in domains outside of computer science or statistics, such that the data-enabled end capabilities will be understood in a context separate from data science (e.g., chemistry, public health, military equipment). Communication gaps between data scientists and SMEs or other stakeholders are common in such contexts, and shared domain knowledge is therefore critical to project success \cite{hou2017hacking, piorkowski2021ai, zhang2020data}.

Sharing domain knowledge between data scientists and SMEs, however, is complicated by the quantity and diversity of collaboration. While more information may yield better results, comprehensive information sharing is inefficient \cite{mao2019data}. It is likely not possible or desirable, for example, to transform a data scientist into a chemistry expert, or a chemist into an expert data scientist. The goal of these collaborations is not to ensure each member of a project has complete understanding of the problem space, but to provide enough of a shared language to facilitate work \cite{mao2019data}. 

While the collaboration between data scientists and SMEs can be understood within the framework of a shared mental model, where translators or brokers help each side understand the other \cite{piorkowski2021ai}, we consider a common ground or third space between data scientists and SMEs: a hybrid environment that can be constructed at the boundary between disciplines \cite{mao2019data}. In this framework, common ground allows each side to ``compare, negotiate, and integrate goals, perspectives and vocabularies, as well as discuss shared meanings and protocols" \cite{mao2019data}. 

The creation of common ground is a bidirectional process that allows data scientists to learn about the project domain, allows SMEs to learn about data science relative to the project, and enables both groups to understand the terms, goals, and processes of one another. This notion of common ground is intentionally distinct from the typical unidirectional framing of data science projects, in which data scientists consume domain knowledge from SMEs \cite{passi2018trust, zhang2020data, muller2019data, hou2017hacking}. Common ground, by contrast, is an environment where data scientists and SMEs learn from one another: data scientists can better understand the domain relative to the project (e.g., via the questions outlined in Section~\ref{knowledge/common/deployed}), and SMEs can better understand data science (e.g., how real-world entities are simplified for AI/ML contexts, as described in Chapter~\ref{data}). As a result, data scientists can ask more nuanced or accurate questions of SMEs who can in turn provide more actionable insights to the AI development process. 

Through common ground, both sides understand the other and can therefore yield better outcomes for the project. In projects where common ground has not been established, conflict can arise, as SMEs may not understand ``the hurdles and contributions of data scientists and vice versa" \cite{piorkowski2021ai}. With common ground, however, data scientists and SMEs can make more effective data curation decisions. 

Once established, common ground must be maintained throughout the project lifecycle \cite{mao2019data}. System requirements may change over time, and it is possible that the initial understanding between data scientists and SMEs becomes out of sync with the needs of the project. Even if the requirements and the evaluation methods of the system remain consistent over time, however, common ground must still be maintained; as the data scientists and SMEs learn more and understand one another better, the shared vocabulary and goals must reflect these changes.

The questions in Section~\ref{knowledge/common/deployed} invite SMEs to describe the background and context for the task in their own words. The data scientist can then clarify the words' meaning. Together, the data scientist and SME arrive at the terminology of their common ground. As the data scientist proceeds to curate the data, each step aims to improve performance on the task, as performance and the task are defined in the common ground with the SME.

\section{Conduct bidirectional communication to choose data transformations}
\label{knowledge/transformation}

In this section, we describe questions that inform data transformations that the data scientist may take to begin curating the data (i.e., creating, from the data available, a new dataset that will be used directly for training and validating the model). Working with domain experts, data scientists must codify what is known about the data in order to facilitate data curation or model design decisions \cite{kerrigan_survey_2021, kornowicz2023aggregating}. Transformations include applying closed-form mathematical functions to individual features or labels, aggregating multiple features or labels into new ones, and removing data points \cite{chicco2022tips}. This section describes how to ask questions about individual features and labels, including steps the data scientist can take to prepare for initial and follow-up interactions with the SME.

\paragraph{Systematic Noise:} What technology was used to measure and record values in the data, and how might that technology systematically introduce noise? Put another way, are any of the features or labels imperfect proxies and, if so, how do they systematically differ from the true features and labels?

Some commonly used technologies, including biomedical measurement devices like electrocardiogram machines, introduce noise for which there are standard denoising techniques in the literature \cite{limaye2016ecg}. Once the data scientist learns what technology was used in dataset creation, they can research appropriate techniques.

\paragraph{Missing Values:} What does a missing value indicate (e.g., not measured, not recorded, or a negative result)? Are missing values completely random, or else are some observations more likely to have a missing value?

In preparation for asking a SME about missingness, the data scientist can first quantify and visualize the patterns of missingness. If a systematic pattern of missingness emerges, then it is unlikely that missingness is completely random. Also, if missingness is rare and completely random, then it may be ameliorated by removing (i.e., filtering) the observations or features with missing values.

If the SME and data scientist conclude that observations with missing values must be filled in and used, the data scientist can propose and experiment with imputation techniques. Simple imputation approaches include filling the missing values in a feature with a centrality statistic (e.g., mean, mode) and propagating nearby values for similar observations (e.g., copying a company's past month's sales volume to the current month). Where simple imputation falls short, a model can be trained to predict missing values. Regression and \emph{k}-nearest neighbors models are more interpretable, hence easier for a SME to review, but neural networks can be used, too \cite{chicco2022tips}. After exploring the options and settling on one or more candidate imputation methods, the data scientist should present the methods to the SME, along with examples of their effects on individual observations, for the SME to review.

\paragraph{Allowable Values:} What is the allowable range or set of values for each feature? Does a certain value in one feature rule out the possibility of the same observation having an otherwise permissible value in another feature?

In preparation for asking a SME about allowable values, the data scientist can compute the range or set of values for each feature across all observations in the original dataset. It may also be helpful to flag possible outliers for the SME to review. Chicco et al. list several techniques for identifying possible outliers \cite{chicco2022tips}.

\paragraph{Domain-Specific Feature Engineering:} Is it possible (i.e., for a trained human analyst or researcher) to determine the label exclusively from the features present in the dataset? If so, what subsets or combinations of features would be most important to making the determination? If not, which important features are missing?

If the SME recommends combinations of features, the combinations could be used as domain-specific features. To supplement, the data scientist can research other feature engineering approaches used on similar data, such as data of the same datatype or representing the same phenomena or objects in the world.

\paragraph{Selected Characteristics:} Are there dimensions of diversity in the data, either explicitly encoded in a feature or latent among the aggregate of features, along which reliable performance is key? 

These could be subpopulations within the data distribution across which users will expect similar performance. As an example from computational chemistry, pharmaceutical and industrial chemicals differ, but a model predicting a chemical's toxicity needs to perform similarly across both kinds.

Such dimensions may be selected characteristics, which are described in Section \ref{data/fairness}. If a selected characteristic is a feature or can be systematically derived from features (e.g., by combining multiple features in a formula), the data scientist can compute group parity metrics for each of the selected characteristics. Common metrics are described and implemented in the Aequitas tool \cite{saleiro2019}. 

If a selected characteristic is latent among the features and cannot be derived systematically, then it may be necessary to label the data with the selected characteristic, which falls outside the scope of data curation and this report.

\section{Gather information about the true distribution}
\label{knowledge/true}

Similar to the Department of Defense principle that its AI capabilities be reliable \cite{dodaiethicalprinciples}, we argue that an AI-enabled system can only be trustworthy when deployed into an environment for which its performance was optimized. To optimize for performance on the true distribution, actionable information on the true distribution is necessary. In this report, we enumerate three distinct kinds of knowledge about the true distribution, expressed as the answers to the following multiple-choice question: 

\begin{quote}
    \emph{Which of the following is known or assumed about the true distribution?}
    \begin{enumerate}[a.]
        \item The true distribution is assumed to be well approximated by a parametric distribution, and estimates for the parameters are available as domain knowledge.
        \item The data used for development is known or assumed to be a representative sample from the true distribution, possibly because it was collected in the deployed environment.
        \item There is a pre-specified subset of the data for development that is known or assumed to be representative of the true distribution, possibly because it was collected in the deployed environment.
        \item As in (c), there is a pre-specified subset of the data for development that is representative of the true distribution, but this subset lacks labels \textit{[only applicable in supervised learning]}.
    \end{enumerate}
\end{quote}

Typically, there is only partial knowledge of the true distribution in the deployed environment. In case (a), for instance, there may only be parametric descriptions of the marginal distributions of some features or labels, rather than a complete description of the joint distribution of all features and labels. In cases (b), (c), and (d), the true distribution is described implicitly via a finite-sized set of data points sampled from the true distribution (i.e., a representative sample). Especially in cases (c) and (d), the size of the set may hinder precise and accurate knowledge of the true distribution.

The data scientist is not assumed to be an expert on the deployed environment and therefore needs authoritative sources attesting that either in case (a) the parameters are sufficiently accurate or in cases (b), (c), or (d) the sample is actually representative.
Authoritative sources include SMEs, decision-makers about the AI development project, and well-established sources of population-level statistics like the U.S. Census (if the true distribution is the U.S. population) and CIA World Factbook (if the true distribution is a population of one or more countries around the world).

The remainder of this section is dedicated to case (a), a common AI development scenario in which the data is not observations or measurements in the deployed environment and does not accurately represent it. Three kinds of discrepancies are elicited in the following questions: data from only part of the true distribution, data whose feature vectors offer a skewed view of the true distribution, and data whose labeling scheme differs from the truth in the deployed environment. Once distributional knowledge has been obtained, techniques in Section~\ref{sec:split_normal} can be used to shrink the size of the validation set to free up more training data while keeping the validation set representative of the true distribution.

\paragraph{Data Coverage} Is each possible input in the deployed environment similar to at least one data point in the dataset? How frequently is the AI-enabled system, once deployed, expected to encounter an input substantially different from all the available data?

The purpose of this question is to assess how much of the true distribution can be represented in a validation set. For example, suppose the deployed environment is the Arctic Ocean and the available data is all from tropical seas. Some fraction of the true distribution of inputs may reflect the presence of sea ice even though none of the data points from the tropics have sea ice present. Only the ice-free parts of the true distribution can be represented in a validation set, and the performance of the system can only be measured on that fraction of the true distribution. When the representativeness of the validation set is limited, only limited trustworthiness can be demonstrated.

\subsection{Eliciting parameters of a true distribution that contrasts from the data's empirical distribution}
\label{knowledge/true/parameters}

\paragraph{Covariate Shift:} Are certain kinds of inputs more or less common in the deployed environment than in the data available for development? How does the true distribution of values for each feature differ from the distribution observed in the data?

This question aims to elicit information about covariate shift, which refers to a discrepancy between the distribution of feature vectors in the training data and deployed environment. To optimize for performance on the true distribution, data scientists must compensate for any such discrepancy using, for example, the resampling and weighting techniques described in Chapters \ref{sec:resample} and \ref{sec:reweight}, respectively.

In preparation for asking the SME about the true distribution, the data scientist can fit a probability distribution to the data and compute statistics. For example, if the empirical distribution of each feature is approximately unimodal and symmetric, a normal distribution might be a decent fit, and sample mean and standard deviation are relevant statistics to compute. SciPy \cite{scipystats} implements methods for fitting to many probability distributions.

It is likely that a SME will only be able to offer information on the true distribution of human-interpretable features. If a feature has semantic meaning, there may be records or SME intuitions on the relative frequency of each of its possible values in the deployed environment. It is less likely that such domain knowledge exists for features with no semantic meaning, unless the same featurization scheme has been used previously on data from the deployed environment. For example, coordinates of a vector embedding typically have no semantic meaning and are not human interpretable, so unless the same embedding scheme has been used widely, it is unlikely that there is information available on the features' true distribution. If all features lack semantic meaning, then eliciting information about the true feature distribution may require labeling data points with human-interpretable features for which information on the true distribution is known. Labeling is outside of the scope of this report on data curation.

\paragraph{Label Shift:} Would a feature vector, if observed in the deployed environment, have a different true label (i.e., “right" answer for the AI-enabled system to predict) than the same feature vector as labeled in the data? For instance, in the case of a binary label, are borderline cases more likely to be labeled with a 0 in the data but a 1 in the deployed environment? Are there other such biases in the data that should not be replicated in the behavior of the AI-enabled system?

This question probes the discrepancy between the task encoded in the data and the task of the AI-enabled system. In general, the AI-enabled system can only be optimized for the task that the data encodes. However, if the tasks differ in a structured way, then mitigation is possible. In the case of binary labels described in the question, the decision boundary for a classifier can be chosen with the deployed environment, rather than the training data, in mind. In effect, the model can be made universally more eager or cautious to predict a label of 1. Section \ref{reweight/labelbias} describes a technique for mitigating other structured kinds of bias in labels.

The data scientist can prepare to ask the SME this question by identifying example data points that are close to a decision boundary. These could be found manually or by training an initial model on all data points, then feeding them back into the trained initial model as inputs and computing the distance to a decision boundary. By presenting these borderline cases and their labels in the dataset to the SME, the data scientist can help elicit whether the decision boundary in the dataset matches the intended decision boundary of the AI-enabled system.

Discussing specific samples with a SME to understand features, their labels, and their relationship relative to the SME's domain knowledge may focus on edge cases or examples likely to be misclassified as a mechanism for understanding the data more broadly \cite{park2021facilitating}.

\subsection{A tool for eliciting distributional parameters}
\label{knowledge/true/shelf}

The Sheffield Elicitation Framework (SHELF) is effective for eliciting true distribution parameters from SMEs \cite{shelf}. The application asks each SME to estimate the true parameter and also to provide information on the uncertainty in their estimate. The interface is based on research into eliciting probabilities from experts \cite{o2006uncertain}. 

SHELF may be useful when interacting with the SME to answer questions in this section that seek estimates of uncertain quantities:
\begin{itemize}
    \item The fraction of the true distribution that is substantially similar to a point in the data
    \item Each feature's true distribution, as contrasted with its empirical distribution in the data
    \item Each label's true distribution, as contrasted with its empirical distribution in the data
\end{itemize}

Once distributional knowledge has been obtained, the data scientist can use rejection sampling \cite{Peng} to winnow the data down to a subset that matches the elicited estimated parameters of the true distribution. Thereafter, splitting techniques \cite{Reitermanova_2010} can be used to shrink the size of the validation set to free up more training data while keeping the validation set representative of the true distribution.


\part{Acting on Data with the Goal of Trustworthiness}

\chapter{Choose among statistical tools for data curation}
\label{statistical_tools}
A common theme in curating data for trustworthy AI is to recognize the differences between the data available for development, viewed in the aggregate, and the true distribution of inputs the AI-enabled system will encounter once deployed. Statistical tools can help translate between the data’s empirical distribution and the deployed environment’s true distribution. Such a translation makes it possible to train the AI-enabled system on the available data while optimizing it for performance on the true distribution. Different tools have different preconditions (e.g., the covariate shift assumption), as we detail in the following chapters.

Unless the training data is a representative sample of the true distribution, optimizing the AI-enabled system for performance on the true distribution will usually require resampling or reweighting\footnote{In this context, resampling refers to the process by which data points are selected from an existing sample; i.e., the data is a sample of the population, but certain data points are resampled to create a dataset more representative of the population's true distribution. Similarly, reweighting refers to the process by which the data points are assigned weights that differ from their (implicit) equal weight.} the training data to accommodate distribution shift. Data scientists must decide which strategy to employ, balancing system requirements of interpretability and model performance. Down-sampling is most interpretable because each data point contributes equally to the model training. Up-sampling and model-agnostic weighting offer the next-best interpretability; data points may contribute unequally, but the inequality can be fully explained as a translation from the original empirical distribution of the data to the deployed environment’s true distribution. Model-specific weights, whether pre- or dynamically computed, blend information from the distributional differences with information about the model architecture’s strengths and weaknesses. If the weights are precomputed, these two components can be analyzed post hoc and possibly somewhat disentangled. Dynamically computed weights may offer model performance gains over other methods, at the expense of interpretability.

The choice of statistical tool depends on which kind of distributional knowledge is available and which resampling or weighting strategy is desirable. The best statistical tool will make efficient use of the knowledge and data available. While some scenarios require the application of multiple tools in sequence, each statistical tool applied to the data and distributional knowledge loses some information, so the best approach is likely to be the one that minimizes the number of tools needed to understand and accommodate distribution shift.

\begin{figure}[ht]
  \centering
  \includegraphics[width=.65\linewidth]{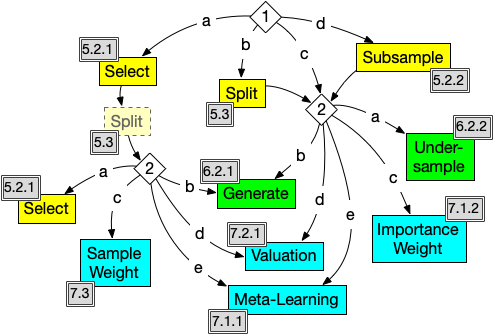}
  \caption{Data curation action decision tree, primarily centered around (1) what is known about the true distribution and (2) what data curation priorities are most important; actions are color-coded according to the relevant report chapters: Chapter~\ref{sec:splitting}, yellow; Chapter~\ref{sec:resample}, green; and Chapter~\ref{sec:reweight}, blue}
  \label{figs/decision_tree}
\end{figure}

We have organized data curation actions to enable trustworthy AI into a decision tree, shown in Figure~\ref{figs/decision_tree}.

Rather than offering a complete prescriptive framework for which data curation actions should be taken, the decision tree offers guidelines for data scientists considering alternative approaches. Sampling and weighting, for example, arise from the same conceptual framework (i.e., that some data points should be emphasized over others), but their mechanism of action is different, as are their trade-offs. Through the decision tree, data scientists walk through the two primary questions, following the choices at each, to arrive at a suggested set of tools for their project.

\section{Make strategic use of knowledge of the true distribution}
\label{sec:statistical_tools/decision_1}
The first decision point hinges on what is known about the true distribution, expressed as the answer to the following multiple-choice question (reproduced here from Section~\ref{knowledge/true} for easy reference):
\begin{quote}
    \emph{Which of the following is known or assumed about the true distribution?}
    \begin{enumerate}[a.]
        \item The true distribution is assumed to be well approximated by a parametric distribution, and estimates for the parameters are available as domain knowledge.
        \item The data used for development is known or assumed to be a representative sample from the true distribution, possibly because it was collected in the deployed environment.
        \item There is a prespecified subset of the data for development that is known or assumed to be representative of the true distribution, possibly because it was collected in the deployed environment.
        \item As in (c), there is a prespecified subset of the data for development that is representative of the true distribution, but this subset lacks labels [only applicable in supervised learning].
    \end{enumerate}
\end{quote}
Option (a) considers contexts in which the parameters of the true distribution are known, likely as the result of domain knowledge elicitation. Development data may or may not originate from the deployed environment, but the characteristics of data in that environment are understood. In such cases, data points can be selected (sampled) to conform to the expectations on the true distribution (Section~\ref{sec:optimal_rep_sample_weighting}), and (optionally) splitting strategies can be used to divide sampled data into training and validation sets (Chapter~\ref{sec:splitting}). This option highlights the trade-off between data representativeness and domain knowledge. In (a), data is not required or expected to be representative of the true distribution of inputs in the deployed environment, but to create representative training and validation datasets, domain knowledge (parameters describing the true distribution) must be elicited.

Option (b) considers contexts where the development data is known to be a representative sample of the true distribution of inputs, possibly but not necessarily because the data was collected from the deployed environment. These cases closely mirror the typical machine learning paradigm, where distinctions between development data (training and validation datasets) and evaluation data (the deployed environment) are assumed to be negligible (e.g., because the data originates from the same place or process).  Since the data considered in (b) is already a representative sample, strategies for splitting can be employed to produce representative training and validation datasets (Chapter~\ref{sec:splitting}). While domain knowledge elicitation is not required to parameterize the true distribution in (b), domain knowledge and statistical methods can be applied to the representative data to better understand the true distribution and thus enable splitting techniques.

In option (c), like option (a), it is not known whether the data is representative of the true distribution. Option (c), however, describes situations in which there exists a prespecified set of data points that \textit{are} representative of the true distribution. In this situation, the unknown data and representative data share an annotation scheme (e.g., both are labeled), and the unknown data is a larger set than the representative data. For example, there may be a large set of data points from benchmark datasets or other sources and a smaller set of representative data points drawn from the deployed environment. If the representative data is large enough and aligned well to the unknown data, it can be used directly as a validation dataset. In cases where the representative data is too small (e.g., a few examples from the deployed environment), typically there are not methods for learning about the true distribution to support data curation. In cases where the representative data is not well aligned to the unknown data, data specification methods (Chapter~\ref{sec:semantic_types}) can help support data curation to ensure that the features as they are understood in the unknown data conform to the expectations of features in the deployed environment.

Like option (c), option (d) posits the existence of two subsets of development data points: one that is known to be representative of the true distribution and one for which the representativeness is not known. The representative subset in option (d), however, is assumed to be unlabeled for the supervised task under consideration. For example, the unknown development data contains images annotated with the kinds of objects they depict, and the representative images are drawn from the deployed environment but are not annotated with object labels. In these contexts, the representative data can be used to understand the true distribution and inform curation actions on the unknown data, but it cannot directly be used as part of training or validation of a supervised system (since the representative data is unlabeled). Strategies for selecting a representative validation set include maximal representative subsampling (MRS) (Section~\ref{sec:maximal_rep_subsampling}), in which the unknown data is split into a smaller unknown set (discarded data points) and a validation dataset that is distributionally similar to the (unlabeled) representative data.

All four options lead to the second decision point, although the choice made in decision 1 impacts the options for decision 2. 

\section{Choose among possible curation actions on the training set}
\label{sec:statistical_tools/decision_2}
The second decision point focuses on data curation actions, expressed as the answer to the following multiple-choice question:
\begin{quote}
    \emph{Which of the following best reflects the priorities of interpretability and utility?}
    \begin{enumerate}[a.]
        \item It is most important that each training data point is real and contributes equally to model training.
	    \item Some data points can be synthesized from real data points, but all should contribute equally.
        \item Some data points can influence the model more than others, but the discrepancy should only correct the data's representativeness of the true distribution, not the model architecture.
        \item Data can be weighted according to distribution shift and model utility considerations, but all weights should be computed up front so that each training data point's contribution can be clearly quantified. 
        \item Model utility is most important; data weights can be uninterpretable.
    \end{enumerate}
\end{quote}

In option (a), the priority is to create training and validation datasets in which each data point is expected to contribute equally to model training. This is accomplished by winnowing the development data according to some criteria. If decision~(2a) was reached via decision~(1a), this winnowing takes the form of a second round of selection, where data is chosen for inclusion according to some criteria parameterized by what is known about the true distribution (Section~\ref{sec:optimal_rep_sample_weighting}). If decision~(2a) was reached via decision~(1b) or (1c), data can be undersampled, where the overall distribution of the development data is representative of the true distribution but some data points should appear less frequently than others to ensure equal weight among selected attributes (Section~\ref{down_sampling}).

In option (b), the priority on interpretability is relaxed a little to allow for generated data points, which recapitulate one or more of the original data points and increase the influence of those original data points (Section~\ref{up_sampling}).

Option (c) introduces the idea that some data points may have more value than others, and data curation involves determining and assigning weights to data points indicating the extent to which a given data point should influence the model. If decision~(2c) was reached via decision~(1a), the data curation action under consideration is sample weighting, in which auxiliary information about the true distribution (e.g., feature X in the deployed environment takes value Y at rate Z) is applied to compensate for the discrepancy between the development data and the expectations of the true distribution of inputs (Section~\ref{sec:reweight/raking}). If decision~(2c) was reached via decision~(1b) or (1c), importance weighting allows a data scientist to introduce weights for each data point in the model's loss function to indicate how important the data point is to the training objective (Section~\ref{sec:reweight/importance}).

Option (d) focuses on weighting the data points prior to training models such that the importance of each data point can be easily understood. In following this priority, it is assumed that some data points will be more useful than others in accounting for distribution shift or model utility. The data curation actions supporting this priority center on data valuation (Section~\ref{sec:reweight/initial_perf}). Data valuation techniques compute or approximate the impact of data points on model performance against the validation dataset, thus optimizing performance of the model against the expectations of the true distribution by way of interpretable precomputed weights.

Option (e) gives primacy to model utility: all that matters is optimizing on the true distribution of inputs in the deployed environment, even if the curation actions are uninterpretable. Data curation actions supporting this priority include meta-learning (Section~\ref{sec:reweight_dynamic}), in which model weights are repeatedly updated during the course of training to reflect the importance of a data point at a given time step. These meta-learning approaches can be useful, but it can be difficult to interpret data point utility since data point weights are dynamic during training. 

\subsection{Role of unlabeled data in curating labeled data}

If following decision~(1d), the available options for decision~(2) are notionally the same as with the other paths with additional nuance from the combination of labeled data (of unknown representativeness) and representative but unlabeled data. In this path, the data curation actions associated with decision~(1d) have produced a labeled validation dataset that mirrors the true distribution as it can be understood from the unlabeled representative data. The data curation actions associated with decision~(2), then, focus on leveraging true distribution knowledge to construct and weight the training dataset. 

Option (a) considers the scenario where the remaining unknown data (i.e., after the winnowing from decision~(1d)) is sufficiently large to use techniques like MRS (Section~\ref{sec:maximal_rep_subsampling}) again to produce a training dataset. That is, decision~(1d) leveraged the representative but unlabeled set of data points to winnow the unknown data into a representative validation set; decision~(2a) then repeats the process using that same representative set and the discarded points from (1d) to produce a representative training set. This approach should be taken with care, as -- depending on the size of unknown data remaining after decision~(1d) -- techniques like MRS may not yield a subset that is sufficiently representative or sufficiently large to train a ML model. 

Option (b) is similar to that of the other paths reaching decision~(2b): data points of unknown representativeness can be recapitulated to increase the training dataset size and, thus, increase the influence of those original data points (Section~\ref{up_sampling}). Unlike in other decision~(2b) contexts, however, there exists true distribution knowledge in the form of unlabeled data points. Leveraging these data points for use in generation or upsampling remains an open research question.

In option (c), the representative but unlabeled data can be used to describe the true distribution, and importance weighting (Section~\ref{sec:reweight/importance}) can leverage this information to introduce weights for each data point in the model's loss function. 

As with other paths leading to decision~(2d), option (d) in this context focuses on data valuation (Section~\ref{sec:reweight/initial_perf}), in which data points are weighted prior to model training. The result of decision~(1d) is a representative but unlabeled dataset, a curated, labeled (and notionally representative) dataset, and data with unknown representativeness (discarded from decision~(1d)). decision~(2d), then, discards the representative unlabeled data and seeks to apply data valuation techniques on the unknown data using the previously curated validation set.

Option (e), like option (d), discards the unlabeled representative data used in decision~(1d) and considers the application of meta-learning techniques (Section~\ref{sec:reweight_dynamic}) using the previously curated validation set.

The nuances of leveraging unlabeled data, and of the overall decision-making process, are summarized in the decision tree. The remaining chapters in Part II address each of the techniques in turn.

\section{Practitioner's perspective on the decision tree}
\label{statistical_tools/practitioners}
To illustrate how a data scientist might use the decision tree in practice, consider the following computer vision scenario.

Following the guidance in Chapter~\ref{knowledge}, knowledge was elicited and recorded about the AI-enabled system to be designed and implemented, including the task, the availability of data, and the true distribution of inputs in the deployed environment. In this scenario, we consider the task of object detection to determine whether a person is in the frame of an image. In the deployed environment, the AI-enabled system would be expected to identify, from uncrewed aerial vehicle (UAV) footage, people in a battlefield setting in need of medical help (triage). The available development data has two parts: a large, publicly available labeled dataset of UAV footage of people in public and a small labeled dataset of UAV images of manikins in a battlefield setting. The large public dataset is not representative of the true distribution of inputs, since the people in frame are not injured and are not in a battlefield setting. The manikin dataset is representative of the true distribution in some ways (manikins appear to be injured in the battlefield setting) but not others (manikins are only proxies for people).

Since, in this scenario, the manikin dataset is as close to the true distribution of inputs as possible, it could be considered to be a prespecified subset that is known or assumed to be representative. The manikin dataset is not large enough to train the model, however, so this case would fall under option (c) for decision 1 in the decision tree. The manikin dataset would be held out as a validation set that is representative of the true distribution in the deployed environment, so no data curation technique would need to be performed for decision point 1 in this case.

For decision 2, it is important that each training point is real as opposed to synthesized, and it is important that each data point contribute equally to the model. Therefore, option (a) would be chosen: undersampling will be performed to adjust any imbalances between the training and validation sets.

As a second example, consider a natural language processing scenario.

In this scenario, we consider the task of named entity recognition to identify and tag chemical and disease entities in text. In the deployed environment, the AI-enabled system would be expected to identify names of toxic chemicals and diseases in newly published scientific articles in \href{https://pubmed.ncbi.nlm.nih.gov/}{PubMed}.\footnote{https://pubmed.ncbi.nlm.nih.gov/} The available development data consists of a large historical dataset of scientific articles from PubMed, each annotated to identify and tag chemical and biological entities. However, this dataset may not be representative of the true distribution since the articles were published over a 100-year span and may not reflect the language and diversity of articles that are currently published. Therefore, domain knowledge about the true distribution of inputs in the deployed environment can be elicited by obtaining publication statistics from PubMed about scientific articles published in the last year.

In this scenario, the data available for development are not known or assumed to be representative of the true distribution. However, the true distribution can be assumed to be well approximated by a parametric distribution, and estimates for the parameters of this distribution are available in the publication statistics. Such parameters could be publication statistics on country of publication, institutional affiliation of author, and type of funding. Therefore, for this scenario, option (a) would be the best choice for decision 1 in the decision tree. A validation set of annotated articles that conform to the true distribution would be selected from the historical dataset of annotated PubMed articles via a method such as representative sample selection.

In considering decision 2 for this scenario, some data points can influence the model more than others to further improve the data's representation of the true distribution. Therefore, option (c) would be the best fit, where a technique such as representative sample weighting could be employed.

The decision-making process outlined for both cases should not be considered comprehensive or prescriptive. This scenario is intended only to illustrate how one might assess the data availability and knowledge of the true distribution in a project. The availability of data and knowledge of the true distribution drive the decisions, as opposed to the specific task dictating the decisions. The nuances of each option are discussed in the previous sections, and care should be taken to assess which options are best suited to the project at hand, focusing on the availability of data and knowledge of the true distribution, as opposed to the specific task.

Implementation of the splitting and resampling techniques for the first scenario are discussed in Sections~\ref{sec:practical_app_splitting} and~\ref{sec:resample/practitioners}, respectively. Assessment of mischaracterization was also applied to the computer vision problem, a discussion of which can be found in Section~\ref{mischar/practitioners}. Finally, the use of pretrained embeddings as a feature in the context of the first scenario is discussed in Section~\ref{subsec:practitioners_perspective_resnet}.

Implementations of representative sample selection and representative sample weighting are discussed in Sections~\ref{subsec:practioners_perspective_rsw_selection} and~\ref{subsec:practioners_perspective_rsw_weighting}, respectively. Debiasing pretrained embeddings was also performed for the second scenario and is discussed in Section~\ref{subsec:practitioners_perspective_wefe}.

\chapter{Split data into training and validation sets}
\label{sec:splitting}

Developing and testing a ML model generally requires three data partitions: training, validation\footnote{The validation set is sometimes referred to as a ``development'' set in the literature, but we avoid this terminology to minimize confusion.}, and testing. The goal of the latter two partitions is to sequester data that the model has not seen for testing and hyperparameter fine-tuning. Both training and validation sets are used in the development of an AI-enabled system; training data is used for learning parameters and validation data for setting hyperparameters. By contrast, testing data is intended for assessing the performance of the model after it has been developed. The roles of developer and tester are often divided among different individuals or organizations, especially for the development and testing of systems that will be deployed to perform highly consequential tasks. The tester is responsible for reserving testing data that is representative of the deployed environment. Meanwhile, the developer is incentivized to curate the data available for development in hopes of developing a model that performs well on the testing data they cannot see. This chapter describes techniques a developer can use to split out a validation set from the data available for development. We note that most of these techniques could also be used by testers to extract a testing set from a larger aggregate of data, though such actions fall outside of the data curation phase.

\section{Avoid leakage between training and validation sets}

Data leakage can occur when the model is able to learn unrealistic connections between the data in the training and validation environments that will not be present in the deployed environment. Data leakage commonly occurs when some of the training data is present in the testing set. If this is the case, the model can memorize these instances and might perform unrealistically well on them. A stealthier form of data leakage can occur if metadata that is present in all the splits is fed into the model. In this case, the model can incorrectly begin to learn from the metadata rather than learn what will be useful in the deployed environment. For an example of such data leakage, take a sentence-based toxicity classifier. If the sentences are parsed from the documents of the input corpus and then randomly shuffled before splitting, sentences from each document could end up in each split. While on its face this randomization might seem good, the model might start to pick up on interdependencies within the documents, such as the topic being discussed, rather than what the data scientist intended for it to learn. For example, if the documents are movie reviews, words such as titles of particularly disliked movies might be labeled as ``toxic," rather than truly toxic words. To avoid this form of data leakage, the documents as a whole should be partitioned into different splits and then the sentences should be parsed for processing. While this is just one example of data leakage, such hidden interdependencies can be present in many datasets and can be detrimental to the learning and validation of the model. It is imperative that there is no data leakage between splits. 

\section{Anticipate a distributional shift from the data to the deployed environment}
\label{subsubsec:split_distro_shift}
As discussed in Section~\ref{intro/truedistro}, it is crucial to understand the true distribution of the classes and features in the deployed environment in order to ensure success of the ML model. 

It is often the case that the amount of data needed for training and validation is not feasibly collected in the deployed environment, and thus the users of the model suffer because the data that the model is trained on does not accurately represent what the model sees in reality. An example is a model used in a classified environment. It is often very difficult, if not impossible, to collect the data that a model will use in a highly classified environment for testing, let alone training and validation. Because of this data scarcity, these models are often trained on unclassified data in the hopes that the unclassified data will approximate the deployed environment.  One way of combating this issue is to make sure that the training and validation sets reflect this distributional shift \cite{koh_wilds_2021}. Stanford's \href{https://wilds.stanford.edu/}{WILDS}\footnote{https://wilds.stanford.edu/} contains a series of datasets that display distributional shift between training and validation sets and can be used to train the models to withstand such a shift.

In cases where the development data is not known to be representative of the true distribution, selection methods can use knowledge of the true distribution to create (or sample) representative datasets. In this paradigm, a (notionally large) set of unknown data points is sampled to create a representative validation set. The remaining data points may then be sampled again to create a representative training set. Representative sample selection, which discards some portion of the available data, can be preferable to weighting when a large dataset size is infeasible (e.g., when more features will need to be collected on existing data points at some expense or when training is expected to scale undesirably with number of data points). These methods may also be preferable when interpretability is a priority because the individual contributions of each data point in the resulting datasets are uniform. 

Selecting a representative validation set from the data for development relies on knowledge of the true distribution, which may come in the form of (a) parameters of the true distribution or (b) a representative sample that may be inappropriate for use as development data (e.g., too few data points or unlabeled for the supervised problem of interest). Depending on the kind of true distribution knowledge available, methods for representative sample selection include optimal representative sample weighting and MRS.

\subsection{Representative sample selection}
\label{sec:optimal_rep_sample_weighting}

Optimal representative sample weighting frames the goal of generating a representative sample as an optimization problem, where, given parameters about the true distribution, weights are found that satisfy two objectives \cite{rsw}:
\begin{enumerate}
    \item Selected characteristics of data points with non-zero weights conform to the true distribution.
    \item Entropy of the weights is maximized, guarding against extreme weights.
\end{enumerate}
Conformity in objective (1) is defined over expected values of functions on selected characteristics. To conduct sample \textit{selection} as opposed to sample weighting, the number of data points, $k$, to be included in the resulting dataset must be specified. Then, the method assigns the representative selection's data points equal weight ($\frac{1}{k}$), and remaining data points are assigned weight $0$. This optimization problem is combinatorial and non-convex but can be well addressed heuristically with a solver based on alternating direction method of multipliers (ADMM).

\subsection{Maximal representative subsampling}
\label{sec:maximal_rep_subsampling}

Maximum Representative Subsampling (MRS) does not require the identification of selected characteristics, so it is useful when we do not know or do not want to specify which qualities of the deployed environment we want reflected in the training set \cite{maximalrepsubsampling}. MRS trains a classifier to differentiate between data of unknown representativeness and representative data. 
This is an example of positive-unlabeled (PU) learning: data points in the representative data are all positive cases, and observations in the unknown data could share characteristics with the true distribution (positive) or sufficiently diverge (negative). The predictions from the initial classifier are used to iteratively downselect the unknown data and retrain the classifier until the classifier cannot distinguish between the representative data and the winnowed unknown data. One benefit of the MRS method is that the distance between the winnowed unknown data and the representative data is decreased in reproducing kernel Hilbert space, which captures nonlinear relationships and interaction effects across the entire feature set. One example MRS implementation is available in Python in the \href{https://github.com/kramerlab/WeightDebiasing/tree/master}{Weight Debiasing}\footnote{https://github.com/kramerlab/WeightDebiasing/tree/master} project.

\section{Split the data while preserving its diversity}
\label{sec:split_normal}
In real-world data, even if the testing set is assumed to be given, there is usually a reason to split the data to ensure that the model is properly learning. Whether the validation set is used for fine-tuning or early stopping, it is crucial that the data is partitioned in a way that is fair and is representative of the true distribution.  Meng et al. demonstrate that choice of splitting method can have a large effect on the performance of the downstream model \cite{Meng_McCreadie_Macdonald_Ounis_2020}. Below are some commonly used splitting methods. None of these methods correct for the true distribution, and, thus, it is assumed that the larger dataset from which the splits are being drawn is reflective of the true distribution. 

When splitting, it is important to keep in mind the selected characteristics that a SME has identified as features that should have equal performance (see Section ~\ref{data/fairness}), as well as proxies for selected characteristics. For the final model to be fair, the selected characteristics must be represented across all of the splits.

\begin{itemize}
    \item \textbf{Simple random sampling (SRS)} splits the data via random sampling along a uniform distribution \cite{Reitermanova_2010}. Should the features be evenly distributed across the dataset, SRS is a simple and quick way to partition the data. However, it is unlikely that the data will be evenly distributed across all of its features. If this is the case, the SRS may distribute the data unequally across the partitions. 
    \item \textbf{Trial and error} tries to overcome the diverse model performance issues using SRS by performing SRS multiple times. It should be noted that this method is resource intensive \cite{Reitermanova_2010}. Trial and error can be a useful approach on small and diverse datasets where training a model using a different SRS split has a nontrivial effect on the output. 
    \item \textbf{Systematic sampling} is an approach for naturally ordered datasets (e.g., time series). $\mathcal{D}$ is first ordered, then a random starting data point is picked. Each successive $\mathit{k}^\mathit{th}$ data point  is chosen where $\mathit{k}$ is calculated based on the size of the dataset. Reitermanová notes that it can often be hard to find a true ordering for a dataset, and if the dataset is misordered, then systematic sampling is inherently the same as SRS. SRS could work for time intervals when the time intervals are ordered and the data is similarly distributed within each time interval. Then data points can be randomly sampled from each time interval in successive order. This method is called \textbf{convenience sampling} \cite{Reitermanova_2010}. State-based data, such as time series data, provides the data scientist with an opportunity to understand why it is so important to mimic the true distribution. Should $\mathit{k}$ yield a data point every five minutes, when in fact a data point in the deployed environment might be fed into a forecasting model every minute, the model would almost certainly overpredict. Therefore, the choice of $\mathit{k}$ in this splitting method must reflect the data that the model will see in the deployed environment. 
    \item  \textbf{CADEX and DUPLEX (extension of CADEX)} are methods that select data points for the split based on Euclidean distance in the feature space. Starting with the two farthest data points in the dataset, data points that are the farthest from the previously selected data points are iteratively selected \cite{Joseph_Vakayil_2022} for the split. These methods aim to capture all of the variance within $\mathcal{D}$ in each split.
    \item \textbf{Stratified sampling} automatically explores the internal structure and distribution of the dataset to create homogeneous groups of data points that are then sampled from to create even splits. These groups are usually created using clustering algorithms, which means that the selection of hyperparameters for these clustering algorithms can have significant downstream effects on the quality of the split and therefore the performance of the final model. Once the data points are split into relatively homogeneous clusters, data points from each cluster can be selected with uniform probability. scikit learn, a commonly used Python library for data science, has a \href{https://scikit-learn.org/stable/modules/generated/sklearn.model_selection.StratifiedShuffleSplit.html}{StratifiedShuffleSplit object (SSS)}\footnote{https://scikit-learn.org/stable/modules/generated/sklearn.model\_selection.StratifiedShuffleSplit.html} that implements this strategy.
\end{itemize}

\section{Practitioner's perspective on splitting}
\label{sec:practical_app_splitting}

\subsection{SSS in scikit-learn}
\label{subsec:practitioners_perspective_sss}

As a practical experiment in splitting data, we tested the Stratified Shuffle Split functionality of the scikit learn Python package using a dataset with quantitative features. The SSS algorithm combines the stratification of Stratified K-Fold splitting and the randomization of Shuffle Split to create stratified randomized folds. The algorithm is described in more detail in Section~\ref{sec:split_normal}.

The SSS method produced folds where -- within each fold -- the average of each quantitative feature for the training set and that of the validation set were the same as that of the original dataset. For example, if the average for Feature 1 was 0.5 in the original dataset across all data points, the average of Feature 1 for each training set and for each validation set in each resulting fold was as close to 0.5 as possible.

As a point of comparison, we also used the \href{https://scikit-learn.org/stable/modules/generated/sklearn.model_selection.train_test_split.html}{``train\_test\_split''}\footnote{https://scikit-learn.org/stable/modules/generated/sklearn.model\_selection.train\_test\_split.html} function in scikit kearn on the same dataset, which randomly splits the data into training and validation sets. This function produced very uneven splits, such that the average for a feature in the training and validation sets could be as little as half or as great as double the average of the feature for the original dataset. From this perspective, the SSS algorithm better preserves feature diversity across all training and validation sets.

SSS can also be used for nonquantitative features, like image embeddings. In this example, the algorithm clusters similar image embeddings into groups and then selects data points such that the diversity of the groups is preserved across all splits and folds. However, because image embeddings are not human interpretable, examining the evenness of the fold will require selecting several images from each set and qualitatively assessing whether the diversity of the data was preserved across all sets and folds as compared with the original dataset. After implementing this technique, our qualitative analysis found the diversity of the data to be preserved across all sets within all folds when using SSS with image embeddings.

An implementation detail is that the random-state variable must be set to an integer in order to reproduce results. In addition, the shuffle element of the algorithm does not guarantee that there will not be overlapping samples between validation sets. If a use case requires mutually exclusive validation sets between the folds, another algorithm that does not incorporate shuffling should be employed.

In contexts where there is ordered data that does not have evenly distributed samples, SSS may not be the best splitting approach. Shuffling could result in overfitting and inflating the validation set. Using another splitting algorithm without shuffling, such as scikit learn’s K-Fold Splitting which does not shuffle the data by default, may be a better fit for the use case of ordered data.

Additional implementations of algorithms for splitting data into training and validation sets can be found in scikit learn's \href{https://scikit-learn.org/stable/modules/cross_validation.html}{documentation of cross-validation}\footnote{https://scikit-learn.org/stable/modules/cross\_validation.html} and \href{https://scikit-learn.org/stable/auto_examples/model_selection/plot_cv_indices.html#sphx-glr-auto-examples-model-selection-plot-cv-indices-py}{visualization of cross-validation}.\footnote{https://scikit-learn.org/stable/auto\_examples/model\_selection/plot\_cv\_indices.html\#sphx-glr-auto-examples-model-selection-plot-cv-indices-py}

The scikit learn package is actively maintained as of December 2024, and instructions on installation and use can be found on \href{https://scikit-learn.org/stable/index.html}{scikit learn's website}.\footnote{https://scikit-learn.org/stable/index.html}

\subsection{Representative sample selection in rsw}
\label{subsec:practioners_perspective_rsw_selection}

The \href{https://github.com/cvxgrp/rsw}{rsw} Python package implements the approach to representative sample weighting and selection that is described in more detail in Section~\ref{sec:optimal_rep_sample_weighting}. To use rsw, clone its repository, then install via the setup script. The package exposes a single method -- \texttt{rsw} -- to which the user can pass input data, functions on selected characteristics that should be balanced, corresponding target values for those functions, and a regularizer. The \texttt{rsw.BooleanRegularizer(k)} regularizer enforces the selection of a uniformly weighted sample of size $k$ (the task of representative sample \textit{selection}, as opposed to weighting). 

To experiment with representative sample selection, 500 PubMed articles from the \href{https://huggingface.co/datasets/bigbio/bc5cdr}{BioCreative V Chemical Disease Relation (CDR)}\footnote{https://huggingface.co/datasets/bigbio/bc5cdr} corpus and 3,500 PubMed articles from the \href{https://huggingface.co/datasets/bigbio/chemdner}{CHEMDNER}\footnote{https://huggingface.co/datasets/bigbio/chemdner} corpus were randomly selected as a dataset. Publication dates range from 1974 to 2017. For the purpose of this exercise, the deployed environment is thought to contain articles added to PubMed in 2023. Metadata about this subset of PubMed can be inferred by comparing the \href{https://datadiscovery.nlm.nih.gov/Literature/MEDLINE-PubMed-Baseline-Statistics-Misc-Report/tap4-sm6y/about_data}{MEDLINE/PubMed Baseline Statistics Reports}\footnote{https://datadiscovery.nlm.nih.gov/Literature/MEDLINE-PubMed-Baseline-Statistics-Misc-Report/tap4-sm6y/about\_data} from 2022 and 2023. Selected characteristics are (i) proportion of articles that report on a randomized controlled trial and (ii) proportion of articles that benefited from grant money from organizations in the United States.

$k$=800 selects 20\% of the training data for validation. After running rsw, the distribution of the selected characteristics in the subset of training set selected for validation is closer to the deployed environment than the training set. Results are shown in Table~\ref{tbl:rsw_selection_results}.

\begin{table}[ht]
\begin{center}
\begin{tabular}{ |p{4.25cm}||p{2.5cm}|p{2.5cm}|p{2.5cm}|p{2.5cm}| }
 \hline
 \multicolumn{5}{|c|}{Observed Values of Selected Characteristics} \\
 \hline
  & \textbf{Deployed\newline Environment} & \textbf{Original\newline Training Set} & \textbf{Training\newline Samples\newline Selected\newline for Validation} & \textbf{Training\newline Samples\newline Not Selected\newline for Validation}\\
    \hline
    \% RCT & 2 & 2 & 2 & 2 \\
    \% Receiving U.S. Money & 23 & 16 & 21 & 15 \\
 \hline
\end{tabular}
\caption{The distribution of selected characteristics (whether the research concerns randomized control trials (RCT) or received US grant money) in the validation set approached that in the deployed environment. Values shown for the training set reflect all 4,000 observations. Values shown for the validation set reflect the 800 observations selected from the training set using rsw.}
\label{tbl:rsw_selection_results}
\end{center}
\end{table}

Moreover, of the training set articles not selected to be in the validation set, 2\% are about randomized controlled trials (RCT) and 15\% are by authors who received grant money from organizations in the United States. Representative sample selection essentially divided the training set into a partition that looks more like the deployed environment and a partition that looks less like the deployed environment.

When using this software, it is worth keeping in mind that the greater the ratio of training set size to $k$, the more candidates there are for every one member of the validation set. When target values for functions on selected characteristics deviate dramatically from the training set, or when many functions are defined, results will be better when the algorithm can be choosier.

\chapter{Resample data to compensate for overt or latent class imbalance}
\label{sec:resample}

In the context of data curation, \emph{imbalance} refers to disproportionate representation between encoded aspects (labels, selected characteristics, or other features)\footnote{In machine learning literature, the most typical term is class imbalance, referring to datasets in which some labels are much more common than others, but we consider imbalance across any encoded aspect.}. In real-world datasets, there are often multiple forms of imbalance, but when developing a trustworthy AI-enabled system it is important to consider the sources of imbalance and mitigate them when doing so will enable performance on the true distribution of inputs. 

ML models learn generalizations about the data. When all features are weighted equally, class imbalance is mainly a concern when the label is imbalanced. When it comes to trustworthiness, depending on the use case, a data scientist may want to artificially increase the number of data points with uncommon feature or label values to force the model to learn more accurate representations of the minority classes.

Proper resampling can combat data imbalance. Without resampling in the face of a data imbalance, a small group may effectively be ignored by the model, since it will play only a small role in the loss function. This in turn can lead to differing performance across groups and poor performance on minority classes.

Resampling is just one approach to handling data imbalance. Chapter~\ref{sec:reweight} discusses weighting as another option that can be used with or without resampling. The benefits of these methods are compared briefly in Section~\ref{learn_sample}. 

\section{Decide whether resampling is appropriate}
\label{sec:calibration}

While data sampling clearly has its place in the data valuation pipeline, it is important to understand when sampling is appropriate or necessary and when it is not.
One of the main practical issues stemming from data sampling is its effect on model calibration.
Sampling increases the ratio of minority-class data points to others in training time.
While this can increase the model's ability to learn a decision boundary, it tends to also increase the overall likelihood that it predicts a minority class.
If the minority class is indeed infrequently present in the deployed environment, there will be a mismatch.
This mismatch between the perceived likelihood of a class and its actual likelihood is called miscalibration. 

Consider the classic example of an imbalanced learning problem: fraud detection.
Instances of fraudulent transactions are thankfully very scarce, say less than 1\% of transactions.
Sampling may be used to train a model to learn a good decision boundary between fraudulent and legitimate transactions.
However, after applying sampling, the fraction of fraudulent transactions in the training data may be increased to very high levels, say 20--30\%.
In this scenario, it is possible for a classifier to learn that fraud is much more common than it is.
Thus, sampling can, in some scenarios, create a mismatch between the training set's empirical distribution and the true distribution, leading to an untrustworthy model.

While changes in model calibration due to data sampling can be problematic, they can also be used to a data scientist's advantage in creating more trustworthy models.
If, for example, it is known that a class will be more prominent in the true distribution than it is in the training set's empirical distribution, sampling could be applied to correct for this and create a more trustworthy model.
Such knowledge could be obtained from SMEs, as described in Section \ref{sec:distributional_knowledge}.
This also overlaps with the correction of data based on knowledge of the distribution, which is further discussed in Section \ref{sec:distributional_knowledge}.

Whether or not the goal is to change a model's calibration, its calibration should be tested and evaluated if sampling is applied as part of the data valuation process.
This testing and evaluation can be facilitated through proper data splitting, as previously discussed.

\section{Resample data to compensate for class imbalance}
\label{sec:resampling}

Once it is decided that the dataset needs to be resampled, the goal is to use various resampling methods to change the relative proportion of classes that a model sees during training, testing, or both. It is up to the data scientist to decide what that proportion should be, and that decision should be informed by model performance on the validation dataset as an estimation of performance on the true distribution. 

\subsection{Up-sampling by synthesizing data} 
    \label{up_sampling}
    Up-sampling involves adding data points to the minority classes to achieve balance within the dataset. Many up-sampling methods involve some form of data synthesis rather than simply duplicating minority-class data points.
    \begin{itemize}
        \item \textbf{Random over-sampling (ROS)} randomly selects data points from the minority group to duplicate \cite{johnson_survey_2019}. This method is useful if the data is quite sparse and there is not much to consider when it comes to up-sampling. Generally, however, it is better to synthesize data rather than just repeat existing data points so that the model can learn generalizations instead of memorizing specifics about the few examples from the minority class. The following up-sampling or data synthesis methods are built to account for more complex or higher-dimensional data.
        \item \textbf{Synthetic minority over-sampling technique (SMOTE)} \cite{chawla_smote_2002} is a method that synthesizes examples from the minority class by interpolating between random data points from the minority class and those data points' nearest neighbors.
        \begin{itemize}
             \item \textbf{Borderline-SMOTE} applies SMOTE only to the data points near class boundaries.
            \item \textbf{Safe-Level-SMOTE} does the opposite of what Borderline-SMOTE does: it defines ``safe" regions to avoid sampling from noisy regions or outliers.
        \end{itemize}
        \item \textbf{Model-based synthesis (MBS)} uses regression models to synthesize data. These models help capture feature relationships and diversity in the synthesis process \cite{Liu_Hsieh_2020}.
        \item \textbf{ADASYN} adaptively generates ``minority data points according to their distributions: more synthetic data is generated for minority class data points that are harder to learn compared to those minority data points that are easier to learn" \cite{he_adasyn_2008}.
        \item Gazzah and Amara \cite{gazzah_new_2008} create synthetic points using an ensemble of interpolation methods. They use four different curve-fitting methods to fill the minority feature subspace and generate diverse minority data points.
        \item Tiwald et al. \cite{Tiwald_Ebert_Soukup_2021} use a generative deep neural network (GNN) with an extra fairness loss to create synthetic data. They find that models trained on this synthetic data maintained accuracy on par with models that were not trained on debiased data but performed nearly equally along selected characteristics, whereas the biased data had a lower performance on minority selected characteristics. The outcome of this method proved successful, although the black-box approach of using a DNN loses accountability. Using this approach over statistical sampling methods might lead to difficulty in describing \emph{how and why} the ``debiased" data no longer contains bias.
    \end{itemize}
    
\subsection{Down-sampling data}
    \label{down_sampling}
    Down-sampling involves removing data points from the majority class to achieve balance. 
    \begin{itemize}
        \item \textbf{Random under-sampling (RUS)} randomly discards data points from the majority group  \cite{johnson_survey_2019}. As with ROS, RUS is best used when the features are evenly distributed across the dataset; otherwise, more complex methods account for noisier data.
        \item \textbf{Near-miss} \cite{johnson_survey_2019} uses \emph{k}-nearest neighbors to select majority-class data points to remove based on their distance to the minority-class data points. The goal of near-miss is to remove borderline data points.
        \item \textbf{One-sided selection} removes noisy or near-duplicate data points from the majority class using various noise detection algorithms \cite{Kubat_Matwin_1997}.
        \item \textbf{Wilson's editing} is a form of noise detection and removal in the majority class \cite{Wilson_1972}. For more on mischaracterized data cleaning, see Section~\ref{mischar}.
    \end{itemize}

\section{Resample data iteratively during training}
\label{curriculum_learning}
Another approach to handling imbalanced data involves exposing the model to ``harder" data later in the training process. This approach, ``curriculum learning," follows the logic of human learning -- curricula begin with easier concepts and slowly progress to the hardest concepts. Similarly, the data is partitioned and the model is trained in phases such that the model first learns generalizations on the ``easy" (majority-class) data and is later fine-tuned for the task at hand with the smaller samples of relevant ``hard" data \cite{Wang_Chen_Zhu_2021}. 

This method can be applied to balanced data. Take the example of classifying military versus civilian ships. Although this dataset might be balanced (50\% military and 50\% civilian), some images might be taken from a distance, while others have only the ship in them. The photos taken from a distance might be considered ``harder" for the model to learn than the photos taken from closer locations. Curriculum learning proposes first training the model on the close photos so that the model might begin to learn which features indicate the presence of a military ship, and then adding the harder photos during a final training to increase the robustness of the model and account for the true distribution, which might include photos taken from a distance. The same concept applies to imbalanced data. If military ships are a minority class, it might be beneficial for the image model to first train on all of the other pictures so it can learn the features of a ship and the environment and then later fine-tune to the classification task at hand. 

Curriculum learning can be applied to any model once the data has been partitioned into increasingly more ``difficult" classes. Curriculum learning has been found to result in more general models with better discriminatory capabilities \cite{Wang_Gan_Yang_Wu_Yan_2019}. Curriculum learning avoids over-sampling, which could lead to overfitting given the magnitude of the class imbalance.

Class imbalance is a common problem within the field of ML. If  resampling is feasible, it can be a simple solution to this issue. However, resampling can lead to overfitting or throwing away useful data. Therefore, other methods, such as model recalibration or curriculum learning, might be better suited for the dataset. Another approach to handling data imbalance is discussed in Chapter~\ref{sec:reweight}.

\section{Practitioner's perspective on down-sampling with imbalanced-learn}
\label{sec:resample/practitioners}
We experimented with under-sampling using the Python \href{https://imbalanced-learn.org/stable/index.html}{imbalanced-learn}\footnote{https://imbalanced-learn.org/stable/index.html} package, which is actively maintained as of December 2024. The most common and applicable of the under-sampling algorithms is \href{https://imbalanced-learn.org/stable/references/generated/imblearn.under_sampling.RandomUnderSampler.html#imblearn.under_sampling.RandomUnderSampler}{RUS}.\footnote{https://imbalanced-learn.org/stable/references/generated/imblearn.under\_sampling.RandomUnderSampler.html} Given a binary unbalanced class and using the default parameters, RUS randomly removes data points from the majority class until the two classes are balanced. The same can be done for multiclass problems, in which case each class is balanced separately.

Several sampling strategies can be used with RUS. By default, all classes but the minority class will be resampled to balance the dataset, and all classes will be balanced equally; since data points are removed, the overall size of the dataset will decrease. Users can specify the sampling strategy to be ``majority" (resampling only the majority class), ``not minority" (default), ``not majority" (resampling all class but the majority class), or ``all" (resampling all classes). A target ratio can also be provided for binary classification (as opposed to the default 0.5, which balances 50/50). The same can be done for multiclass problems, but a dictionary of data point target counts for each class must be provided.

In addition to class label imbalance, under-sampling can also be used to adjust an imbalance of a selected characteristic (defined in Section~\ref{data/fairness}). For example, consider an object detection task where a selected characteristic is whether an image is zoomed in on the object to be detected or not. The concept of "zoomed" could be quantified for each image by identifying the percentage the object bounding box occupies in the entire image (given that the images are all the same size or scaled to the same size). Labels for zoomed versus non-zoomed would then be obtained by setting a threshold for the percentage that qualifies as zoomed. 

For this example, the user would want the model to perform equally well on both zoomed and non-zoomed images. However, the ratio of zoomed images to non-zoomed images may be different in the deployed environment than in the training data. To ensure that the model performs equally well on the selected characteristic, the training data can be under-sampled such that the ratio of zoomed to non-zoomed images in the training data is the same as in the validation set that has been selected as representative of the deployed environment. The same process used to balance class labels with RUS can be applied to a selected characteristic. Zoomed versus non-zoomed would be used as the label, and the ratio passed to the sampling strategy parameter would be the ratio of zoomed to non-zoomed images in the validation set.

While we only tested the RUS algorithm, imbalanced-learn implements several other under-sampling algorithms, which fall into two categories: generation and selection. The generation technique \href{https://imbalanced-learn.org/stable/references/generated/imblearn.under_sampling.ClusterCentroids.html#imblearn.under_sampling.ClusterCentroids}{Cluster Centroids}\footnote{https://imbalanced-learn.org/stables/references/generated/imblearn.under\_sampling.ClusterCentroids.html} is still considered an under-sampling method despite using synthesized data in the final selection. Rather than using a subset of the original data, cluster centroids applies a \emph{k}-means method, thereby generating centroids that are used to synthesize a new set of data. 

All other imbalanced-learn under-sampling algorithms fall into the selection category. One such example is the \href{https://imbalanced-learn.org/stable/references/generated/imblearn.under_sampling.NearMiss.html#imblearn.under_sampling.NearMiss}{near-miss}\footnote{https://imbalanced-learn.org/stable/references/generated/imblearn.under\_sampling.NearMiss.html} technique, which incorporates heuristic rules through the nearest-neighbors algorithm. Several near-miss versions are implemented in imbalanced-learn, with varying levels of susceptibility to noise. imbalanced-learn also implements several cleaning under-sampling methods that aim to identify and remove noise (related to concepts in Chapter~\ref{sec:distributional_knowledge}).

More information on all of these concepts can be found in the imbalanced-learn user guide section on \href{https://imbalanced-learn.org/stable/under_sampling.html#}{under-sampling}.\footnote{https://imbalanced-learn.org/stable/under\_sampling.html}

The imbalanced-learn package also provides methods for over-sampling, including naive ROS, SMOTE, ADASYN, and SMOTE variants. While we did not test the over-sampling functionality of the package, readers can find more information on the implementation of these algorithms, along with best practices for choosing an algorithm, in the imbalanced-learn user guide section on \href{https://imbalanced-learn.org/stable/over\_sampling.html}{over-sampling}.\footnote{https://imbalanced-learn.org/stable/over\_sampling.html}

\chapter{Weight training data based on representativeness, uniqueness, and utility}
\label{sec:reweight}
Suppose, as previously, that there is a (hopefully large) set of data available for training a model that will be part of an AI-enabled system, as well as a (likely smaller) dataset that is approximately a representative sample from the true distribution of inputs the model will encounter in the deployed environment. The training data can be seen as a sample from some distribution, too. The challenge of trustworthiness is to train a model that performs optimally in the deployed environment even though it was trained on data from the training distribution, which is generally different from the true distribution.

One way to address this challenge is to apply weights to each training data point to capture the relative influence each training data point should have during the training process.
Intuitively, some regions of the true distribution might be poorly represented or might exhibit important but nuanced differences between data points, and up-weighting the training data points from these regions can draw attention to those regions in the training process, hopefully yielding a model that performs optimally across the true distribution.
This section introduces several ways that this intuitive idea has been formalized and implemented, highlighting the trade-offs between them.

Note that weighting training data can serve two goals.
The first, more traditional, goal is to assist the model's performance on the true distribution when there is no difference between the development data and true distributions.
This is especially important in the case of imbalanced learning, where models trained without resampling or weighting the data yield poor results on the minority class.
In the case of classification, data weigthing emphasizes the importance of the boundary surrounding the minority class.
See Anand et al. \cite{anand2010approach} for one example of this in practice.
The second potential goal of data weighting is to assist the model's performance on the true distribution when the development data distribution differs from it in some way.

Among the many metrics that might be selected to measure performance, average accuracy across the true distribution is one option. It is estimated by computing the empirical accuracy on the (likely smaller) dataset that is approximately representative of the true distribution. Most data weighting techniques assume the performance metric is average accuracy or another metric estimated by computing an average of some value across a dataset. This assumption limits the direct applicability of many weighting techniques to projects in which the performance metric is not an average.

\section{Weight data points agnostically to the model architecture}
\label{sec:reweight_model_agnostic}

Model-agnostic weights aim to compensate for distribution shift (sometimes called dataset shift), in which the data available for training a ML model is composed of samples from a different distribution than the true distribution the AI-enabled system (encompassing the trained model) will encounter in the deployed environment. 

Conceptually, there are two types of methods by which model-agnostic weights can be found. If auxiliary information about the true distribution (and, critically, the feature value probabilities therein) is available, statistical methods can be applied to compensate for the discrepancy between the true distribution and the empirical distribution. Methods that do not directly leverage this kind of auxiliary information tend to learn an initial (or throwaway) model\footnote{Typically, the architecture of an initial weight-learning model does not match the model to be trained for use in the AI-enabled system, and the weights themselves impose no requirements on that downstream model architecture.} to determine weights; information derived from initial models must be interpreted as information about the empirical, rather than the true distribution. 

\subsection{Weighting data points to conform to true distribution expectations}
\label{sec:reweight/raking}

In scenarios with human-interpretable features (e.g., the time of day an image was taken), domain knowledge can be elicited from experts (Section~\ref{knowledge/true}) to better understand how the empirical distribution in the data differs from that of the true distribution with respect to certain feature values (e.g., night images make up 20\% of the data, but in the deployed environment, images are expected to be 50/50 night/day). With this information, model-agnostic weights can be found to compensate for the discrepancy between the data and the deployed environment by weighting samples in accordance with the prevalence of their feature values in the true distribution (e.g., a night image might be weighted at $\frac{0.5}{0.2}=2.5$, and a day image at $\frac{0.5}{0.8}=0.625$). While this is trivial for single features, there are statistical techniques for finding data point weights given expectations on true distribution (or population-level) feature values.

One technique is iterative proportional fitting, also known as matrix scaling or \emph{raking}, in which the sample weights on specified features are adjusted to have marginal totals equal to the corresponding population-level feature totals \cite{deming1940least, deville1992calibration, rote2007matrix}. Fundamentally, raking seeks to weight data based on partial auxiliary information about the true distribution \cite{albertus2019auxiliary}.

Consider the following scenario:
\begin{itemize}
    \item In the deployed environment, (a) images are equally as likely to be taken during the day as during the night (i.e., 50\% night images), and (b) images are twice as likely to contain military ships than commercial ships (i.e., 66\% military images).
    \item In the development data, (a) images are four times as likely to have been taken during the day (i.e., 20\% night images), and (b) images are equally likely to contain military ships as commercial ships (i.e., 50\% military images).
\end{itemize}
In this scenario, the raking procedure fits weights to the data such that the weighted sum of the night images is approximately 50\% of the weighted total (i.e., night images have larger weights than day images, since they appear less frequently in the data than in the deployed environment) and the weighted sum of the military ship images is approximately 66\% of the weighted total (i.e., military ship images have larger weights than commercial ship images). Since the procedure focuses on feature totals,\footnote{The raking procedure can support any number of feature groupings given that the joint probability in the deployed environment is known and that there are sufficient data points containing all features in the group. This section focuses on single features for clarity (since feature groups could be conceptually understood as single features) and because the probability of single features in the deployed environment (e.g., the likelihood of night images, the likelihood of military ship images) is usually more readily available than joint distributions (e.g., the probability of images of military ships at night).} these weight adjustments can be at cross purposes: samples showing commercial ships at night will require both smaller weights (to correct for the military ship total) and larger weights (to correct for the night total).

The raking procedure addresses this tension by iteratively adjusting weights for each specified feature in turn, following the analogy of smoothing soil with a rake alternately in perpendicular directions \cite{battaglia2009practical}. In the two-feature scenario, raking would weight samples by multiplying each by the ratio of the deployed environment total to the weighted sample total for that feature such that the adjusted total would agree with the deployed environment total (e.g., the weighted sum of the night images would be 50\% of the total). As this may have altered the total for the other feature, raking next weights samples by multiplying each by the appropriate ratio such that the adjusted total agrees with the deployed environment (e.g., the weighted sum of military ship images would be 66\% of the total). This process continues until (a) a maximum number of iterations is reached or (b) the marginal totals reach a specified tolerance (i.e., close enough to the deployed environment totals) \cite{battaglia2009practical}. 

While raking can be applied to any data for which there are known true distribution totals (e.g., feature X should take value Y at rate Z in the deployed environment), there are practical considerations to its use. In the statistical literature, raking is often performed on data points with relatively few categorical features with relatively few possible options (e.g., discretized gender, binned age), which helps the process converge after a small number of iterations \cite{battaglia2009practical, valliant2013practical}. While the more general matrix scaling formulation supports real-valued features \cite{rote2007matrix, pukelsheim2014biproportional}, it may be impractical to elicit this kind of information about the deployed environment without binning or similar approaches to approximate the true distribution of features. There are also specific scenarios in which raking may be problematic, including (a) when an attribute is encoded as part or by proxy in more than one specified feature (e.g., the time an image was taken -- day/night -- and the time an image was taken -- binned by hour), (b) when there are a large number of specified features, and (c) when specified features depend on one another (e.g., the time an image was taken -- day/night -- and the median pixel brightness value) \cite{brick2003identifying}.

Representative sample weighting is closely related to iterative proportional fitting but frames the problem as entropy maximization. Section~\ref{sec:optimal_rep_sample_weighting} reviews the \texttt{rsw} Python package for representative sample selection. To implement the package for representative sample \emph{weighting}, the same steps are taken, but a non-Boolean regularizer should be chosen.

Methods for leveraging auxiliary information about the true distribution can help correct for discrepancies between development data and the deployed environment, but such approaches should be taken with care to ensure that the direct outcome of the process -- weights for samples -- remains useful for the overall goal of the process: that the resulting AI-enabled system is optimized for performance in the deployed environment.

\subsection{Importance weights for covariate shift}
\label{sec:reweight/importance}

Covariate shift is a form of distribution shift in which the distribution of features shifts but the conditional distribution of labels given the features, which we can think of as a labeling scheme, does not. In other words, if a training data point and an input from the true distribution had identical features, then they would have identically distributed labels. Intuitively, the scheme used to label the training data is the same scheme that should be used to judge inputs that the model encounters in the deployed environment.\footnote{Covariate shift has been studied widely, but it is only one of several types of distribution shift for which mathematical techniques have been introduced. See Chapter 1 in  Qui\~nonero-Candela et al. \cite{Quinonero-CandelaJoaquin2008DSiM} for a more comprehensive list. A recent survey, released as a preprint in March 2024, catalogs the importance-weighting techniques currently available for addressing each of these kinds of distribution shift \cite{kimura2024short}.}

A foundational method for addressing covariate shift is importance-weighted empirical risk minimization (IW-ERM), which introduces importance weights for each data point into the loss function used to train the model. An importance weight, in the original formulation, is the ratio of the probability densities for a data point in the training and true distributions. If the feature space were discrete, the importance weight of a data point with features $x$ would be the probability of the model encountering input $x$ in the deployed environment, divided by the prevalence of $x$ in the training data. For continuous feature spaces, the probabilities are replaced by (finite-sample estimates of) the probability densities. These weights are agnostic to the model because they are chosen on the basis of the training and true distributions. The theory of IW-ERM is described in chapter 6 of Qui\~nonero-Candela et al. \cite{Quinonero-CandelaJoaquin2008DSiM}, and more numerically stable variants are mentioned in Kimura and Hino \cite{kimura2024short}.

IW-ERM and its variants can be used for any ML model that is trained through empirical risk minimization. However, both empirical evidence \cite{Byrd2018WhatIT} and theory \cite{Xu2021UnderstandingTR} suggest that IW-ERM does not have much effect on deep learning models once they have been trained for a long duration. These results were shown using a loss function with exponential tails; using a polynomial-tailed loss function may recover the utility of importance weighting for deep learning models \cite{Wang2021IsIW}.

\subsection{Weights to overcome biased labeling}
\label{reweight/labelbias}

Covariate shift is not the only way to theorize the difference between the available training data and the true distribution in the deployed environment; systematic bias in training data points' labels is another way to frame the discrepancy. In contrast to covariate shift, biased labeling assumes that the scheme for labeling training data differs from the scheme for correctly labeling data in the true distribution. Under covariate shift, two inputs with identical features, one in the training data and the other in the deployed environment, would be equally likely to bear each label; under the paradigm of biased labeling, they would not. 

The training data labels are assumed to be biased insofar as they violate functional constraints that encode notions of group fairness, such as equal opportunity. Jiang and Nachum's approach \cite{jiang_identifying_2019} learns a parametric characterization of the bias in training data labels and uses it to develop a weighting function for the training data points. Under certain conditions, training on the weighted data points and their biased labels is equivalent to having trained on the unweighted data points with unbiased labels, were they to have been available.

\section{Compute weights specified to the model architecture}
\label{sec:reweight_model_spec_pre}

If the model's architecture has been determined, the weights of training data points can be chosen to reflect individual data points' utility. Training data points receive a higher weight if their inclusion in the training set improves, or is thought to improve, the trained model's performance on the validation set. Data valuation techniques aim to learn such weights from preliminary model performance. Then, once weights have been computed, the model is trained from scratch with weighted data.

Many data valuation methods, including leave-one-out and cooperative game-theory-based approaches, train models with the same architecture on various subsets of the data and evaluate each of these preliminary models' performance on the validation set \cite{sim_data_2022}. Weights derived in this way quantify the utility of each data point to a model's performance on the validation set and, by extension, the true distribution. If training many models is infeasible or impractical, uncertainty quantification can be used to produce weights instead. Uncertainty quantification does not depend on the validation set, so these weights are not as directly related to trustworthiness as defined in this report. We include one uncertainty-based method in our discussion because it incorporates concern for a dimension of the true distribution.

\subsection{Weights based on initial model performance on the validation set}
\label{sec:reweight/initial_perf}

The most direct way to quantify the relative utility of different data points to the trustworthiness of the trained model is to vary the training data of an initial model and observe how its performance on the validation set varies in response. Leave-one-out approaches are straightforward ways to compute weights, but they can be brittle to seemingly irrelevant changes in the dataset overall. Instead, many modern approaches draw on Shapley values, originally proposed in cooperative game theory \cite{sim_data_2022}.

A leave-one-out approach computes the weights of data points by training an initial model with all the data points and then training initial models omitting each data point in turn. All initial models are evaluated by their performance on the validation set, which is an estimate for their performance on the true distribution. A data point is up-weighted if model performance degrades significantly when it (alone) is omitted from the training set. Performance might degrade because the data point represents an underrepresented (in the data) part of the true distribution or because inclusion of the data point ameliorates a weakness in the model's architecture for the task at hand. Either way, up-weighting valuable data points can help optimize the eventual model's performance on the true distribution, improving its trustworthiness.

Leave-one-out approaches have some weaknesses, though, as illustrated by the case of duplicate data. If one data point is an exact duplicate of another data point in the set, then omitting it (alone) from the training set likely results in little or no performance drop-off. Whatever the model was going to learn from the features and labels of the data point, it already learned from the original copy. In this case, the duplicate data point might be severely down-weighted when training the model. A leave-one-out approach would come to the same conclusion about the original data point; leaving it out does not degrade performance much because the copy is still present. While the de-duplication of data might not degrade performance much, severely down-weighting both copies of the data point is a mistake. Just because a data point is duplicated does not mean all copies of it are useless.

Data Shapley values \cite{ghorbani_data_2019} more holistically capture the value each data point offers. A Shapley value measures the drop in performance that results when a given data point is omitted from any training subset selected from the overall pool of data. For instance, in the case of duplicate data, the data Shapley considers the marginal contribution of a copy to both a set of training data containing the original (likely a small or negligible contribution) and a set of training data without it (possibly a much larger contribution). By considering a data point's contribution in the context of various training subsets, the Shapley value avoids a pitfall of leave-one-out approaches.

The original data Shapley \cite{ghorbani_data_2019} value was defined in reference to the overall training dataset, but it was reformulated and generalized into distributional Shapley \cite{ghorbani_distributional_2020}. The distributional Shapley value views the dataset as a sample from some distribution, which we assume to be different from the true distribution, and it captures the marginal contribution of a data point in the context of the whole distribution.

For large datasets and complex model architectures, leave-one-out approaches are computationally infeasible because they require training as many initial models as there are data points. Shapley values are worse yet because they require training exponentially many (in the number of data points) initial models. For Shapley values, researchers have developed approximation techniques that require training fewer models \cite{sim_data_2022}. Because distributions often have infinite possible data points, distributional Shapley values are almost always approximated, and approximating them can be more efficient than data Shapley values \cite{ghorbani_distributional_2020}.

\subsection{Uncertainty quantification in an initial model}

There are methods for quantifying uncertainty in many models; a taxonomy is provided by Canal et al. \cite{canal2024decision}. Uncertainty weights can be computed by training an initial model on the whole unweighted training dataset, then feeding each training data point to the initial model and quantifying the model's uncertainty on each point. Intuitively, the initial model may have high uncertainty on its own training data in regions of the training distribution that exhibit patterns the model is not well suited to learn. This scheme, however, does not necessarily optimize performance on the true distribution of inputs, and so may require preliminary steps to align the training dataset to the true distribution of inputs. However, Pombal et al. \cite{pombal_fairness-aware_2023} introduce a weighting technique that uses uncertainty quantification to both improve model performance on the training data and achieve fair performance across a sensitive attribute.

Pombal et al. suggest training one initial model to predict the labels and another initial model to predict sensitive attribute, then measuring the Shannon entropy for each model on each training data point. The entropy values are averaged together (arithmetically or geometrically) to form a weight value. Training data points receive a higher weight if the initial label-predicting model has more uncertainty and also if the initial protected attribute-predicting model has more uncertainty \cite{pombal_fairness-aware_2023}. The rationale is that a higher weight will sensitize the ultimate model to training data points where the model is likely to struggle and also where the protected attribute is less salient. It is unclear from the paper whether the performance attained is due to the strength of the approach or to the means of evaluation. The evaluation scheme seems to reward scattered results rather than high average performance. However, this approach is appealing for its simplicity because it tries to weight the training data while balancing the considerations of model utility and fairness.

\section{Weight data dynamically during training}
\label{sec:reweight_dynamic}
Until now, the weighting techniques discussed have resulted in a static list of weights, with a single weight per data point.
A static list of weights is useful because it is easy to interpret the weighting scheme's impact on a training procedure given your model and classifier.
However, another approach to data weighting is to learn weights dynamically at train time.
Under this paradigm, an algorithm is used to determine the relative weights of each data point based on the current state of a model.
As the model changes through training, so too will the data weights, making them dynamic.
The core intuition behind this approach is that a model's bias toward features of the training distribution can change over time and a weighting scheme should be able to adapt to those changes.
Consider a multiclass classification problem with two minority classes.
It is conceivable that initially the two minority classes will be up-weighted to compensate for their size.
However, during training with these weights, the model may begin to prioritize one of the minority classes because of the ease of classification.
In this case, it would be beneficial for the weights to be updated such that the more difficult minority class can be even further emphasized.
The following methods are designed to handle such a scenario.

Dynamic data weighting methods fall broadly into two categories: those that derive weights to optimize the original performance metric and those that introduce a new performance metric to fit weights to.
While the former is a simpler approach, the latter enables a data scientist to bake multiple notions of performance into the model. 

We will describe these methods in the following subsections, but before doing so, it should be noted that there are drawbacks to the dynamic data weighting approach.
Most notably, dynamic data weighting is less interpretable because it is not straightforward to determine the relative contribution of each training data point to the model trained.
While static weights of data points can be directly compared, dynamic data weighting approaches produce an entire time series of weights for each data point, which are more difficult to compare.

\subsection{Weighting based on original performance metric}
One view of dynamic weighting is as a means to better meet the initial performance metric.
Under this view, there is a single notion of performance, quantified through a single performance metric.
See Section~\ref{sec:performance} for further discussion on the choice of a performance metric.

Methods in this category add a data weighting module within the module, which assigns weights to training points as they are seen.
The weights are then used in conjunction with the performance metric.
One such approach is data valuation using reinforcement learning \cite{yoon_data_2020}.
This approaches defines a reinforcement learning sub-model, which learns to weight data points as the overall prediction model is trained.
This is done by feeding these weights to a multinomial sampler, and then feeding the sampled data points to the prediction model.
As the whole model is trained, the reinforcement learning component learns to highly weight (and thereby sample) the most effective data points for optimizing the performance metric.
Another way of viewing this process is that the model assigns weights to the data points and then stochastically binarizes them.
Thus, this approach falls at the intersection of dynamic data weighting and data sampling.

\subsection{Weighting based on additional performance metrics}
More powerful dynamic data weighting methods introduce a secondary performance metric that is jointly optimized with the primary performance metric.
The introduction of this additional performance metric allows the practitioner to train models that meet necessary standards for multiple notions of performance.
Additional objectives may include class-based performance, performance on a ``clean dataset" (in our language, a validation dataset that is representative of the true distribution), or alternative definitions of performance.
The intuition for using these approaches is that given the current state of a model, some data points may be more important to the training objective, and the magnitude of that importance can be derived from a second performance metric.

For example, adaptive sensitive reweighting is a classification method in this family of approaches that uses average performance over the classes as a secondary performance metric \cite{krasanakis_adaptive_2018}.
During training steps, the per-class performance metrics are calculated, and those metrics directly inform the weights of the data points in the following training step.
That is, if the model is performing poorly on a class, data points from that class will be weighted as more important in subsequent training steps.
Thus, this secondary objective encourages equal performance across classes, while the primary objective encourages strong performance overall.
This example also highlights the need for dynamic weighting.
We can imagine that a model performs poorly on one class of data in the early training steps.
As that class is emphasized through weighting, it is possible that the model then begins to perform worse on another class.
By dynamically weighting the data, the weights can adapt to improve the model throughout the whole training procedure.

Chai and Wang \cite{chai2022fairness} take a similar approach, which optimizes for some more recent fairness-based metrics.
Other efforts like FORML and DVRL have explored more complex weighting procedures, effectively adding a data weighting sub-model \cite{yan2022forml,yoon_data_2020}.

In a similar effort, Roy et al. proposed ``multi-fair boosting post Pareto," which actually optimizes three performance metrics to balance the three-way trade-off among accurate, class-balanced, and fair results \cite{roy_multi-fairness_2022}.
Specifically, these objectives are accuracy, average class-based accuracy, and discrimination on protected attributes.
The data is dynamically weighted in support of this optimization through the boosting algorithm.

Ren et al. \cite{ren_learning_2019} provide an alternative approach when considering deep learning models.
Instead of using heuristics to optimize models on the true distribution, they suggest optimizing on a small set of data pulled from the true distribution itself.
Their assumption is that the practitioner can obtain a small, clean dataset that is representative of the true distribution.
Then, the secondary performance metric is performance on this validation dataset.
Specifically, data points are weighted according to their gradient's alignment with that of the performance metric evaluated on the validation dataset.
Thus, data points that can nudge the model toward better performance in the deployed environment become more important during training.
In a sense, the underlying approach of altering the update procedure to make the model more robust is similar to that taken in model-agnostic meta-learning, which has the slightly different goal of learning a model that can quickly be trained for a new task \cite{finn2017model}.

\section{Choose between resampling and weighting}
\label{learn_sample}
In many ways, data resampling and data weighting are alternative approaches to solving the same problem: certain data points may be more or less important, and some types of data points may be over- or underrepresented.
Some techniques explicitly link the two approaches, by learning weights that inform data sampling \cite{yoon_data_2020}.
The weighting procedure may be as complex as a reinforcement learning algorithm or as simple as a uniform class-based sampling.
A simple example is as follows: Under a two-class scenario, a weighting algorithm may determine that class one should be weighted $0.75$ and class two should be weighted $0.25$.
This method might also assume a binomial distribution.
Then, on further training, class one would be sampled with probability $p=0.75$, and class two sampled with probability $p=0.25$, increasing the number of class one examples that the model will see.
This approach is an alternative use of data weights and is contrasted with using the weights to augment the model optimization procedure directly.

Since sampling and weighting can be viewed as alternative approaches, some studies have attempted head-to-head comparisons of their effectiveness for imbalanced learning.
An et al. \cite{an2020resampling} focused on weak classifiers but found that sampling outperforms data weighting.
Seiffert et al. \cite{seiffert2008resampling} also compared the approaches within the context of boosting algorithm implementations (so weighting or sampling as part of the ensemble construction, not as a single pretraining step), and that resampling also outperforms weighting in this context.

However, it is less clear that sampling is even necessary if more powerful models are available.
Elor and Averbuch-Elor \cite{elor2022smote} posit that the advantages of resampling, and specifically SMOTE, may be smaller than those of just using a more expressive classifier like XGBoost \cite{chen2016xgboost}. 
They test this hypothesis by comparing the results of class balancing and decision boundary optimization over a grid of imbalanced datasets, models, and their hyperparameters.
They find that there is no need to balance data through resampling when using a strong classifier and optimizing the decision boundary for performance.
However, they confirmed that balancing data provides benefits when only weak classifiers are available. 
They further compared random rebalancing to SMOTE and SMOTE variants and found that SMOTE-based approaches were no better on average than random, unfair oversampling techniques.

Overall, it seems that the class of models available for the task at hand will be a major factor in deciding which data valuation techniques to apply.
If only weak classifiers are available, which may be the case for regulatory, resource, or interpretability reasons, these data valuation techniques may be more useful.
However, if stronger classifiers are available, quantitative evaluations should be done to confirm that these data valuation techniques are actually needed for the problem at hand.

\section{Practitioner's perspective on weighting with rsw}
\label{subsec:practioners_perspective_rsw_weighting}

Representative sample weighting is closely related to iterative proportional fitting but frames the problem as entropy maximization. Section~\ref{subsec:practioners_perspective_rsw_selection} reviews the rsw Python package for representative sample selection on a corpus of PubMed documents. To implement the package for representative sample weighting, the same steps are taken, but a non-Boolean regularizer should be chosen.

In that PubMed example, 800 samples were selected from the 4,000 training set articles to serve as the validation set. The validation set was chosen to match the proportion of randomized controlled trials and the proportion of articles benefiting from U.S. grant money in the deployed environment (articles added to PubMed in 2023). The remaining 3,200 articles in the training set -- while collectively less similar to the deployed environment than the 4,000 original training articles -- can now be weighted using rsw to balance on the chosen selected characteristics.

\begin{table}[ht]
\begin{center}
\begin{tabular}{ |p{4.25cm}||p{2.5cm}|p{4cm}|p{4cm}| }
 \hline
 \multicolumn{4}{|c|}{Observed Values of Selected Characteristics} \\
 \hline
  & \textbf{Deployed\newline Environment} & \textbf{Training\newline Samples\newline Not Selected\newline for Validation} & \textbf{Weighted Training\newline Samples\newline Not Selected\newline for Validation}\\
    \hline
    \% RCT & 2 & 2 & 2 \\
    \% Receiving U.S. Money & 23 & 15 & 16 \\
 \hline
\end{tabular}
\caption{After weighting the 3,200 training articles that remained after removing the validation set, the distribution of selected characteristics (whether the research concerns randomized control trials (RCT) or received US grant money) is slightly closer to that of the deployed environment. The target percentage of randomized controlled trials is maintained, and the percentage receiving U.S. grant money is elevated.}
\label{tbl:rsw_weighting_results}
\end{center}
\end{table}

Table~\ref{tbl:rsw_weighting_results} illustrates both the utility of rsw and the challenge of the problem. The remaining 3,200 articles are more similar to the deployed environment when weighted. However, the percentage of articles that benefited from U.S. grant money is much more favorable in the 800 articles chosen for validation (and omitted from representative sample weighting). When using representative sample selection to choose a validation set, the observations that best approximate the deployed environment according to the selected characteristics are chosen. It will naturally be the case that the observations not selected for validation are, collectively, less similar to the deployed environment. Representative sample weighting can compensate to an extent.

\part{Further Topics}

\chapter{Describe and validate data with semantic types}
\label{sec:semantic_types}
In many real-world scenarios, the data available for development is not well aligned to the true distribution of inputs. This may occur in scenarios where development data encodes aspects differently from how SMEs understand or communicate them (e.g., columns have different names or describe different phenomena from how SMEs describe the deployed environment) or the development data contains multiple constituents (e.g., decision 1c and 1d in Section~\ref{sec:statistical_tools/decision_1}, the union of several benchmark datasets). To apply the techniques and strategies described in this report, from splitting and selection to detecting mischaracterization, it is essential that the data scientist understand what each encoded aspect means, how those encoded aspects relate to the true distribution, and how to unify potentially disparate sources of information. 

\section{Recognize semantic types}
\label{sec:semantic_types/recognize}
Encoded aspects can often be understood as instantiating \emph{semantic types}, which describe the kinds of entities represented in the data. Semantic types may be (human-interpretable) feature identifiers (e.g., column headers), or -- in the context of the semantic web -- correspond to classes in an ontology. For example, a feature might correspond to the semantic type “Date" and may be expressed by a standardized datatype like ISO 8601 \cite{iso_8601}. Lists of semantic types and their association with a corresponding set of datatypes and constraint rules are frequently used for validation, where a datatype consists of (a) a set of distinct values, called its “value space,” (b) a set of lexical representations, called its “lexical space,” and (c) a set of facets that characterize properties of the value space, individual values, or lexical items.

Example semantic types are shown with corresponding datatypes in Table~\ref{tbl:smpl_semantic_types}.

\begin{table}[ht]
\begin{tabular}{ |p{3cm}||p{3cm}|p{4cm}|p{4cm}|  }
 \hline
 \multicolumn{4}{|c|}{Sample Semantic Types} \\
 \hline
 \textbf{Semantic Types} & \textbf{Datatype} & \textbf{Constraint} & \textbf{Example}\\
 \hline
 Email Address & String & Must contain “@" and a domain name & jane.doe@domain.com \\
  \hline
 URL&   String  & Must follow URL format, to include scheme, domain, and path   &https://www.sample.gov\\
  \hline
 Phone Number & String & Must follow a valid phone number format, relative to country & +0-800-666-6666\\
  \hline
 Date of Birth & Date & Must be a date on or before today & 06/01/2024\\
  \hline
 Temperature&   Float or integer & May include units such as Celsius or Fahrenheit in valid ranges&-273.15 to 56.7 F\\
  \hline
 \raggedright{Social Security Number} & String  & Integers in the form xxx-xx-xxxx & 123-45-6789\\
  \hline
 Zip Code & String  & Must follow a valid postal code format & 55555\\
  \hline
 ISBN & String & Must follow the ISBN-10 or ISBN-13 formats & 978-1-83-802958-6\\
  \hline
 IP Address & String & Must follow the format for IPv4 or IPv6 addresses & 128.254.12.4\\
  \hline
 Geographic Coordinates & Pair of floats & Must represent valid latitude and longitude values & 37.2431° N, 115.7930° W\\
  \hline
 Pixel Coordinates & Pair of integers & Values must be positive in valid ranges & 1024, 768\\
     \hline
    Person Name & String & May include first, middle, and last name & Jane Doe\\
     \hline
    File Path & String & Must follow syntax form relative an operating system & \texttt{C:/username/}\\ 
    \hline
    Person Age & Integer  & Usually a number between 0 and 122 & 42\\
 \hline
\end{tabular}
\caption{Sample semantic types shown with typical datatypes and constraints.}
\label{tbl:smpl_semantic_types}
\end{table}

Many common commercial software packages use regular expression and dictionary lookups to automate the detection of semantic types, leveraging lexical matches to link data to a set of semantic types with corresponding datatypes. Such dictionaries may recognize that terms such as “temporal region,” “duration,” and “time period” are all associated with the semantic type “Time Span,” and that Time Span may be implemented in a number of different formats conforming to the datatypes “date” or “dateTime.” Chevallier et al. \cite{chevallier_semantic_2023} identify a number of examples of commercial products using this approach. Some of these products may recognize a semantic type based solely on lexical matches to column headers, while others may detect semantic type by evaluating the datatype and format of the associated data. Google’s Looker Studio, for instance, allows for automatic semantic type detection based on detection of associated datatype property values \cite{noauthor_data_nodate}. 

In addition to commercial tools, many open-source tools leverage type registries for this purpose for either form of validation \cite{begtin_semantic_2022}. Yan and He \cite{yan_synthesizing_2018} report development of a system that allows users to provide a set of positive examples for a target datatype and a search word, and their system automatically identifies relevant code and synthesizes type-detection functions using execution traces. 

Generative AI is also greatly impacting the discovery of semantic types in a data workflow. Korini and Bizer \cite{korini_column_2023} report that ChatGPT is able to competitively align columns in relational tables to semantic types with no or minimal task-specific demonstrations. Li et al. \cite{li_table-gpt_2024} report that fine-tuning LLMs, such as GPT-3.5 and ChatGPT, on tasks related to tabular-formed data improved the models' performance on held-out tasks such as data transformation and data cleaning.  

Common methods of identifying the semantic type for data elements include identification of lexical matches between the schema name and the corresponding semantic type, identification of data values corresponding to datatypes associated with a semantic type, manual assertion, tailored ML approaches, or some combination. In all cases, a review phase involving a SME is required because misalignment can easily lead to misinterpretation. In particular, none of the automation approaches listed here can account for decisions that may have occurred in collecting data. For instance, if “adult” is a semantic type in both datasets, one should ask if the criterion used for asserting that a person is an adult was the same in both cases (e.g., was the legal age of 18 used, or were survey respondents asked if they were adults?).

In the context of computer vision, discrepancies in how a feature is conceptualized in the development data as compared to the deployed environment can be particularly impactful on model performance. Discrepancies can occur because those acquiring the data had different ways of conceiving of what appears to be a common semantic type; for instance, “military vehicle” may have meant “owned by the military” in one dataset and “designed to serve a military function” in another dataset. 

\section{Leverage semantic types for data cleaning and more}

Identifying features in both the training and validation datasets with their corresponding semantic type enables many distinct capabilities:

\begin{enumerate}
\item	Data mining – Semantic types are used to determine the most relevant patterns to extract from data.
\item	Data cleaning – Semantic types are used alongside rules to facilitate tasks such as data validation (e.g., ensuring data accords with expected datatypes and flagging violations for review) and transformation (e.g., converting data formats to known datatypes associated with a known semantic type).
\item	Data interoperability – Semantic types are used to ensure that data schemas expected by software systems can be adequately supported by the semantic types, for instance, through conformance to a REST application programming interface (API) standard. 
\item	Data enrichment – Semantic types are used to ensure that the data schemas one wishes to integrate are appropriately matched, such that portions of datasets (e.g., individual columns in two tabular sets of data) may be merged. This technique can also be used in the context of SME review, where matched semantic types can be paired with a definition for verification by a SME. 
\end{enumerate}

Using Internationalized Resource Identifiers (IRIs) is also helpful in documenting mappings. Where natural language can be ambiguous, IRIs can provide an exact identifier for an entity represented in data. Such entities may be kinds of things (e.g., ship), instances of such kinds (e.g., the Titanic), or relations that hold between instances (e.g., X captain\_of Y). The IRI standard builds on the Uniform Resource Identifier (URI) standard by expanding its list of permissible characters. 

One way to understand a workflow using IRIs is to compare them with using a traditional dictionary, where the text is indexed according to natural language words in alphabetical order. Each entry in a dictionary describes a single word and provides multiple definitions for this word based on different senses of the word. Since different senses can be referred to by different words, a thesaurus can also be used to see synonyms for words that may refer to one or more of the same meanings as the term in the dictionary. The practice of using IRIs accords with assigning an IRI to each definition in the dictionary and not to each word. Multiple words may then be associated with each IRI to facilitate querying a set of IRIs and selecting the one that accords with an end user's intended meaning. 

For instance, a system may associate a moderate degree of confidence that “birth” in a column header refers to the event of a person’s birth, and this confidence may be raised when a string appears to be one of a recognized set of data formats in the corresponding values under the column. On this basis, a system designed to automate semantic type assertions for datasets may assign a candidate class IRI to the column header and normalize the appearance of dates in the values beneath it while also asserting that each value is an xsd:dateTime. 

\section{Verify semantically typed data with queries}
\label{sec:semantic_types/verify}

Once semantic types have been established and data has been cleaned and validated, the features and distributions may be reliably compared across datasets. This comparison may be desirable between training and validation datasets, between development data and the deployed environment, or when merging multiple development datasets (Chapter~\ref{statistical_tools}).

Executing these tasks can be supported by multiple frameworks. TensorFlow includes an evaluator component, which allows a model to be validated against a serving model to determine whether it is good enough relative to baseline.\footnote{\url{https://www.tensorflow.org/tfx/guide/evaluator}} However, there can be advantages in working with semantic web standards outside a tool-specific framework, and if semantic types are already represented with IRIs, then it may be useful to represent data in the Resource Description Framework (RDF).

RDF is a widely used W3C knowledge representation language that allows data to be represented as a graph. Tabular data is often transformed to create RDF, and this can happen in a number of different ways. In addition, Python libraries like RDFlib \cite{noauthor_rdflib_nodate} and a W3C standard mapping language, R2RML, may be used to transform data.\footnote{\url{https://www.w3.org/TR/r2rml/}} RDF is the basic language of the semantic web and is enabled using semantic web techniques for tasks such as data enrichment. In this case, definitions, labels, source information, and other kinds of terminological support can be automatically associated with the IRIs in a data schema, as they may be easily queried for in standard sources such as Wikidata, DBpedia, schema.org, or other well-known or authoritative knowledge graphs and ontologies.  RDF can also be saved in many different syntaxes, including JSON-LD.

\begin{figure}[h]
    \centering
    \vspace{3pt}
    \includegraphics[width=0.6\textwidth]{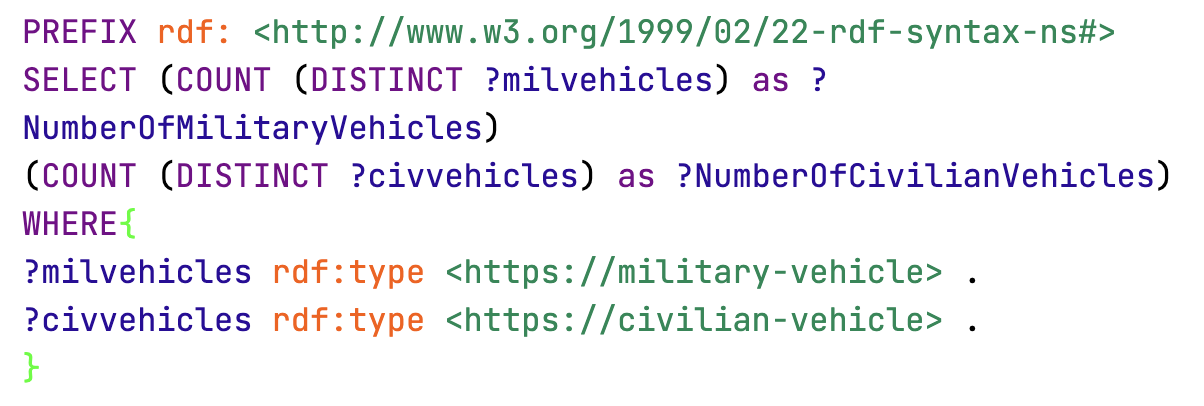}
    \caption{Example SPARQL query for the total count of military vehicles and the total count of civilian vehicles.}
    \label{fig:count-query}
\end{figure}

Consider a dataset with 19 military vehicles and 20 civilian vehicles. Representing these facts in RDF involves creating IRIs for each vehicle, asserting that they each associated as types of the classes military vehicle and civilian vehicle, respectively (where these classes are also represented by IRIs). Once this transformation has taken place, this can be confirmed with a SPARQL query, shown in Figure~\ref{fig:count-query}.

The query in Figure~\ref{fig:count-query} computes the total count of all distinct instances of civilian vehicles and military vehicles. This query returns that there are 19 military vehicles and 20 civilian vehicles, confirming that our data was correctly represented in the RDF transformation of our dataset. A validation query such as this should be run prior to performing analyses to ensure that the transformation of data into RDF was executed reliably and that the results faithfully reflect the original data source. 

\begin{figure}[h]
    \centering
    \vspace{3pt}
    \includegraphics[width=0.6\textwidth]{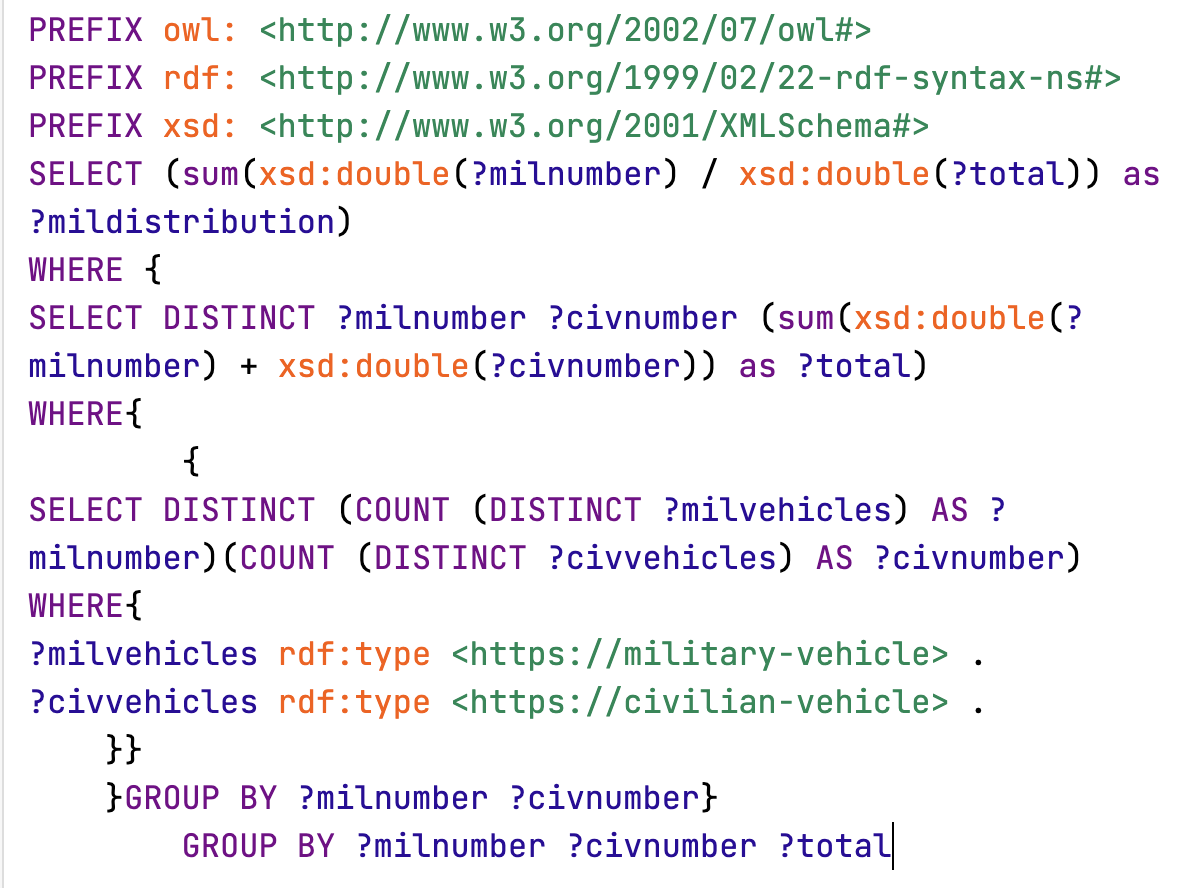}
    \caption{Example SPARQL query for the percentage of military vehicles.}
    \label{fig:distro-query}
\end{figure}

Figure~\ref{fig:distro-query} depicts a query that returns the distribution of military vehicles over the total number of vehicles, with the result: \texttt{0.48717948717948717\textasciicircum\textasciicircum xsd:double}.

Once this process is repeated for the features in the validation dataset, these distributions may be queried and compared to assess the expected reliability of our model.

\section{Capture data provenance alongside semantic types}

Data provenance refers to the record of the history of a dataset, including elements such as its date and method of creation, source information, and author. We have noted that different ways of conceptualizing a feature may influence model performance in the deployed environment. However, differences can also occur based on where the data itself is acquired. For instance, two datasets may contain data conforming to the semantic type “truck,” but if the data for the validation set data was acquired from a different country than the data in the training dataset, then trucks may look quite different based on the distribution of the makes and models specific to a geographic market. 

Data provenance standardization is often seen as low-hanging fruit for many research groups looking for ways to make nonsymbolic AI, if not explainable, at least traceable \cite{longpre2024dataauthenticityconsent}. Within industry, the rush to adopt provenance standards has also been driven by interest in generative artificial intelligence and concerns regarding both the risks of generative AI as well as the potential overreach of government regulation \cite{noauthor_prov-o_nodate}. 

For this reason, there has been a proliferation of overlapping and competing initiatives to establish provenance standards, including data cards and metadata tagging standards. Prominent provenance initiatives include the Massachusetts Institute of Technology (MIT) Data Provenance for AI project, which audits over a thousand AI datasets \cite{noauthor_project_nodate}, and the Data \& Trust Alliance data provenance standards, which represent a broad set of partnerships with industry to provide a series of provenance standards for datasets \cite{noauthor_data&trust_alliance_nodate}. The Coalition for Content Provenance and Authenticity (C2PA) is another high-profile initiative that seeks to support both provenance and content credentials to combat the threat of disinformation due to the rise of synthetic media \cite{noauthor_c2pa_nodate}.    

\urldef{\urlOSF}\url{https://help.osf.io/article/217-how-to-make-a-data-dictionary#:~:text=A\%20data\%20dictionary\%20is\%20critical,in\%20your\%20spreadsheet\%20really\%20mean.} 
In this section, we have surveyed how semantic typing can aid in a number of data curation tasks that facilitate the automatic confirmation that our models are trustworthy. For those already executing these tasks using IRIs and RDF,  these approaches may also be extended to serve a further benefit: recording the provenance of the dataset. This is true regardless of the provenance standard adopted within a workflow.\footnote{This applies broadly to data dictionaries as well. As an example, see Open Science Framework (OSF) data dictionary guidance: \urlOSF Creating a data dictionary should be a part of the data curation process and not a separately created article.}

Within the semantic web, representing data provenance has a long history. The Provenance Ontology (PROV-O) is only one prominent example of a W3C standard ontology created specifically to track the provenance of datasets and other informational entities \cite{noauthor_prov-o_nodate}. Another is the Data Catalog Vocabulary (DCAT), which facilitates the representation of data objects, such as datasets, for curation within a data catalog to aid in their discoverability \cite{archer2014catalog}. Having identified the semantic types within a dataset for tasks such as data cleaning, validation, and enrichment, the labels for these semantic types can now serve as keywords in a DCAT-conformant representation to aid in the discoverability of the dataset by others.

\chapter{Record important information in an accompanying information system}
\label{sec:information_system}
Following the actionable definition of trustworthiness, the primary goal of data curation is to train and select models as part of an AI-enabled system. In practice, however, the development of trustworthy AI-enabled systems benefits from thorough documentation regarding data, models, and their contributions to the overall system. This parallel process can facilitate trustworthiness by recording decision criteria and rationale (helping stakeholders to trust the process) as well as domain knowledge (ensuring future curation decisions benefit from previously elicited SME insights). 

Important information may include metadata (e.g., creation date, column headers for tabular data, file names) and model performance (typically first recorded in logs). Some important information may be compiled in data, model, or system cards: artifacts that aim to make the information about the dataset, trained models, or AI-enabled systems available to downstream or end users. However, there is no consensus on what information should and can feasibly be expected as part of a data, model, or system card. 

Part of the confusion around what to include in a data card, for instance, arises from the ambiguity about what information a dataset creator can be expected to know, especially regarding downstream uses of a dataset. A data scientist creating a dataset in isolation can only speculate about the ways a subsequent data scientist might curate the dataset, the models a developer might train with the dataset, and the AI-enabled systems that might incorporate the dataset or a derivative thereof. Many data card templates call for dataset creators to advise data card readers about appropriate and inappropriate uses of the dataset. We contend that it is unreasonable to expect a data card creator to foresee all possible uses of a dataset and unrealistic for subsequent users of a dataset to be restricted to using it only in ways the creator foresaw.

To complement data, model, and system cards, which aim to document components individually, we envision an information system\footnote{Here, an ``information system'' refers to a computer program that enables users to input, store, and retrieve text and numbers. The term ``information system'' abstracts the functions of such a program away from its implementation, which could be, for example, a relational database or a graph database equipped with a query engine. In other contexts, the term ``information system'' can refer to a computer network; our usage is different.} that unites details about the data, the data's curation, the model, the model's performance, and the task and deployed environment. All are ingredients in assessing the trustworthiness (using our actionable definition) of a model in an AI-enabled system. The information system would be populated with relevant information during the data curation phase and into model training. The information would consist of questions for data scientists and model developers to answer as they elicit domain knowledge, transform and split data, and train and select models.

\begin{figure}[ht]
  \centering
  \includegraphics[width=1\linewidth]{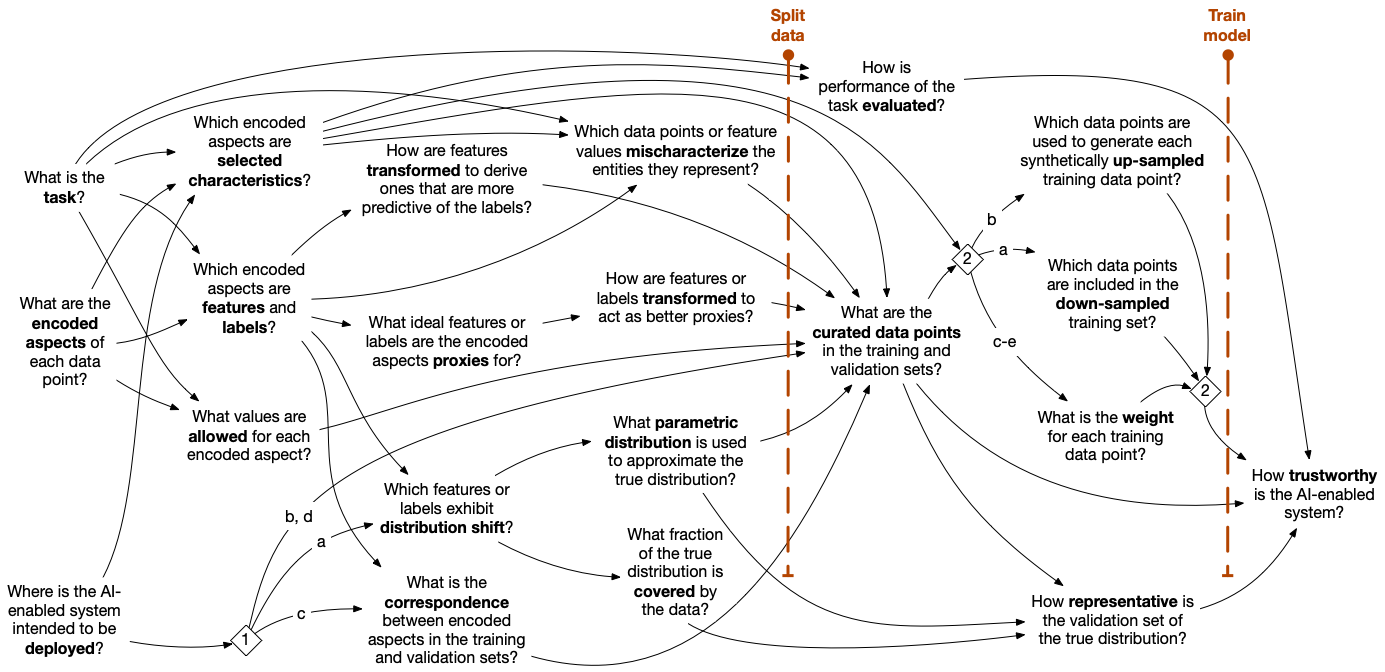}
  \caption{Questions to answer in an information system accompanying an AI-enabled system under development. Arrows trace dependencies; each arrow points from an upstream question to one whose answer is best understood in the context of the upstream question. Two key AI development milestones are noted: splitting data for development into training and validation sets and training a model.}
  \label{figs/information_system}
\end{figure}

Figure~\ref{figs/information_system} is a schematic of the envisioned information system. 

\section{Gather information cumulatively}

In Figure~\ref{figs/information_system}, questions about an AI-enabled system under development are arranged from left to right in an approximate chronological order of being answered. The questions are cumulative, and arrows indicate a dependency structure between the questions: a change in the answer to an upstream question can trigger changes to the answers of downstream ones. For example, consider several of the questions in the upper left corner: only after encoded aspects are enumerated can certain aspects be designated as features or labels for the given task, and after that, data scientists, likely in collaboration with SMEs, can devise a scheme for transforming the features to be more predictive of the labels. Answering the questions from left to right ensures that upstream questions will be answered before ones that depend on them.

Another indicator of chronology in Figure~\ref{figs/information_system} is the inclusion of two milestones: splitting data and training a model. In the data curation phase, data is typically split into disjoint training and validation sets. The training set is used to fit the parameters of a model, while the validation set is used to set hyperparameters of the fitting process, such as the number of iterations. The second milestone is training the model. This typically occurs after data curation has concluded, though there can be some temporal overlap, as in the case of dynamic data reweighting (see Section~\ref{sec:reweight_dynamic}). The milestone is included in Figure~\ref{figs/information_system} because training is prerequisite to answering the rightmost question in the figure: how trustworthy is the AI-enabled system? This question is the cumulation of all the other questions. Here, trustworthiness is defined according to our actionable definition, which is stated in Section~\ref{intro/scope}.

\section{Populate relevant parts of the information system}

Figure~\ref{figs/information_system} calls out decisions 1 and 2 (Chapter~\ref{statistical_tools}) in rhombi, allowing for multiple paths through the question flow. Each decision is among several options, and depending on the development team's choices, different questions are relevant to answer. The decision points and their respective options are detailed in Chapter~\ref{statistical_tools}. For example, in decision 2, options (a) and (b) correspond to down- and up-sampling, respectively, while options (c) through (e) correspond to different kinds of weighting. Only the questions relevant to the development team's choice need to be answered.

\section{Incorporate and supplement information from data cards}

Complementary to data cards, this information system incorporates some information that data scientists may find in data cards.

The following questions concern only the process of encoding information about a given set of real entities or phenomena in data. It is likely that this information could be found in a data card.
\begin{itemize}
    \item What are the encoded aspects of each data point?
    \item What values are allowed for each encoded aspect?
\end{itemize}

All other questions in Figure~\ref{figs/information_system} directly or indirectly depend on questions about the task or deployed environment, neither of which the dataset and data card creator can be expected to know. A thorough data card may contain information about the suitability of the dataset for certain tasks and deployed environments, but it is impossible for it to be comprehensive. New tasks may emerge after the data card was created, and the data card creator may be unaware of some tasks and deployed environments. For example, a dataset could be published publicly with its data card and subsequently used for a national security application, where information about the deployed environment or task is sensitive and therefore would not be mentioned in the data card.

However, it is possible that the task is similar enough to one that the dataset creator envisioned, and in this case, more information can be harvested from the data card. Some of the following questions may be answered in a data card, though the answers should be carefully reviewed to see whether they make sense for the current project's use of the dataset.
\begin{itemize}
    \item Which encoded aspects are features and labels?
    \item What ideal features and labels are the encoded aspects proxies for?
    \item How are features or labels transformed to act as better proxies?
    \item How are features transformed to derive ones that are more predictive of the labels?
\end{itemize}

Other questions, especially those along the bottom row of Figure~\ref{figs/information_system}, are likely to have answers unique to the project at hand. Distribution shift, coverage, and representativeness entail relationships between the data and the deployed environment. Unless the data have previously been used to train a model for this same environment, it is likely that data scientists will need to construct \emph{de novo} answers to the bottom row of questions from domain knowledge they elicit and from their direct experience with the data. These answers are key to assessing trustworthiness of the AI-enabled system under development because they address how well the conditions under which the model is quantitatively validated match those of the deployed environment.

\chapter{Identify and correct mischaracterized data}
\label{sec:distributional_knowledge}
\label{mischar}

Any dataset is the result of a human-centered creative process in which real-world entities or phenomena are simplified into machine-readable representations (Chapter~\ref{data}). The goodness of this process is reflected in how well that data represents, or characterizes, the real world. Drawing on the classic definition of outliers \cite{hawkins1980identification}, we consider \emph{mischaracterized data} to be any data points that deviate sufficiently from the rest of the data that they could have been drawn from a different distribution. The term ``mischaracterization" is intended to subsume similar terms and reflect an agnosticism toward the causes of mischaracterization. That is, given the definition in Section~\ref{data/def}, data points may be considered mischaracterized because of errors, anomalies, or outliers in their features, labels, and/or aspects not encoded in either.

Conceptually, mischaracterized data should be considered to be data points that are ``not right," but this can mean different things in different contexts. Mischaracterized encoded aspects (features or labels) may (a) differ from the overwhelming majority of other aspect values and/or (b) fall outside the allowable range of values. A mischaracterized data point may be (c) highly improbable in the true distribution of inputs in the deployed environment and/or (d) erroneously encode aspects of the real-world entity or phenomenon it represents. Of these, (b), (c), and (d) can only be understood, identified, and resolved through careful elicitation and consideration of domain knowledge (e.g., knowing the allowable range of values, understanding the real-world entities). Identifying the micharacterized data typified by (a), however, requires considering all data points that deviate from the expectations of the true distribution in the deployed environment inclusive of varied types of deviation possible in encoded aspects.

This encoded aspect mischaracterization is possible in any dataset creation or annotation process, whether due to task difficulty, human bias, lack of expertise, or other factors. Building on the taxonomy of noise from \cite{frenay2013classification}, we consider four types of mischaracterization:

\begin{enumerate}
    \item \textbf{Random mischaracterization:} Refers to features or labels that are in error\footnote{By ``in error,'' we refer to an erroneous translation from the real entity or phenomenon of interest to the encoded data point.} but have no statistical relationship between the true features or true labels.
    \item \textbf{Feature mischaracterization:} Refers to features that are in error and have some statistical relationship with the true features (e.g., feature A is typically in units X but is often incorrectly entered in units Y). 
    \item \textbf{Label mischaracterization:} Refers to labels that are in error and have some statistical relationship with the true labels (e.g., label R is challenging to distinguish from label S, so data points with label R have a higher likelihood of being mislabeled).
    \item \textbf{Feature-label mischaracterization:} Refers to data points that are in error (features, labels, or both) such that the error is not random relative to the data point's characteristics (e.g., the values of features A and B make label S more likely to be in error).
\end{enumerate}

Random mischaracterization reflects the category of errors without systemic cause (e.g., an annotator accidentally pressed the wrong key) or possibility for detection via standard statistical methods. While this type of mischaracterization is always possible, it is not typically possible to identify or correct in an automated way.

Most literature on mischaracterization detection or correction considers feature mischaracterization (i.e., outlier or anomaly detection) or label mischaracterization (i.e., mislabel detection), where features or labels are assumed to be fixed and the other is assumed to be a candidate for error \cite{nicholson_label_2015, m_teng_correcting_1999}.
Feature-label mischaracterization considers interrelated feature-label error and thus can only be understood by viewing the data point as a whole. In all such cases, mischaracterization can be said to be a deviation from the true distribution in the deployed environment. Methods for detecting and correcting these deviations are attempts to align development data with the true distribution to improve performance in the deployed environment. 

This section details the intersection of mischaracterization with fairness and trust in AI (Section~\ref{mischar/fairness}), detection of mischaracterized data (Section~\ref{mischar/detect}), correction or removal of mischaracterized data (Section~\ref{mischar/correct}), and challenges to consider when addressing mischaracterization (Section~\ref{mischar/challenge}).

\section{Consider selected characteristics when addressing data mischaracterization}
\label{mischar/fairness}

Handling mischaracterization through the lens of fairness aims to ensure that the development data is corrected for data points that contain values of selected characteristics that weight the dataset away from the true distribution. Properly correcting mischaracterized data often means that the data scientist must pay specific attention to relevant selected characteristics rather than treating each encoded aspect equally. 

Blum and Stangl \cite{Blum_Stangl_2019} frame the problem: 
\begin{quote}
    Rather than argue whether or not these demographic constraints encode intrinsically desirable properties of a classifier, we instead consider their ability to help a learning algorithm to recover from biased training data and to produce a \textit{more accurate} classifier. [...]  [fairness] constraints might actually help prevent the optimizer from being led astray, with a higher quality solution when accuracy is \textit{measured on the true distribution}. 
\end{quote}
Blum and Stangl argue that including fairness metrics in mischaracterization detection and correction enables the data scientist to create a more accurate dataset. While fairness has the goal of making sure that members of minority selected characteristics are not overlooked by the model, taking fairness into consideration in ML leads to better results along the true distribution. 

Verma et al. \cite{Verma_Ernst_Just_2021} find that when they preprocess their dataset to rank and remove the data points that contribute to biased decision-making, their model is not only more fair, according to the metric of individual discrimination, but also more accurate than other comparable models. 

\section{Detect mischaracterized data}
\label{mischar/detect}

Following a taxonomy of mislabel detection \cite{xiong_towards_2024}, this report considers three categories of mischaracterization detection approaches:
\begin{enumerate}
    \item{\textbf{Expert review} includes activities where a human reviews the data directly, including ad hoc inspection and interactive analysis.}
    \item{\textbf{Statistical detection} includes methods to try to find all data points that deviate statistically from the expected distributions of features and labels (e.g., kernel methods for anomaly detection).}
    \item{\textbf{Model impact analysis} refers to understanding how individual data points influence model outputs. 
    These approaches seek to identify mischaracterization not from the perspective of the data itself but from the data's influence on the model.}
\end{enumerate}

Exhaustive expert review can be effective but prohibitively time-consuming and expensive \cite{frenay2013classification}.
Visually identifying outliers in low-dimensional space using methods like uniform manifold approximation and projection (UMAP) \cite{mcinnes2018umap} can help simplify this 
process for SMEs and data scientists by depicting data points in (or away from) major clusters in feature or label space. Tools like OoDAnalyzer \cite{chen_oodanalyzer_2020} can also help data scientists visualize and understand which features are learned by models and how they affect model output.

Statistical detection utilizes different algorithms to identify data points that significantly deviate from the expected value along a certain axis. Different outlier detection algorithms are appropriate for different types of outliers. Below is a nonexhaustive list of outlier detection algorithms. 

\begin{itemize}
    \item \textbf{Deep Support Vector Data Description (SVDD)} \cite{pmlr-v80-ruff18a}: Learns a network that minimizes the volume of the hypersphere that encloses the data; normative points are in the center of the hypersphere, outliers are by definition not enclosed in the hypersphere; the intuition is that the network has learned a representation of the data that captures the most salient aspects of the data's true distribution.
    \item \textbf{Deep Fair SVDD} \cite{zhang_towards_2021}: Extends original Deep SVDD with a notion of group fairness; an adversarial network is learned to de-correlate selected characteristics and learned representations.
    \item \textbf{Local Outlier Factor (LOF):} Calculates the local density around a data point $\mathnormal{X}$. First the maximum distance between $\mathnormal{X}$ and its $\mathnormal{n}$ nearest neighbors $\mathnormal{X_n}$ is found. This distance is then used to calculate the \emph{Local Reachability Density}, which is then used to calculate the \emph{LOF} of the data point \cite{p_fair_2020}.
\end{itemize}

Model impact analysis often uses data valuation techniques to identify specific data points, or even features or labels, that seem to erode model performance. Section~\ref{sec:reweight_model_agnostic} describes data valuation techniques as means for computing weights, but some of the same techniques can be used to detect mischaracterized data \cite{Liu_Just_Chang_Chen_Jia_2023}. Unlike other data curation techniques for detecting and mitigating mischaracterized data, model impact analysis has results that change depending on the choice of model architecture. This raises questions about whether the techniques are actually detecting data points that have been erroneously recorded, and it requires data scientists to know the model architecture during data curation. 

\section{Correct and prune data points}
\label{mischar/correct}

In contexts where the deployed environment is expected to be noisy (i.e., contain outlier data points), ML models can be optimized with robustness toward outliers in mind. Certain ML model classes, such as ensemble methods and decision trees, are designed to be robust to noise \cite{frenay2013classification}. Additionally, methods can model label noise directly as part of the training paradigm, simultaneously learning models that predict, from a data point's features, its label and its likelihood of being in error \cite{frenay2013classification}.

If the data in the deployed environment is expected to be generally noise-free, however, then outlier detection methods should be used and the outliers should be handled accordingly.  Once outlier data points are detected, they can either be corrected or pruned from the dataset. 
To train the model on the most accurate data possible, relabel or remove data through heuristics or learned models before model-fitting \cite{frenay2013classification}. 

\subsection{Outlier correction or pruning}
\label{mischar/correct/general}

Many mischaracterization detection and correction algorithms go hand in hand: these are algorithms built to detect and correct data points simultaneously. The label correction methods explained in this section are framed with respect to a supervised binary classification problem but can be modified to fit a multiclass classification problem. 
These data cleaning methods appear to be particularly actionable. However, these methods make a few key assumptions:
\begin{enumerate}
    \item There is enough correctly labeled data to represent the true distribution of labels in the dataset such that a  ML system will be able to discover the appropriate labels through creative subsampling methods. Specifically, it is assumed that part of the dataset already reflects the true distribution. 
    \item All features are initially weighted the same; therefore, selected characteristics are not treated differently from the rest of the features.
    \item These methods generally assume that there exists some set of outliers $\mathcal{O}$ that can be detected in the dataset. 
\end{enumerate}

Any and all of these assumptions are not necessarily true for most real-world data. 
Nicholson et al. \cite{nicholson_label_2015} compare three label correction methods, all of which involve training multiple classifiers on the training set in order to ``smooth" the outliers of the data:
\begin{itemize}
    \item \textbf{Polishing labels} \cite{nicholson_label_2015} is a label correction method based on Teng's work \cite{m_teng_correcting_1999}. Although Nicholson et al.'s version does not account for mischaracterized features, Teng's original work does, and thus is more relevant to the fairness discussion. Nicholson et al.'s version trains 10 classifiers with a single classification algorithm on 10-fold split of the data. Each of the classifiers then predicts the label for each observation, and the classifiers vote to select the final label. 
    
    Teng's work uses a similar method, but it trains classifiers for every feature one at a time, using the other features and the label column as the training features. It uses these classifiers to identify ``weak" attributes. Then, a 10-fold cross validated classifier is trained on the training set. For the observations this classifier misclassifies, 
    the values of the weak attributes that were nominated in the initial phase are iteratively tweaked to see if these tweaks will improve the classification rate of the classifier. If no attribute manipulation will fix the classification, then the label for the observation is deemed incorrect and will be corrected here. This method effectively smooths both the label space and the feature spaces. 
    \item \textbf{Self-training correction} \cite{nicholson_label_2015} uses a noise-filtering algorithm to split the data into ``clean" and ``noisy" sets. A classifier is trained on only the clean set, and the noisy observations are run through this classifier. The noisy observations are then ranked by the likelihood that they are misclassified, and the top $\mathnormal{n}$ observations' labels are flipped.
    \item \textbf{Cluster-based correction} \cite{nicholson_label_2015} repeatedly clusters the data based on features alone and then assigns each data point a label that was common within the data point's clusters. This method works well if the label space is noisy but the feature space is not, as the clustering results are unsupervised and therefore label-agnostic.
   \end{itemize}

The cluster-based correction method performed the best in Nicholson et al.'s survey of these label correction methods. While their survey focused on noisy label space rather than noisy label and feature spaces, any of these methods could be applied to any given feature.

Another popular data correction method is called confident learning (CL), which is the backbone architecture of a popular data-cleaning library called \href{https://cleanlab.ai/}{Cleanlab}\footnote{https://github.com/cleanlab/cleanlab} \cite{northcutt2021confidentlearning}. CL counts the labels that are likely mischaracterized using a ``confident joint." It then prunes the noisiest data and ranks the confidence of the remaining training data \cite{Quaresmini_Primiero_2023}. Note that CL does not attempt any form of label correction; therefore, it is assumed that there is enough development data to withstand pruning.

\subsection{Fair correction and pruning}
\label{mischar/correct/fair}

The examples discussed in Section~\ref{mischar/correct/general} correct mischaracterized data and features regardless of the existence of a selected characteristic. However, it may be the case that the correction of data must account for whether or not the value being changed is part of a selected characteristic. As discussed in Section~\ref{mischar/fairness}, smoothing the data too much might be detrimental to the overall fairness of the dataset, even if it makes it more amenable to the downstream model. 

P and Abraham \cite{p_fair_2020} created a fair version of the LOF outlier detection algorithm, an unsupervised method that quantifies the ``local object density" of each data point, comparing it to the other data points in its neighborhood. Their LOF method aims to align the distribution of data points with selected characteristics in the outlier set with the distribution in the development data. 
This is referred to as representational parity or disparate impact avoidance. Fair-LOF aims to ensure that individual minority features are not discriminated against. Outlier detection in this framing is challenging in part because selected characteristics may be encoded as proxy variables (e.g., last name may indicate ethnicity) or may not be directly encoded at all.

Unlike previous fair anomaly detection methods, counterfactual fairness defines fairness as a function on the causal effect that the selected characteristic has on model outcomes. Counterfactually fair anomaly detection (CFAD) aims to create an anomaly detection algorithm that is fair as well as ``counterfactually fair." Han et al. \cite{Han_Zhang_Wu_Yuan_2023} define a counterfactually fair anomaly detection algorithm as one that yields the same anomaly score for a data point even if the value of its selected characteristics changed.

When the selected characteristic is only partially labeled, it is beneficial to impute the dataset with pseudo-labels for the missing selected characteristic values before passing the data to a fair classifier. Jung et al.'s \href{https://github.com/naver-ai/cgl_fairness}{confidence-based group label assignment (CGL)}\footnote{https://github.com/naver-ai/cgl\_fairness} \cite{Jung_Chun_Moon_2022} proposes training a classifier for the selected characteristic on the data points that are labeled and using it to assign pseudo-labels for high-confidence unlabeled values (based on a fine-tuned threshold). For the low-confidence values, labels are randomly selected. The authors show that using CGL outperforms dropping unlabeled data, random pseudo-label assignment, and relying solely on a classifier for pseudo-labels \cite{Jung_Chun_Moon_2022}.

CL as a noise detection and pruning algorithm does not cover all real-world scenarios because it assumes ``categoricity," or features fitting perfectly into categories. For example, if time of day is a selected characteristic in an image classification problem, the binary labels of ``night" and ``day" are not appropriate for images of ``twilight" in either the training set (mislabeled cases) or the deployed environment (insufficient coverage). Because the domain of this label set does not encompass the true distribution of values, CL will not work. In general, categorical values can develop and change over time. This means that the deployed environment might have more categories for a certain feature than exist in the training set because of updating labeling schemata. One might need to detect both mislabeled data points and incomplete feature sets in the training set. Either of these can affect accuracy measurements. Quaresmini et al. \cite{Quaresmini_Primiero_2023} propose evaluating the dataset for completeness over time to the data evaluation process. 

\section{Exercise caution when identifying and correcting mischaracterized data points}
\label{mischar/challenge}
The attributes of the dataset must be considered when weighing which mischaracterization detection and correction methods are appropriate. If the dataset has few observations, it may not be large enough to be able to withstand pruning or correcting data points. 

Alternately, if the dataset is too large, a method that requires training classifiers could be too resource-intensive. Even if computational load is not a concern, there might not be enough representative data points for every relevant class. Correction methods will not smooth correctly along dimensions with sparse data, resulting in decreased accuracy \cite{Guha_Khan_Stoyanovich_Schelter_2024}. For observations that are members of minority classes, extra care must be taken. Correction methods must not simply drop or change the value of all minority variables to the majority value. 

Finally, it is important to make sure that the architecture used in the mischaracterization detection and correction steps is different from the architecture used in the final model. If the curation architecture is also used for the overall model architecture, then unexpected behavior can be encoded. Matching architectures might result in the curation step removing data points that the model architecture struggles to learn, when our goal is to remove data points that are uncharacteristic of the true distribution. Data points that are common in the true distribution but difficult for a model to perform well on must be retained during curation for later exposure to the model.

\section{Practitioner's perspective on detecting mischaracterized data with Cleanlab}
\label{mischar/practitioners}

We assessed mischaracterized data by using the Python package Cleanlab, which implements CL. As mentioned previously, the CL method identifies labels that are likely mischaracterized and then prunes the noisiest data and ranks the confidence of the remaining data. The algorithm is discussed in more detail in Section~\ref{mischar/correct/general}.

We used Cleanlab in an object detection scenario using the large, publicly available dataset VisDrone. While the VisDrone dataset was curated specifically for object detection, the dataset is large and the method by which the labeling was performed is unknown. Mischaracterized labels could exist in the dataset, but manually examining each image is far too time-consuming. Cleanlab can be used to identify and correct mischaracterized labels of people in the dataset, thereby increasing trustworthiness in our model.

The CL method in Cleanlab requires an out-of-sample prediction for every data point being assessed for mischaracterization. An out-of-sample prediction is defined as a prediction for a data point on which the model was not trained. If you want to assess mischaracterization only in the validation set, then you can use the model trained on the training data to generate out-of-sample predictions for the validation set and subsequently use these predictions with the Cleanlab package. 

However, if you want to assess mischaracterization in the entirety of your data, both the training and the validation sets, there are two options for obtaining out-of-sample predictions:

\begin{enumerate}
    \item Use cross-validation for training such that for each data point, there is a model trained where that data point is only included in the validation set, and thus its predictions for that model are out-of-sample.
    \item Use a pretrained model that was not trained using any of your data.
\end{enumerate}

Option 1 would avoid the use of a pretrained model, yielding better transparency in the model training. However, training one model, let alone training multiple models, using cross-validation can be computationally expensive. For our object detection scenario, training just one of the cross-validation models would take over two days. The computational expense led us to consider using a pretrained object detection model for obtaining out-of-sample predictions.

Option 2 uses a pretrained model to obtain out-of-sample predictions. In our case, we found that the pretrained object detection ResNet model was not trained on the VisDrone dataset and could therefore produce out-of-sample predictions for our training data.

Even when using a pretrained model, conducting inference using an object detection model is still computationally expensive. We opted to produce out-of-sample predictions only for a small subset of our training data to illustrate the exercise of assessing mischaracterized data. Ideally, this process would be performed on all training and validation data, but the computational cost would need to be weighed against the potential benefits of identifying mischaracterized data.

A final consideration is that Cleanlab removes those data points that are found to be mischaracterized. Either the labels would need to be manually fixed for these removed data points or the training set would need to be large enough to withstand this pruning. In our scenario, the VisDrone dataset is large enough to withstand pruning, so Cleanlab can be used to identify and remove mischaracterized data points in the training set.

Cleanlab provides a \href{https://docs.cleanlab.ai/stable/tutorials/object_detection.html}{demonstration}\footnote{https://docs.cleanlab.ai/stable/tutorials/object\_detection.html} of assessing mischaracterization for object detection, along with two tutorials:
\begin{itemize}
    \item Producing out-of-sample predictions for a validation set \href{https://github.com/cleanlab/examples/blob/master/object_detection/detectron2_training.ipynb}{only}\footnote{https://github.com/cleanlab/examples/blob/master/object\_detection/detectron2\_training.ipynb}
    \item Producing out-of-sample predictions for both the training and validation \href{https://github.com/cleanlab/examples/blob/master/object_detection/detectron2_training-kfold.ipynb}{sets}\footnote{https://github.com/cleanlab/examples/blob/master/object\_detection/detectron2\_training-kfold.ipynb} 
\end{itemize}

Note that Cleanlab uses the term ``test" set to refer to our concept of a ``validation" set. Also note that the tutorials involve training a \href{https://github.com/facebookresearch/detectron2}{Detectron2}\footnote{https://github.com/facebookresearch/detectron2} model. We found the Detectron2 framework rather difficult to install, and we found training with the framework to be very computationally expensive. These issues also contributed to our decision to use a pretrained model to obtain out-of-sample predictions.

The \href{https://github.com/cleanlab/cleanlab}{Cleanlab package}\footnote{https://github.com/cleanlab/cleanlab} is actively maintained as of December 2024, and instructions on installation and use can be found in the relevant \href{https://docs.cleanlab.ai/stable/index.html}{documentation}.\footnote{https://docs.cleanlab.ai/stable/index.html}

\chapter{Curate data for AI-enabled systems that involve pretrained models}
\label{sec:pretrained}
In many AI-enabled systems, capabilities leverage or extend existing AI/ML tools. In some cases the data used to train the original models is available (e.g., a project requires updating or refining a model that was trained by the same team), but usually training data is unavailable, as in the case of many pretrained embeddings or commercial generative language models. Data curation to support AI trustworthiness can be applied not only to new AI-enabled systems but also to contexts where some relevant data is inaccessible to the data scientist for curation. 

\section{Learned embeddings are features}
\label{perspective/embed}
When considering the data curation paradigm represented in this report, particularly relative to the knowledge elicitation described in Section~\ref{knowledge/transformation}, it is possible to misconstrue data curation as applying only to contexts in which the data and their aspects are interpretable to data scientists and SMEs. For example, if the data is represented as a table -- each row a data point, each column a feature -- domain knowledge can be elicited about the range of allowable values in a column or what a particular column means. Data curation for trustworthy AI, however, is not limited to this framing and should be understood inclusive of all contexts in which ML models are trained for use in a deployed environment, regardless of how the data is represented.

While data representations can be restricted to human interpretability (e.g., a date, a temperature, the frequency of a word), data prepared for ML contexts is typically embedded as dense vectors (Section~\ref{data/def}). These dense vectors are often learned parameters that generalize across inputs, such that similar inputs yield similar vectors \cite{goldberg2022neural}. From the perspective of the model, these representations are features (inputs over which model parameters will be learned) even though they are not explicitly interpretable by humans. These opaque representations are used as inputs largely for their flexibility, and in the context of neural networks they can enable effective learning of task-relevant parameters, since -- if the network was provided with human-interpretable features -- the first layer(s) of the network would be devoted to learning a dense representation as a consequence of the network architecture \cite{goldberg2022neural}. 

As defined in Section~\ref{intro/scope}, data curation is concerned with the creation of training and validation datasets from available data. That is, the development-environment inputs to an AI-enabled system are curated with the goal of manipulating that system (i.e., the parameters of a model) to enable trustworthiness in its performance in the deployed environment . In this broad framing, knowledge elicitation (Section~\ref{knowledge/transformation}), reweighting data points (Section~\ref{sec:reweight_model_agnostic}), and other curation activities can be seen as trying to understand and curate data at a degree of abstraction inclusive of, but not limited to, the mechanism by which the data is encoded. 

\begin{figure}[ht]
  \centering
  \includegraphics[width=.85\linewidth]{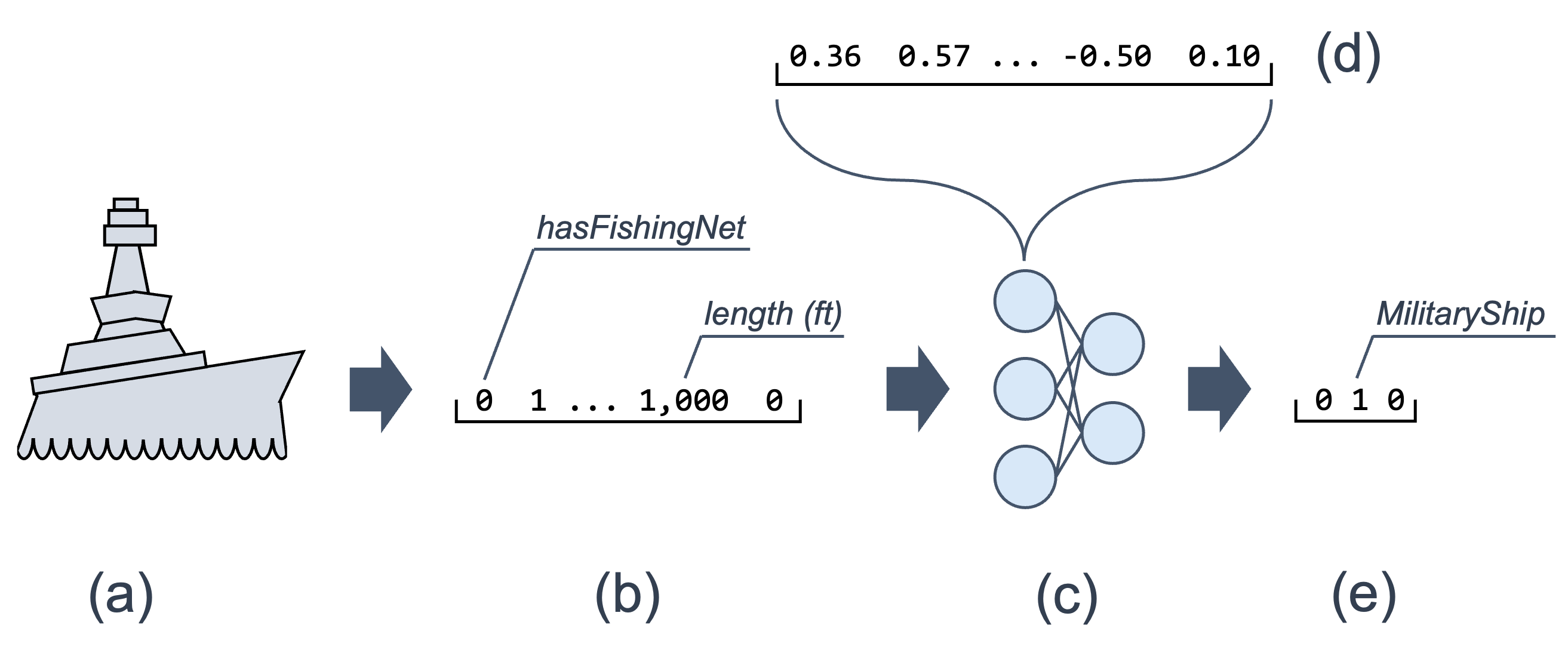}
  \caption{Example AI pipeline. A real-world entity (a) is encoded with human-interpretable attributes into a feature vector (b); input vectors are passed to a neural network (c), which learns a dense representation of the input (d) and yields a prediction (e).}
  \label{figs/pipeline_ex}
\end{figure}

Consider the example AI pipeline shown in Figure~\ref{figs/pipeline_ex}. In this setting, real-world entities (e.g., a ship) are encoded with human-interpretable features (e.g., whether there is a fishing net, the length in feet), which are then passed to a network that learns a dense representation as part of the process of yielding a prediction (e.g., that a given ship is a military rather than civilian watercraft). Data curation in this setting involves selecting/transforming data points (b) in support of training the task-specific model (c). As discussed in Section~\ref{knowledge}, these curation activities can benefit from knowledge elicitation about the input entities (a), the process by which those entities are encoded as features (b), and the expectations of the final model's predictions (e).

\begin{figure}[ht]
\centering
\begin{subfigure}{\textwidth}
  \centering
  \includegraphics[width=.7\linewidth]{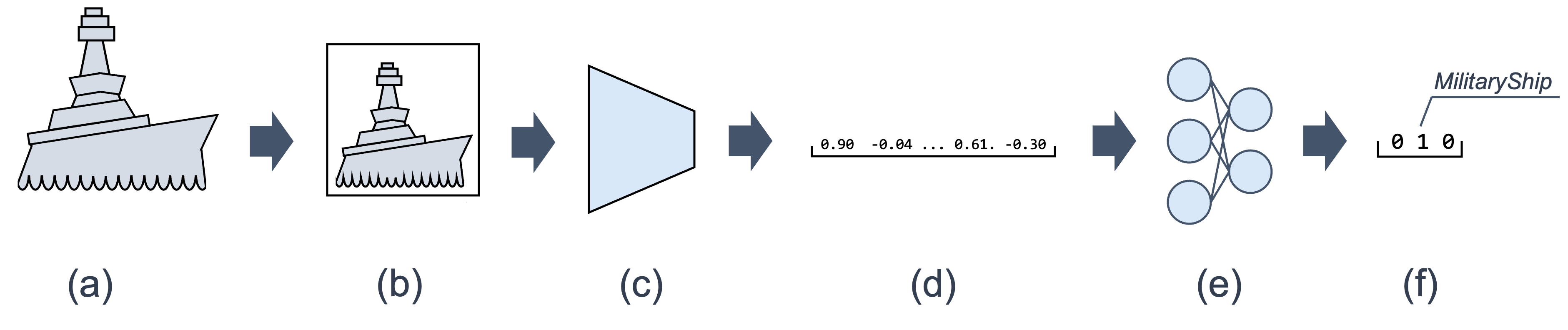}
\end{subfigure}

\begin{subfigure}{\textwidth}
  \centering
  \includegraphics[width=.7\linewidth]{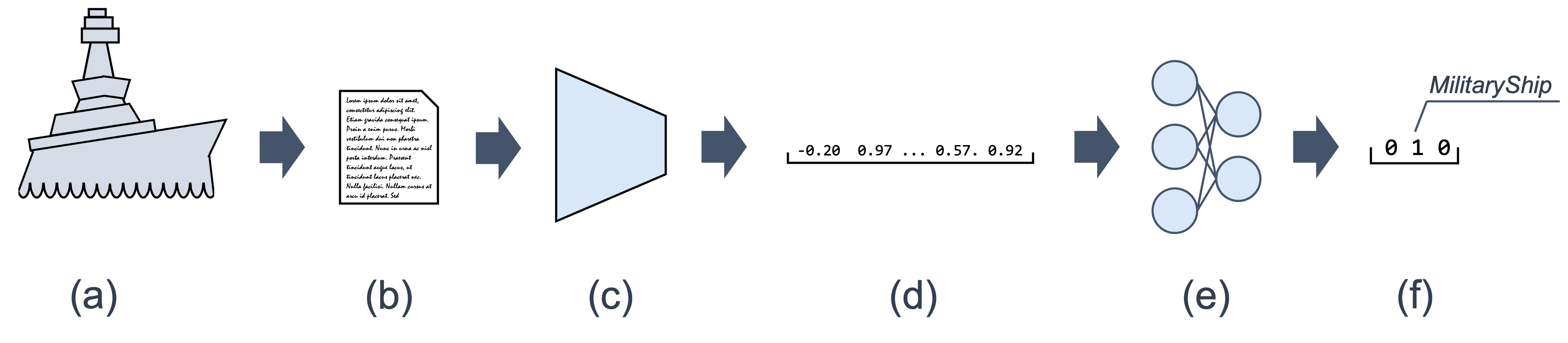}
  \end{subfigure}
  \caption{Example AI pipelines. A real-world entity (a) is represented as an image (top) or text (bottom) (b); some process (c; e.g., pretrained model) embeds the representation as a dense vector (d); input vectors are passed to a neural network (e), which yields a prediction (f).}
  \label{figs/pipeline_ex_modality}
\end{figure}

Other AI pipelines, such as those shown in Figure~\ref{figs/pipeline_ex_modality}, conceptually vary on their input modality. Real-world entities may be represented as images (top) or natural language descriptions (bottom), which are then encoded by some process to produce dense input vectors. This process may be transparent to the data scientist (e.g., naively vectorizing fixed-sized images by pixel values) or opaque (e.g., a pretrained language model), but the result is a dense vector without human-interpretable features. These vectors are then used to train the task-specific model. These pretrained models are often utilized in practice, as the large amounts of training data available to the companies that deploy such models lead to rich and generalizable representations.

In practice, data curation for the kind of AI-enabled system shown in Figure~\ref{figs/pipeline_ex} may differ from the kinds of systems shown in Figure~\ref{figs/pipeline_ex_modality}. For example, while SMEs can provide insight into single features in the former, they cannot in the latter and may not have insight into the process by which input data modalities (e.g., images) are transformed into dense vectors. These differences, however, are few when compared to the breadth of data curation activities that remain consistent across these and other kinds of AI-enabled systems.

AI-enabled systems seek to make inferences (e and f, respectively) about real-world entities (a) using ML models (c and e, respectively). Data curation is concerned with the data used for training those models and the relationship among the real-world entities, the available data, and the expected behavior of the system. The activities described in this report are thus applicable even in contexts where the inputs are not interpretable by humans or where the process by which one kind of data modality (e.g., text) is transformed into another (e.g., word embeddings) by an opaque process (e.g., a pretrained model).

\begin{figure}[ht]
  \centering
  \includegraphics[width=.8\linewidth]{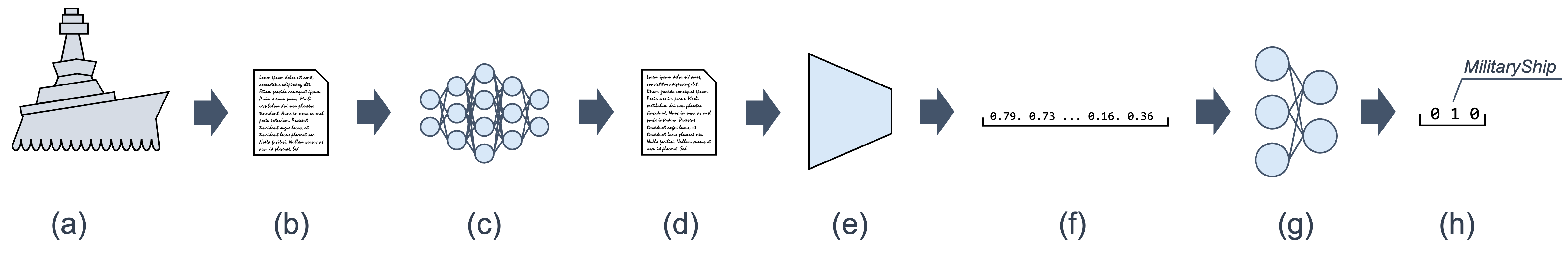}
  \caption{Example AI pipeline. A real-world entity (a) is represented as text (b; e.g., natural language description); a pretrained generative model (c; LLM) generates task-dependant text output (d); some process (e; e.g., pretrained model) embeds the text as a dense vector (f); input vectors are passed to a neural network (g), which yields a prediction (h).}
  \label{figs/pipeline_ex_llm}
\end{figure}

This perspective even applies to AI-enabled systems with generative components like LLMs, such as that pictured in Figure~\ref{figs/pipeline_ex_llm}. In such systems, real-world entities may be represented as text over which an LLM makes inferences and outputs new text, and this new text is then embedded and passed to a task-specific network. While an LLM is an opaque component in the overall pipeline, data curation proceeds in the same way as it does in Figure~\ref{figs/pipeline_ex_modality} (bottom): knowledge about the task and data must be elicited, data points can be weighted and sampled, etc. However, ensuring trustworthiness while using generative components leads to more complexity in data curation, as the training data for the models are often inaccessible to the user. This can lead to an incompatibility between the knowledge stored in the pretrained model used to generate the features and the knowledge learned from the data at hand.  The rest of this chapter delves further into these complexities. 

By taking a sufficiently broad perspective, data curation activities to enable trustworthy AI can be applied to a wide variety of systems, even when there are opaque components or representations without explicit human-interpretable features.

\section{Apply trustworthy data curation to pretrained embedding models}
\label{perspective/embedding}
Embedded text or images are often used as features for ML models (Section~\ref{perspective/embed}). In practice, these embeddings often originate from a pretrained model for which the data scientist typically does not have access to the training data. If the pretrained model is unable to perform sufficiently on the true distribution of inputs in the deployed environment, downstream AI-enabled systems (including other models) using these embeddings may not be trustworthy according to the actionable definition (Section~\ref{intro/scope}) in this report. To support curation actions in this environment, the difference between the true and expected behavior of these embeddings must be quantified, and where possible, bias must be corrected.

\subsection{Measure bias in pretrained embeddings}
\label{sec:measure_we_bias}
We consider \emph{biased} embeddings as those that lead to unequal performance along the dimension of a given selected characteristic where performance is expected to remain equal (Section~\ref{data/fairness}). For example, it may be a requirement that object detection performance for ships should remain the same for images that do and do not include coastline. In that case, the image embeddings might need to be neutralized along this dimension to ensure similar performance between data points.  To remove any bias from embeddings, it is crucial to be able to measure their performance along a predefined dimension of interest. 

All measurement and correction techniques depend on model architecture and its dynamism. Visual media is constantly changing and requires a trained model to generate embeddings at runtime. In contrast, there is a limited number of words in a vocabulary; therefore, text embedding models are often described as either static or contextual \cite{Zhang_Lu_Abdalla_McDermott_Ghassemi_2020}. Static word embeddings such as Word2Vec or GloVe \cite{Pennington_Socher_Manning_2014, Mikolov_Chen_Corrado_Dean_2013} do not change given the surrounding context of the word in question. These embeddings inherently remain the same in all contexts, which means that  each individual embedding can be transformed to ensure trustworthiness. Contextual word embeddings, such as those retrieved from transformer models (e.g., BERT \cite{Devlin_Chang_Lee_Toutanova_2019}), are trained on a wide variety of natural language tasks and return different representations depending on the context provided. These embeddings can therefore distinguish between homonyms as well as differences between word usage depending on the domain being discussed. While this significantly improves expressivity in word embeddings, the debiasing quantification or correction methods become more complicated, because it is not as simple as performing the correction method on $v$ word embeddings, where $v$ is the size of the vocabulary. Image bias measurements would be similar to contextual word embedding measurements, as there is no static set of image embeddings representing all images.

Measuring bias can be difficult depending on the modality and the selected characteristic for which equal performance matters, but it can be measured by the effect a set of embeddings has on a downstream ML task. While there is no all-purpose solution for this, training a lightweight model on a simple task and measuring cross-class performance can help reveal how embeddings will perform on the larger system. Bias in embeddings has been defined to be mathematically detectable via geometric distance, clustering, or downstream tasks \cite{Schroder_Schulz_Kenneweg_Feldhans_Hinder_Hammer_2024}. Some methods compare embeddings across different groups within a selected characteristic to ``target words," which represent positive and negative bias (e.g., comparing gendered words to ``pleasant" and ``unpleasant"), and the geometric distance between the group's embeddings and the target words is used to calculate bias \cite{Bolukbasi_Chang_Zou_Saligrama_Kalai_2016, Caliskan_Bryson_Narayanan_2017}. An ``unbiased" embedding would have similar distances between the target words for each class and the attribute descriptors. A biased embedding might have one class situated closer in the embedding space to an attribute than another. 

These distance-based detection and debiasing methods are popular but may not be completely thorough \cite{Gonen_Goldberg_2019}. Goenen and Goldberg \cite{Gonen_Goldberg_2019} instead suggest using clustering or classification tasks rather than distance measurements for evaluating bias in embeddings. For the clustering method, the goal is to make sure that concepts that are supposed to be neutral cannot be clustered (consider again the case of images of military and civilian ships with and without shoreline). Finally, there has been a recent push to measure bias based on downstream task performance. For example, consider evaluating whether the coastline interferes with the performance of an object detection model using image embeddings \cite{Schroder_Schulz_Kenneweg_Feldhans_Hinder_Hammer_2024}.

In cases where selected characteristics are legally defined demographics (e.g., race, gender), open-source datasets and packages, such as WinoBIAS \cite{Zhao_Wang_Yatskar_Ordonez_Chang_2018} or MDGENDER \cite{dinan-etal-2020-multi}, aim to measure bias in word embeddings. WEAT \cite{Caliskan_Bryson_Narayanan_2017}\footnote{WEAT: \url{https://wefe.readthedocs.io/en/latest/api/generated/wefe.metrics.WEAT.html}} and SEAT \cite{May_Wang_Bordia_Bowman_Rudinger_2019} are static distance-based measurement packages for bias within word embeddings. SEAT is a sentence-based package using a contextual model to create the embeddings but treats sentences as static embeddings. A sentence embedding can be a more useful measurement than using a single word, since words carry a different meaning depending on the words surrounding them. Generally, however, bias measurement requires task-specific data and models.

\subsection{Correct for bias in embedding models}
\label{sec:correct_we_bias}
The seminal word embedding debiasing paper by Bolukbasi et al. \cite{Bolukbasi_Chang_Zou_Saligrama_Kalai_2016} projects word embeddings into a space orthogonal to the dimension in which the debiasing is going to occur. This space is found using definition words describing the axis of bias. The use case for this paper was to remove gender bias from nongendered words. However, this same concept can be applied to any selected characteristic for which one would want equal performance.  Manzini et al. \cite{Manzini_Lim_Tsvetkov_Black_2019} extend this method to nonbinary definitions of bias; they use a method called \emph{soft debiasing} to calculate ``a projection of the
embedding matrix that preserves the inner product between biased and debiased embeddings while minimizing the projection onto the bias subspace of embeddings that should be neutral". This method is more complex than the method of Bolukbasi et al., and it allows for multiclass debiasing. All of these methods work by identifying the selected characteristic subspace of the embedding model and then removing the bias component from the vector representation. While debiasing often needs to be implemented on a case-by-case basis,  \href{https://github.com/dccuchile/wefe}{Word Embedding Fairness Evaluation (WEFE)} \cite{Badilla_Bravo_Marquez_Perez_2020} implements a debiasing package for single-class and multiclass debiasing, which applies the debiasing method of Bolukbasi et al. DebIE \cite{Friedrich_Lauscher_Ponzetto_Glavas_2021} is a rest API for measuring and debiasing static word embeddings.

These methods apply only to situations where a fixed number of embeddings is used, which is most common when using static embeddings.

Contextual models are made up of many layers and parameters. Biases can be attributable to any one or combination of these parameters or layers. Therefore, simple projection methods (such as that of Bolukbasi et al. and similar) cannot be used \cite{Kaneko_Bollegala_2021} unless the embeddings belong to a fixed set and can thus be predetermined. Adversarial representation learning (ARL) is an approach used for debiasing embeddings across modalities \cite{Sadeghi_Boddeti_2020, Zhang_Lemoine_Mitchell_2018}. The core idea behind ARL is to optimize three separate networks: a network that encodes the input data while ``intentionally and permanently eliminating the information corresponding to a sensitive attribute," a network whose goal is to identify the target attribute from the embeddings, and an adversarial network whose goal is to identify the sensitive attribute \cite{Sadeghi_Boddeti_2020}. The goal of this approach is to maximize the target attribute predictor's ability to achieve the task at hand while minimizing the sensitive attribute predictor's performance \cite{Zhang_Lemoine_Mitchell_2018}. Although this is a promising debiasing method, biases have been shown to persist in the resultant vectors \cite{Zhang_Lu_Abdalla_McDermott_Ghassemi_2020}. 

Alternatively, many newer approaches focus on fine-tuning large foundation models to eliminate bias in the dimension of a selected characteristic. By fine-tuning a model that is already successful at a general task, one can continue to take advantage of the information the model learned from a vast amount of training data while minimizing the effect of biased, unstructured data \cite{Kaneko_Bollegala_2021}. An expanded discussion on fine-tuning foundation models can be found in Section \ref{sec:finetune}. However, as was the case with adversarial debiasing, it is unlikely that all bias will be eliminated from the final fine-tuned product, and fine-tuning can lead to catastrophic forgetting, where the model's performance on previous tasks is significantly harmed. Other approaches aim to return to the simplicity of the Bolukbasi et al. approach. Rakshit et al. compare soft debiasing to a similar method that uses a neural network to learn the transformation rather than only a projection matrix (more on transformation and projection methods can be found in Section \ref{sec:projection}). In theory, any of these could be post-processes after embeddings are retrieved from the foundation model \cite{Rakshit_Singh_Keshari_Chowdhury_Jain_Chadha_2024}. 

\section{Apply trustworthiness techniques to improve the quality of the output of generative AI}
\label{perspective/gen}

In Section~\ref{perspective/embed}, we argue that the embeddings from large pretrained models can be used as features in downstream models. These features are inherently not human interpretable. With the recent boom in the development of generative foundation models, some of which fall into the category of LLMs (e.g., GPT), the \emph{output} of such generative models is often used in downstream ML tasks \cite{Yang_Jin_Tang_Han_Feng_Jiang_Zhong_Yin_Hu_2024}. All of these models, regardless of size and modality, are probabilistic models -- meaning that they learn and abstract information solely from their training data. In many cases, LLMs work by predicting the next most likely word-piece conditioned on the information that came before it (the prompt and whatever output has been generated so far by the model). Although these models are classically considered ``black boxed," by understanding how these models are trained, a data scientist can begin to consider trustworthiness modifications that can be applied to their LLM-enhanced system. The following section primarily focuses on text models; however, many of these methods can be applied to a general foundation model, a similarly pretrained and large model that exists in a variety of modalities.

\subsection{The challenge of ensuring trustworthiness in closed-source models}
The output of large generative models models may be human interpretable, but -- because of cost and other factors -- many companies consider the internal aspects of the models (like embeddings, pre-processing steps, and internal weights) proprietary \cite{Yang_Jin_Tang_Han_Feng_Jiang_Zhong_Yin_Hu_2024}. Because a model's internal workings are often be inaccessible to the user, they cannot be used to remove bias. 

Data scientists can be using LLMs accessible via API endpoint, or may be restricted in their available computational resources such that they cannot edit the embedding space of the LLM. In this case, the embedding transformations discussed in Sections \ref{perspective/embedding} and \ref{sec:finetune} might not be applicable. Ensuring fairness from this output is a harder task, as the embeddings themselves (and therefore the downstream output) cannot be directly transformed along a predefined dimension of fairness. Additional trustworthiness concerns in generative models stem from their nondeterministic nature, the opaqueness of their decision-making processes, and the scale of their training data (i.e., a model may be too large to realistically vet for nontrustworthy sources).

\urldef{\urlHFGrammar}\url{https://huggingface.co/docs/text-generation-inference/en/guidance#grammar-and-constraints-} 
Despite these drawbacks, it is possible to encourage trustworthy output from these models. Some models provide the user the ability to constrain the model to output only certain tokens; however, this requires a grammar definition from the data scientist \cite{geng2023grammarconstrained}. These constraints usually work by taking the logits from a model's prediction, passing them through the grammar to cancel out any tokens that are invalid, and then selecting the best option from the remaining available tokens. Constrained grammar implementations are relatively new innovations and are quickly being integrated into many proprietary models and model-hosting platforms.\footnote{Hugging Face incorporated a grammar parameter for the relevant text generation models that it supports: \urlHFGrammar} Data scientists can use  other methods to encourage trustworthy output from these models. Measurements exist to quantify the trustworthiness of the models before usage, and methods such as prompt engineering and retrieval-augmented generation (RAG) can help prevent inappropriate or ungrounded outputs, or hallucinations. Finally, if the data scientist has the ability to train these models or access their embedding spaces, it is possible to shift the model's output to encompass the domain of the true distribution.

It is not curently possible to completely ensure trustworthiness in large pretrained models because of the lack of documentation and structure in vast training corpora and the general applicability of these models. Evaluating and ensuring widespread trustworthiness in LLMs across all their potential use cases is a ``game of fairness whack-a-mole [that] seems indefinitely intractable" \cite{Anthis_Lum_Ekstrand_Feller_DAmour_Tan_2024}. Even if a data scientist were able to reliably encourage some LLM output to be trustworthy, care is required to ensure any subsequent model trained on that data is also trustworthy. While this report aims to outline potential approaches to evaluate and enable the trustworthiness of large pretrained models, it is important for data scientists to remember that the use of generative AI comes with inherent risks that may limit the possible trustworthiness of the AI-enabled system being designed. 

\subsection{Generative model selection and trustworthiness measurements}
\label{llm_trustworthiness_measurements}

Data scientists often have many pretrained models to choose from for a given task. While many factors (e.g., resource requirements, task, proprietary restrictions, cost) influence this choice, a data scientist should initially determine the model output's performance in a relevant downstream task given selected characteristics. In the context of trustworthiness, the goal of this experimentation is to determine which candidate pretrained model will have the most trustworthy (i.e., optimized for the true distribution of inputs) output. 

LLMs often refer to causal language models (CLMs) with many parameters that are trained on a large amount of data. Causal language modeling is the task of predicting the next token given previous context. These models are often referred to as autoregressive, meaning they use their past output as future input for the next step in the series. LLMs can sometimes refer to large masked language models (MLMs) (e.g., BERT), which are trained on the task of predicting words that are masked (hidden to the model at prediction time). Among the different model training procedures, one procedure might be a better fit for a given task. Tasks such as named entity recognition, part of speech tagging, or semantic similarity might benefit more from the MLM training procedure because of the available forward context, whereas tasks that involve text generation would be more suited to the CLM training procedure. Thus, data scientists should consider training procedures before selecting a models for their task. 

One flexible way to assess candidate models is to devise a simple classification task using the classes and selected characteristic of interest. For example, if the input to the downstream model is a generated description of the image of a ship, the data scientist may want to generate image descriptions and classify them into civilian or military vessels (e.g., using debiased word embedding models). Those class predictions could be taken into account while balanced performance across a selected characteristic (e.g., images that do and do not contain the coastline) is ensured.  

LLMs are often trained on large collections of potentially disparate datasets. These models' performance can therefore vary greatly depending on the task. A data scientist may want a model that is robust and trustworthy at summarization, for example, even if the mathematical reasoning capabilities are far worse than alternative candidate models. Chang et al. \cite{Chang_Wang_Wang_Wu_Yang_Zhu_Chen_Yi_Wang_Wang} provide a list of relevant benchmarks and datasets for multiple tasks. For online models that experience retraining over time, an open area of research is to continuously track which model is the best in light of these updates \cite{Xia_Kong_Yu_Guo_Rossi_Kim_Li_2024}. While there is no single source comparing all closed- and open-source LLMs for all tasks, there are incomplete open-source leaderboards for a wide variety of tasks \cite{Muennighoff_Tazi_Magne_Reimers_2023, liu2023is, DBLP:conf/icml/ChiangZ0ALLZ0JG24}\footnote{Massive Text Embedding Benchmark Leaderboard: \href{https://huggingface.co/spaces/mteb/leaderboard}{https://huggingface.co/spaces/mteb/leaderboard}\\
EvalPlus Leaderboard (for code generation): \href{https://evalplus.github.io/leaderboard.html}{https://evalplus.github.io/leaderboard.html}\\
Chatbot Arena LLM Leaderboard: \href{https://lmarena.ai/?leaderboard}{https://lmarena.ai/?leaderboard}}, and they can help a data scientist select models with which to experiment.

While this report is typically concerned with bias as it relates to the more general concept of mischaracterization (Chapter~\ref{mischar}), the more traditional framing of bias in model outputs can be one useful criterion for selecting between pretrained models. However, the field of bias measurement and detection incldudes many different bias and meta metrics. Bias metrics are usually formulated from a base metric and measure the amount of bias that each group experiences in a given model. Meta metrics measure the amount of bias in a model as a whole \cite{Lum_Zhang_Bower_2022}. Unfortunately, there is no best one-size-fits-all base or meta metric for evaluating bias in large pretrained models. Furthermore, as these metrics build on each other as the scope is widened (model performance versus group parity versus model fairness), differences in base metrics can become compounded and can have large downstream effects on meta metrics. Hutiri et al. \cite{Hutiri_Patel_Ding_Scharenborg_2024} even demonstrate that choice of metrics can lead to differing if not opposite conclusions about a given model. This research compares different group bias metrics (group-to-min difference, group-to-average ratio, and group-to-average log ratio) as well as two meta metrics -- fairness discrepancy rate (FDR) and normalized reliability bias (NRB) -- that are based on these bias metrics. They find that ratio-based bias metrics lead to the most reliable meta measures when base metrics are small or have different orders of magnitude. In general, to select the best model for the task at hand, the recommendation is to use and understand multiple metrics when quantifying bias of different generative models.

\subsection{Prompt engineering for trustworthy generation}
\label{sec:prompt_engineering}
Prompt engineering is the process of iteratively testing different generative model inputs, or \emph{prompts}, to determine the most effective input for a given task. Generative models can be very sensitive to changes in their prompts -- small changes can yield significantly different outputs \cite{Chen_Zhang_Langrene_Zhu_2024, Marvin_Hellen_Jjingo_Nakatumba-Nabende_2024} -- so prompt engineering can help data scientists select among candidate models and can guide outputs toward performance on the true distribution of inputs. Prompt engineering can involve changing wording or punctuation, adding examples, or incorporating contextual information. It is a cost-effective and relatively low effort modification to any generation pipeline and is therefore favored by many users to improve their model's output for their task. 

Prompt engineering is often described as the ``ability of GenAIs to learn skills and tasks by providing them with exemplars and or relevant instructions within the prompt, without the need for weight updates/retraining" \cite{DBLP:journals/corr/abs-2406-06608}. While it is true that editing the prompt might improve the quality of the output, prompt engineering by nature does not involve changing the weights of a model, nor does it add any ``guardrail" to prevent hallucinatory output. Rather, prompt engineering should be seen as ``nudging" the model in the direction of the desired output rather than enforcing that the output will be of a certain form or level of accuracy. 

A common form of prompt engineering follows the paradigm of in-context learning (ICL), where a model receives an enhanced prompt that usually contains examples analogous to the desired output \cite{ICL}. The autoregressive model uses its own output as input and can therefore pull information from the prompt rather than from the training data (as encoded in pretrained weights). Methodology such as chain of thought (CoT) prompt engineering -- encouraging the model to produce step-by-step explanations for its output -- has been shown to significantly improve the quality of the output of logical reasoning tasks for the same reason. Since a generative model uses its own output as input, this long-form ``logical" explanation can help drive the model toward a more consistent output \cite{zhang2023automatic}. 

Dwivedi et al. \cite{prompt_generation_fairness} experimented with both debiased prompts with and without in-context examples, and they found that while they were able to debias output by around 40\%, too much of this ``sanitized" output can lead to incoherence. The task of prompt engineering can be quite fickle, as models can be responsive to concepts such as word choice and example order \cite{Zhao_Wallace_Feng_Klein_Singh_2021}. Additionally, it has been found that language models do not respond to prompts in the same way as humans do. This makes prompt engineering more of a trial-and-error task than a task that encourages the language models to act in a more human-like manner \cite{Webson_Pavlick_2022}. Note also that prompt engineering will be limited by the context window of the model. As models increase in size, so do context windows, but there is always a limit to the amount of data that one can fit into any prompt. Therefore, the choice, amount, ordering, and format of examples included in the prompt are all important design decisions in prompt engineering \cite{DBLP:journals/corr/abs-2406-06608}.

\subsection{Trustworthy retrieval augmented generation}
RAG is used to achieve consistency in generative model output, mitigating against issues such as hallucinations or the limitations of training data (e.g., a model is used on a new domain). RAG involves searching in a database for relevant media (e.g., text, images, audio) to provide to the model with context. This enables to model to respond to a query (in a prompt) with more relevant information than may be latent in its weights after training \cite{Gao_Xiong_Gao_Jia_Pan_Bi_Dai_Sun_Wang_Wang_2024}. The goal of a RAG system is to encourage the model toward desirable outputs (e.g., optimizing for performance on the true distribution of inputs), which may improve the the trustworthiness of the AI-enabled system as a whole. However, the added complexity may inadvertently present opportunities for unintended bias. 

RAG is centered on the concept of information retrieval (IR): the task of identifying and ranking relevant pieces of information, given some query or topic of interest. In some cases, the exact order of the ranking is important to the downstream ML task (note that models are sensitive to example ordering in the prompt. as discussed in Section \ref{sec:prompt_engineering}), and in others, IR simply involves returning the top $k$ documents regardless of order. In RAG pipelines, $k$ relevant documents are retrieved such that (i) documents are incorporated into $k$ individual prompts and model outputs are combined or (ii) documents are concatenated into one large prompt. Order is especially impactful in this latter case, particularly when the length of the model's context window may require input truncation. 

Embedding spaces can contain bias (Section~\ref{perspective/embedding}), and if an IR system leverages biased embeddings, that bias may propagate into the ranked results. Much like the general approaches described earlier, the first step in creating a trustworthy RAG pipeline is to evaluate the bias inherent in the rankings based on selected characteristics. While IR systems often incorporate embeddings (and, thus, the methods described in Section \ref{perspective/embedding} are directly applicable), sometimes the algorithms are more naive. In the case where there are no embeddings, the goal of IR evaluation is to look at candidates that have similar features but belong to different classes along the selected characteristic. Fair IR would lead those data points to have similar rankings, otherwise called individual fairness \cite{Lahoti_Gummadi_Weikum_2019}. Additionally, a fair IR system would also ensure that within a selected characteristic, members of certain groups  do not have significantly lower rankings as a whole in comparison to members of other groups. 

Because of a wealth of data or other processing limitations, it is common that the top $k$ items returned by any IR algorithm do not represent all of the valid options for a given query. Unfair and untrustworthy IR would yield a ranking such that with a given $k$, a disproportionate amount of high-scoring members of a sensitive class fall above or below the threshold $k$. Therefore, one goal of trustworthy IR is to provide \emph{equal exposure} to high-scoring members of groups within a selected attribute \cite{Singh_Joachims_2018}. Zehlike et al. \cite{Zehlike_Bonchi_Castillo_Hajian_Megahed_Baeza_Yates_2017} outline three fair ranking criteria that algorithms should adhere to selecting a permutation $\mathcal{T}$ such that $\mathcal{T}$ exhibits:
\begin{enumerate}
    \item Ranked group fairness: $\mathcal{T}$ should fairly represent the protected group.
    \item Selection utility: $\mathcal{T}$ should contain the most qualified candidates.
    \item Ordering utility: $\mathcal{T}$ should be ordered by decreasing qualifications.
\end{enumerate}

Some fair ranking methods attempt to create trustworthy lists by incorporating fairness constraints in the ranking algorithms or reranking after prediction \cite{Gale_Marian_2024, Zehlike_Castillo_2020, Zehlike_Bonchi_Castillo_Hajian_Megahed_Baeza_Yates_2017}.\footnote{https://github.com/fair-search/\\
https://github.com/plahoti-lgtm/iFair} Other methods learn fair representations of each item being ranked, which means that these same representations can be used for other tasks, such as clustering \cite{Lahoti_Gummadi_Weikum_2019} 

Many of these methods are created with discrete, human-interpretable features in mind. Since ranking text documents for RAG involves transforming the documents into embeddings, text document IR can be less interpretable than IR on simpler data. Zhou et al. \cite{Zhou_Liu_Li_Jin_Qian_Liu_Li_Dou_Ho_Yu_2024} define trustworthiness in RAG systems across six dimensions: factuality, transparency, robustness, fairness, privacy, and accountability. These six pillars of RAG trustworthiness maximize the reliability and traceability of any data that is generated by a large model and therefore lead to a more trustworthy downstream model. All of these dimensions aim to encourage the generated output to adhere to the true distribution -- one that performs equally across a selected characteristic and is traceable and interpretable to the source information. Though there is not a well-known RAG system that actively uses fair IR, Kim and Diaz \cite{Kim_Diaz_2024} experimented with a stochastic retrieval method (over a deterministic ranker) and found that it consistently improves retrieval performance given a selected characteristic. RAG is a new and open area of natural language processing research, and as RAG systems become more complex, they have rapidly improved across all six of Zhou et al.'s trustworthiness dimensions. 

\subsection{Adapt foundation models for out-of-distribution tasks}

Although many state-of-the-art foundation models are closed-source, some powerful models remain open-source and available for fine-tuning or adaptation. Unless the deployed environment is the open web, it is likely that the task for which an AI-enabled system is being designed will contains out-of-distribution (OOD) inputs relative to the foundation model's training data. The topic of adapting foundation models to OOD tasks is a developing area of research, but two common techniques are fine-tuning and alignment.

\subsubsection{Fine-tuning and pretraining foundation models}
\label{sec:finetune}
If there is sufficient representative data from the deployed environment (e.g., option (b) in decision 1 from Section~\ref{sec:statistical_tools/decision_1}), foundation models can be adapted to generated data that is more representative of the deployed environment. This adaptation can be in two forms: pretraining and fine-tuning. 

Pretraining is the initial self-supervised step in model training where weights are learned from a large corpus and labels are generated from the data (e.g., as in CLMs or MLMs) \cite{Mao_2020}. Adapting a model via pretraining is a computationally expensive process that involves adding task-relevant data to the initial training corpus and training from the combined corpora. There are cases where the model's task has remained the same but the data scientist wishes to expand the domain of the input in some dimension. For example, consider a model trained on English that is pretrained to operate over inputs in Spanish as well. 

Fine-tuning is the process by which a foundation model that is trained on one task adapts a subset of the model's layers toward a new, relevant, task \cite{Vrbancic_Podgorelec_2020}. Foundation models are by nature generalized, and it has been found that model adaptation will enhance performance on individual tasks \cite{Gururangan_Marasovic_Swayamdipta_Lo_Beltagy_Downey_Smith_2020}. Fine-tuning is thus a practical step that data scientists often use to improve output performance. 

One common issue with fine-tuning is called catastrophic forgetting, a process in which the model learns the new task but its performance on previous tasks degrades significantly \cite{Kar_Castellucci_Filice_Malmasi_Rokhlenko_2022}. Becuase of this limitation, fine-tuning is not always an improvement over the base model, and care must be taken to ensure performance as the model is adapted to the new task. 

As the size of models has increased, the need for computationally efficient fine-tuning methods has increased correspondingly. One popular set of efficient fine-tuning methods is parameter-efficient fine-tuning (PEFT) \cite{houlsby2019parameter}, which selects parameters for fine-tuning. Low-ranked adaptation (LoRA), for example, is a common PEFT method for reparameterization. Many fine-tuning methods have implementations for publicly available models.\footnote{ \href{https://huggingface.co/docs/peft/en/index}{PEFT (https://huggingface.co/docs/peft/en/index)} and  \href{https://huggingface.co/docs/peft/main/en/conceptual_guides/lora}{LoRA (https://huggingface.co/docs/peft/main/en/conceptual\_guides/lora)}}

\subsubsection{Fine-tuning with pseudo-labels}
Fine-tuning methods often require labeled data from the deployed environment, but data is not readily available in many settings. It is important to emphasize that without representative data, a model -- even a large foundation model -- cannot be considered ``trustworthy" according to the actionable definition. It is possible, however, to create pseudo-labels that attempt to mimic the deployed environment as much as possible. In environments where labeled data is scarce, data scientists can leverage semi-supervised learning methods to generate data to train their models \cite{Yang_Song_King_Xu_2023}. Pseudo-labeling techniques often involve using pretrained models or models trained on a small amount of labeled data to obtain appropriate labels for the unlabeled training data. Many pseudo-labeling methods learn labels using a model's confidence metrics -- this means that any biases within the model will be propagated downstream into the pseudo-labels. This process is referred to as confirmation bias, specifically: ``confirmation bias is overfitting to incorrect pseudo-labels predicted by the network" \cite{Arazo_Ortego_Albert_OConnor_McGuinness_2020}. There are many regularization techniques to correct for confirmation bias, including requiring a minimum number of true labels per batch, ``mixup augmentation" (weighted random combination of previous samples), or other data synthesis methods described in Section \ref{sec:resampling}.

\subsubsection{Domain generalization, projection, and alignment}
\label{sec:projection}
There may be cases in which a data scientist wants to shift the distribution of a foundation model (e.g. image embeddings to text, embeddings in one language to another). If the data scientist has a reasonable amount of data from the deployed environment but still not enough to train or even successfully fine-tune a large model, the data scientist could consider training a projection layer in order to shift the distribution of the input data.

A typical solution to the domain shift problem is to use importance weighting to ``weight the samples in the source domain based on the ratio of target and source domain densities" \cite{Farahani_Voghoei_Rasheed_Arabnia_2021}. More on importance weighting can be found in Section ~\ref{sec:reweight/importance} \cite{Farahani_Voghoei_Rasheed_Arabnia_2021}. One ``shallow" way of aligning two distributions is minimizing the distances between them. To do this, the data scientist would need access to data from both distributions. The most-used distance measures in the domain adaptation are maximum mean discrepancy \cite{Yan_2017_CVPR}, Wasserstein metric \cite{Shen_Qu_Zhang_Yu_2018}, correlation alignment \cite{coral, coral-lda}, KL divergence \cite{nguyen2022kl}, and contrastive domain discrepancy \cite{Kang_2019_CVPR}. Another way to utilize these distance metrics is to train a neural network as a layer on top of the foundation model and using distribution distance measures as the loss function. A common method of using OOD input within a foundation model is to train a projection layer using the expected outputs  \cite{Lester_Al-Rfou_Constant_2021, Kang_Roy_2024}. For example, to feed audio input into a text-based foundation model without a speech-to-text step, one can create audio ``token embeddings such that the LLM’s response to a spoken prompt matches its response to a text prompt with the same semantic content" \cite{Kang_Roy_2024}. The networks needed to create a foundation model-interpretable prompt take far less data to train than a foundation model would need to fine-tune; however, this method does not affect the output distribution of the foundation model. 

Pretrained models are powerful tools that can be useful to a data scientist when curating data. However, using such a tool requires knowledge of how exactly that tool is going to transform the data at hand. Any pretrained model trained on data that is not sampled from the deployed environment is inherently not trustworthy as defined by this report. Minimally, such models should be tested to quantify the difference between the output distribution and the true distribution. Ensuring trustworthy output from such models is a growing field of research, and there are many adaptations that a data scientist can employ to shift the output of the model to match the true distribution of the deployed environment as much as possible. 

\section{Practitioner's perspective on using pretrained models}
\label{perspective/practitioners}

\subsection{Low-dimensional image representations from pretrained ResNet}
\label{subsec:practitioners_perspective_resnet}

The use of pretrained embeddings in AI-enabled systems is an important concept to consider when taking data curation actions to enable trustworthy AI. Consider an object detection scenario where a data point consists of an image and metadata about the image. A technique like splitting (Chapter~\ref{sec:splitting}) could divide the data based on image metadata, balancing the average size of object bounding boxes between splits. However, two images can look very different but have the same metadata. If we want to include information about the actual image itself (i.e., the colored pixels that make up the image's visual representation) in splitting, we can utilize a pretrained embedding space to obtain numerical embeddings for every image. These embeddings can then be used as features for splitting, to preserve diversity of the images between data splits. This can be thought of as taking advantage of the semantic space of embeddings; in this way, splits can occur on (fuzzy, machine-interpretable) concepts like \emph{patch-of-redish-pixels-around-here} or \emph{edges-in-a-certain-configuration}.

To demonstrate this concept, we obtained image embeddings for an object detection dataset using ResNet \cite{he2016deep}. ResNet is an image classification model, but the vector of final hidden state values before classification output can be used as an image embedding. To use these embeddings with splitting techniques (which typically require lower dimensionality), we reduce the length-2048 vectors from ResNet to length 20 using principal component analysis. Splitting was then performed using these vectors as features to ensure that the diversity of the original dataset was preserved among the splits. While there is no quantitative method for verifying diversity preservation, the splits resulting from the use of pretrained embeddings as features can be qualitatively compared to the splits resulting from using only the metadata as features. Qualitative assessment can show the preservation of diversity between splits.

Using pretrained embeddings as features, however, comes with assumptions and risks. The human interpretability of the splits decreases as embeddings are used, and the process assumes the semantic space of the embeddings has meaning that will be useful in data curation, which may or may not be true depending on how the model was trained. While these assumptions and risks are important to acknowledge, they do not preclude the use of pretrained embeddings as features, and the benefits of using pretrained embeddings as a feature should be weighed against the downsides.

Chapter~\ref{sec:pretrained} provides an in-depth overview of the data curation process when using pretrained models.

\subsection{Debiasing Word2Vec embeddings with WEFE}
\label{subsec:practitioners_perspective_wefe}

Pretrained embeddings can encode latent biases from their training data (Section~\ref{sec:measure_we_bias}), and mitigating these biases can be an important part of data curation to support trustworthy AI. To demonstrate debiasing in practice, we use the WEFE toolkit \cite{Badilla_Bravo_Marquez_Perez_2020} to identify and mitigate dialectal bias (American vs. British spellings) in domain-specific word embeddings.

\paragraph{Data.}
Rather than using existing, general-purpose word embeddings, we train a Word2Vec model \cite{Mikolov_Chen_Corrado_Dean_2013} on the BioCreative V Chemical Disease Relation corpus (bc5cdr) \cite{li2016biocreative}. It is expected that these embeddings capture domain-specific language while minimizing the introduction of unexpected external biases (e.g., as may occur with text from the open web). The bc5cdr dataset splits were combined prior to debiasing activities to ensure consistency across subsets.

\paragraph{Bias detection and mitigation.}
The WEAT metric was used to measure dialectal bias in the embeddings by comparing attribute sets (e.g., ``color'' versus ``colour'') and target sets (e.g., ``chemistry'' versus ``biology''). When measured, the model yielded a WEAT score of 0.357, indicating associations along the dialectal axis. To mitigate this bias, WEFE's HardDebias function geometrically adjusts the embedding space. This function requires \emph{definitional pairs} (e.g., ``color'', ``colour'') to define the bias axis, and \emph{equalizer pairs} (e.g., ``chemistry'', ``biology'') to ensure that unrelated concepts are not affected by the procedure. After debiasing with this function, the WEAT score was 0.117, indicating a meaningful reduction in bias along the dialectal axis. 

\paragraph{Additional considerations.}
While WEFE is a powerful tool, it assumes that biases in embeddings can be geometrically defined and mitigated. This assumption may not hold in all scenarios, and care should be taken to determine whether a tool like WEFE accomplishes the goal of reducing embedding bias. In cases where WEFE is appropriate, practitioners should consider several key items.

\emph{Selecting definitional and equalizer pairs.} \hspace{1em} 
The quality of debiasing depends heavily on the accuracy and representativeness of these pairs. Poorly chosen pairs may fail to capture the intended bias or introduce unintended distortions. It is therefore important to understand the dataset’s context and the type of bias being addressed. Where possible, it can be helpful to collaborate with SMEs to define meaningful pairs. 

\emph{Interpreting WEAT results.} \hspace{1em} 
While WEAT provides a quantifiable metric, interpreting its scores requires understanding the relationships between attribute and target sets. It can be helpful to use absolute WEAT scores for comparison (e.g., biased versus debiased) and focus on the direction and magnitude of changes to assess the effectiveness of debiasing.

\emph{Debiasing the entire dataset.} \hspace{1em}
Recombining subsets for debiasing ensures consistency. After debiasing, however, the data should be re-split into training and validation sets.

\emph{Focusing on domain-specific data.} \hspace{1em}
Training embeddings specific to the dataset ensures better alignment with the task and reduces the risk of introducing external biases from general-purpose models.
    
\emph{Iteratively testing.} \hspace{1em}
Definitional and equalizer pairs should be tested iteratively. Subtle adjustments can significantly impact the effectiveness of debiasing.

\emph{Evaluating beyond WEAT.} \hspace{1em}
While WEAT is a robust tool for measuring bias, consider testing embeddings in downstream tasks to validate their practical fairness and performance.

The WEFE toolkit demonstrates the balance between precision and complexity in data curation. For practitioners, it offers a structured framework for addressing bias in embedding spaces, but success relies on careful planning, understanding of the dataset, and iterative refinement. By leveraging tools like WEFE, practitioners can create embeddings that are not only fairer but also better suited for real-world applications.

\chapter{Conclusion}
\label{sec:conclusion}
This report aims to support stakeholders pursuing AI trustworthiness by offering thoughtful and practical guidance specific to the data curation phase of the AI development lifecycle. We contend that data curation is an underutilized opportunity to promote trustworthiness, where trustworthiness entails optimizing an ML model for performance on the true distribution of inputs in the deployed environment.

Though this report includes necessary conceptual frameworks synthesized from academic literature (e.g., what is meant by data or trustworthiness), the majority of the report collects, collates, organizes, compares, and offers practical guidance on data curation tools and techniques. These range from decades-old statistical approaches to present-day generative AI research. We highlighted individual capabilities in sprawling, well-known open-source toolkits and uncovered obscure but elegant one-off tools that solve key challenges. We pieced together a puzzle of techniques that address myriad data curation scenarios, especially in the national security domain where directly relevant datasets are scarce and reusing data is a matter of necessity.

In one particular area where a gap made it impossible to stitch together a coherent process, we created a new tool: twelve essential questions to elicit domain knowledge. These questions elicit why the AI-enabled system is being built (i.e., the task with which AI is to assist), how the data scientist can make nuanced decisions about data transformations, and the where the AI-enabled system will be deployed (i.e., discrepancies between the available dataset and the true distribution). 

There are more data curation techniques than could fit in this report, and new ones, especially for use with pretrained models, are proposed every day. To complement the blossoming of new techniques, it is important that existing approaches mature into well-known, dependable, and smoothly integrable tools. This report highlights disparate tools that, in some cases, are difficult to find and to use together. We recommend prioritizing these tools for integration because they form a coherent workflow, with well-defined alternatives for distinct project circumstances. This report constructs such a workflow.

We see value in taking a similar lens to other phases or subphases of the AI lifecycle, such as synthetic data generation, metrics selection, model training, and evaluation.

\paragraph{Synthetic data generation.} When an insufficient volume or diversity of data can be collected, developers generate data synthetically. We addressed a few of these techniques in Section~\ref{sec:resampling}, but there are many more. Of particular interest are simulators or data generation templates that, while inspired by real observations, are entirely fabricated. Autonomous robots, for instance, often have models trained on data generated in physical simulations; the ``reality gap'' refers to the wide and problematic chasm between simulations and the physical world. As stated in our actionable definition of trustworthiness, a trustworthy AI-enabled system ultimately must be optimized to perform in the deployed environment, not a simulated or synthetic one. As a future direction, researchers could determine how and in what contexts synthetic data generation techniques succeed in approximating the real-world deployed environment.

\paragraph{Metrics selection.} Machine learning models are trained by optimizing a loss function, which determines the notion of ``good performance'' the training algorithm is pursuing. We refer to the loss function (and validation metrics) in our actionable definition with the general term ``performance.'' A common loss function is the average accuracy on the training dataset; a ``good'' model is one that gets as many correct answers on the training set as possible. However, many circumstances can (i) complicate the notion of a ``correct answer'' or (ii) prioritize correct answers on a diversity of training data points rather than simply a large number of them. This report partially addresses the latter issue by curating training and validation sets that approximately represent the true distribution. The former issue, however, is not addressed in this report. For example, a ``correct answer'' for a document summarization system is difficult to define; different summaries of the same document can use different language but be equally useful and coherent. In this and other cases, we want a loss function to measure how ``incorrect'' the model's answer is so that we can train the model to output better answers. Future work could catalog loss functions, their strengths, weaknesses, and appropriate contexts. 

\paragraph{Model training.} Models are typically trained using iterative optimization algorithms, like gradient descent, to adjust the parameters in a way that reduces the loss function's penalty for ``incorrect answers.'' The methods of performing the optimization originate from a blend of theoretical mathematics and experimental computer science. Sometimes, theoreticians come up with promising algorithms that are difficult to implement in practice due to numerical instability or other issues, and sometimes empiricists develop tactics that provide measurable benefits across a variety of problems but lack theory to explain why they work or what specific kinds of problems they help with. For the purposes of trustworthy AI, it would be best to have both theory and practice justifying a choice of training algorithm. A future work could explore training algorithms and adaptations that are well-justified for use on wide classes of models.

\paragraph{Evaluation.} After using a loss function for training and a validation metric for model selection, developers turn the trained model over for testing and evaluation. Comprehensive testing may produce many measurements; evaluators are responsible for synthesizing together the measurements into a recommendation for whether to acquire and deploy the model or to send the model back to developers for further refinement. Future research could determine which measurements are important in which contexts, and can explore how evaluators can reconcile evaluation metrics into a clear picture of a model's trustworthiness.

Each phase of the AI lifecycle can contribute to the trustworthiness of the AI-enabled system. The focus of this report is the data curation phase: an oft-overlooked phase which we found to be a rich opportunity for promoting trustworthiness. We anticipate a similar wealth of opportunities elsewhere in the AI lifecycle, and contend that practical guidance for these phases is necessary.

As AI continues to be developed and deployed in government and industry, it is important for researchers to continue creating and synthesizing tools and techniques to support trustworthy AI, developers to integrate trustworthy AI principles into practice, and all stakeholders to consider trustworthiness as a necessary part of any AI-enabled system.

\chapter*{Acknowledgements}
Copyright 2025 Carnegie Mellon University and JHU APL

This material is based upon work funded and supported by the Department of Defense under Contract No. FA8702-15-D-0002 with Carnegie Mellon University for the operation of the Software Engineering Institute, a federally funded research and development center. 

The view, opinions, and/or findings contained in this material are those of the author(s) and should not be construed as an official Government position, policy, or decision, unless designated by other documentation.

[DISTRIBUTION STATEMENT A] This material has been approved for public release and unlimited distribution.  Please see Copyright notice for non-US Government use and distribution.

This work is licensed under a Creative Commons Attribution-NonCommercial 4.0 International License.  Requests for permission for non-licensed uses should be directed to the Software Engineering Institute at \url{permission@sei.cmu.edu}.

Carnegie Mellon® is registered in the U.S. Patent and Trademark Office by Carnegie Mellon University.

DM25-0327

\bibliographystyle{plainurl}
\bibliography{0_cate_dcv}

\begin{thebibliography}{100}

\bibitem{noauthor_c2pa_nodate}
{C2PA} {Implementation} {Guidance} :: {C2PA} {Specifications}.
\newblock URL:
  \url{https://c2pa.org/specifications/specifications/1.3/guidance/Guidance.html#1.4@specs:C2PA_Specification.adoc}.

\bibitem{noauthor_data&trust_alliance_nodate}
The {Data} \& {Trust} {Alliance}.
\newblock URL: \url{https://dataandtrustalliance.org/}.

\bibitem{noauthor_data_nodate}
Data types and semantic types {\textbar} {Community} {Connectors}.
\newblock URL:
  \url{https://developers.google.com/looker-studio/connector/semantics}.

\bibitem{Zehlike_Bonchi_Castillo_Hajian_Megahed_Baeza_Yates_2017}
Fa*ir: A fair top-k ranking algorithm.
\newblock URL: \url{http://arxiv.org/abs/1706.06368}, \href
  {https://doi.org/10.1145/3132847.3132938}
  {\path{doi:10.1145/3132847.3132938}}.

\bibitem{Yang_Jin_Tang_Han_Feng_Jiang_Zhong_Yin_Hu_2024}
Harnessing the power of llms in practice: A survey on chatgpt and beyond.
\newblock 18.
\newblock URL: \url{https://dl.acm.org/doi/10.1145/3649506}, \href
  {https://doi.org/10.1145/3649506} {\path{doi:10.1145/3649506}}.

\bibitem{noauthor_project_nodate}
Project {Overview} - {Data} {Provenance} for {AI}.
\newblock URL:
  \url{https://www.media.mit.edu/projects/data-provenance-for-ai/overview/}.

\bibitem{noauthor_prov-o_nodate}
{PROV}-{O}: {The} {PROV} {Ontology}.
\newblock URL: \url{https://www.w3.org/TR/prov-o/}.

\bibitem{noauthor_rdflib_nodate}
rdflib 7.0.0 — rdflib 7.0.0 documentation.
\newblock URL: \url{https://rdflib.readthedocs.io/en/stable/#}.

\bibitem{sklearn-metrics}
Scikit-learn user guide — 3.4. metrics and scoring.
\newblock URL: \url{https://scikit-learn/stable/modules/model_evaluation.html}.

\bibitem{mayer_1995_integrative}
An integrative model of organizational trust.
\newblock {\em The Academy of Management Review}, 20(3):709--734, 1995.

\bibitem{ai_strategy_summary_2018}
{SUMMARY} {OF} {THE} 2018 department {OF} {DEFENSE} artificial {INTELLIGENCE}
  strategy.
\newblock Technical report, United States Department of Defense, 2018.
\newblock URL:
  \url{https://media.defense.gov/2019/Feb/12/2002088963/-1/-1/1/SUMMARY-OF-DOD-AI-STRATEGY.PDF}.

\bibitem{albertus2019auxiliary}
Mickael Albertus and Philippe Berthet.
\newblock Auxiliary information: the raking-ratio empirical process.
\newblock 2019.

\bibitem{an2020resampling}
Jing An, Lexing Ying, and Yuhua Zhu.
\newblock Why resampling outperforms reweighting for correcting sampling bias
  with stochastic gradients.
\newblock {\em arXiv preprint arXiv:2009.13447}, 2020.

\bibitem{anand2010approach}
Ashish Anand, Ganesan Pugalenthi, Gary~B Fogel, and PN~Suganthan.
\newblock An approach for classification of highly imbalanced data using
  weighting and undersampling.
\newblock {\em Amino acids}, 39:1385--1391, 2010.

\bibitem{Anthis_Lum_Ekstrand_Feller_DAmour_Tan_2024}
Jacy Anthis, Kristian Lum, Michael Ekstrand, Avi Feller, Alexander D’Amour,
  and Chenhao Tan.
\newblock The impossibility of fair llms.
\newblock (arXiv:2406.03198), May 2024.
\newblock arXiv:2406.03198 [cs, stat].
\newblock URL: \url{http://arxiv.org/abs/2406.03198}.

\bibitem{Arazo_Ortego_Albert_OConnor_McGuinness_2020}
Eric Arazo, Diego Ortego, Paul Albert, Noel~E. O’Connor, and Kevin
  McGuinness.
\newblock Pseudo-labeling and confirmation bias in deep semi-supervised
  learning.
\newblock In {\em 2020 International Joint Conference on Neural Networks
  (IJCNN)}, page 1–8, Glasgow, United Kingdom, 2020. IEEE.
\newblock URL: \url{https://ieeexplore.ieee.org/document/9207304/}, \href
  {https://doi.org/10.1109/IJCNN48605.2020.9207304}
  {\path{doi:10.1109/IJCNN48605.2020.9207304}}.

\bibitem{archer2014catalog}
Phil Archer.
\newblock Data catalog vocabulary (dcat) (w3c recommendation).
\newblock Online, January 2014.
\newblock URL: \url{https://www.w3.org/TR/vocab-dcat/}.

\bibitem{Badilla_Bravo_Marquez_Perez_2020}
Pablo Badilla, Felipe Bravo-Marquez, and Jorge Pérez.
\newblock Wefe: The word embeddings fairness evaluation framework.
\newblock In {\em Proceedings of the Twenty-Ninth International Joint
  Conference on Artificial Intelligence}, page 430–436, Yokohama, Japan,
  2020. International Joint Conferences on Artificial Intelligence
  Organization.
\newblock URL: \url{https://www.ijcai.org/proceedings/2020/60}, \href
  {https://doi.org/10.24963/ijcai.2020/60} {\path{doi:10.24963/ijcai.2020/60}}.

\bibitem{rsw}
Shane Barratt, Guillermo Angeris, and Stephen Boyd.
\newblock Optimal representative sample weighting.
\newblock {\em Statistics and Computing}, 31(2):19, March 2021.
\newblock \href {https://doi.org/10.1007/s11222-021-10001-1}
  {\path{doi:10.1007/s11222-021-10001-1}}.

\bibitem{battaglia2009practical}
Michael~P Battaglia, David~C Hoaglin, and Martin~R Frankel.
\newblock Practical considerations in raking survey data.
\newblock {\em Survey practice}, 2(5), 2009.

\bibitem{begtin_semantic_2022}
Ivan Begtin.
\newblock Semantic data types. {Systematic} approach and types registry, April
  2022.
\newblock URL:
  \url{https://medium.com/@ibegtin/semantic-data-types-systematic-approach-and-types-registry-a2c2a60a467b}.

\bibitem{aif360-oct-2018}
Rachel K.~E. Bellamy, Kuntal Dey, Michael Hind, Samuel~C. Hoffman, Stephanie
  Houde, Kalapriya Kannan, Pranay Lohia, Jacquelyn Martino, Sameep Mehta,
  Aleksandra Mojsilovic, Seema Nagar, Karthikeyan~Natesan Ramamurthy, John
  Richards, Diptikalyan Saha, Prasanna Sattigeri, Moninder Singh, Kush~R.
  Varshney, and Yunfeng Zhang.
\newblock {AI Fairness} 360: An extensible toolkit for detecting,
  understanding, and mitigating unwanted algorithmic bias, October 2018.
\newblock URL: \url{https://arxiv.org/abs/1810.01943}.

\bibitem{Blum_Stangl_2019}
Avrim Blum and Kevin Stangl.
\newblock Recovering from biased data: Can fairness constraints improve
  accuracy?
\newblock (arXiv:1912.01094), December 2019.
\newblock arXiv:1912.01094 [cs, stat].
\newblock URL: \url{http://arxiv.org/abs/1912.01094}, \href
  {https://doi.org/10.48550/arXiv.1912.01094}
  {\path{doi:10.48550/arXiv.1912.01094}}.

\bibitem{Bolukbasi_Chang_Zou_Saligrama_Kalai_2016}
Tolga Bolukbasi, Kai-Wei Chang, James Zou, Venkatesh Saligrama, and Adam Kalai.
\newblock Man is to computer programmer as woman is to homemaker? debiasing
  word embeddings.
\newblock (arXiv:1607.06520), July 2016.
\newblock arXiv:1607.06520 [cs, stat].
\newblock URL: \url{http://arxiv.org/abs/1607.06520}.

\bibitem{brick2003identifying}
J~Michael Brick, Jill Montaquila, and Shelley Roth.
\newblock Identifying problems with raking estimators.
\newblock In {\em annual meeting of the American Statistical Association, San
  Francisco, CA}, 2003.

\bibitem{Byrd2018WhatIT}
Jonathon Byrd and Zachary~Chase Lipton.
\newblock What is the effect of importance weighting in deep learning?
\newblock In {\em International Conference on Machine Learning}, 2018.
\newblock URL: \url{https://api.semanticscholar.org/CorpusID:83458523}.

\bibitem{Caliskan_Bryson_Narayanan_2017}
Aylin Caliskan, Joanna~J. Bryson, and Arvind Narayanan.
\newblock Semantics derived automatically from language corpora contain
  human-like biases.
\newblock {\em Science}, 356(6334):183–186, April 2017.
\newblock URL: \url{https://www.science.org/doi/10.1126/science.aal4230}, \href
  {https://doi.org/10.1126/science.aal4230}
  {\path{doi:10.1126/science.aal4230}}.

\bibitem{canal2024decision}
Gregory Canal, Vladimir Leung, Philip Sage, Eric Heim, I~Wang, et~al.
\newblock A decision-driven methodology for designing uncertainty-aware ai
  self-assessment.
\newblock {\em arXiv preprint arXiv:2408.01301}, 2024.

\bibitem{chai2022fairness}
Junyi Chai and Xiaoqian Wang.
\newblock Fairness with adaptive weights.
\newblock In {\em International Conference on Machine Learning}, pages
  2853--2866. PMLR, 2022.

\bibitem{Chang_Wang_Wang_Wu_Yang_Zhu_Chen_Yi_Wang_Wang}
Yupeng Chang, Xu~Wang, Jindong Wang, Yuan Wu, Linyi Yang, Kaijie Zhu, Hao Chen,
  Xiaoyuan Yi, Cunxiang Wang, Yidong Wang, Wei Ye, Yue Zhang, Yi~Chang,
  Philip~S. Yu, Qiang Yang, and Xing Xie.
\newblock A survey on evaluation of large language models.
\newblock {\em ACM Transactions on Intelligent Systems and Technology},
  15(3):39:1--39:45, March 2024.
\newblock URL: \url{https://dl.acm.org/doi/10.1145/3641289}, \href
  {https://doi.org/10.1145/3641289} {\path{doi:10.1145/3641289}}.

\bibitem{chawla_smote_2002}
N.~V. Chawla, K.~W. Bowyer, L.~O. Hall, and W.~P. Kegelmeyer.
\newblock {SMOTE}: {Synthetic} {Minority} {Over}-sampling {Technique}.
\newblock {\em Journal of Artificial Intelligence Research}, 16:321--357, June
  2002.
\newblock URL: \url{https://www.jair.org/index.php/jair/article/view/10302},
  \href {https://doi.org/10.1613/jair.953} {\path{doi:10.1613/jair.953}}.

\bibitem{Chen_Zhang_Langrene_Zhu_2024}
Banghao Chen, Zhaofeng Zhang, Nicolas Langrené, and Shengxin Zhu.
\newblock Unleashing the potential of prompt engineering in large language
  models: a comprehensive review.
\newblock (arXiv:2310.14735), September 2024.
\newblock arXiv:2310.14735.
\newblock URL: \url{http://arxiv.org/abs/2310.14735}, \href
  {https://doi.org/10.48550/arXiv.2310.14735}
  {\path{doi:10.48550/arXiv.2310.14735}}.

\bibitem{chen_oodanalyzer_2020}
Changjian Chen, Jun Yuan, Yafeng Lu, Yang Liu, Hang Su, Songtao Yuan, and
  Shixia Liu.
\newblock {OoDAnalyzer}: {Interactive} {Analysis} of {Out}-of-{Distribution}
  {Samples}, February 2020.
\newblock arXiv:2002.03103 [cs].
\newblock URL: \url{http://arxiv.org/abs/2002.03103}.

\bibitem{chen2016xgboost}
Tianqi Chen and Carlos Guestrin.
\newblock Xgboost: A scalable tree boosting system.
\newblock In {\em Proceedings of the 22nd acm sigkdd international conference
  on knowledge discovery and data mining}, pages 785--794, 2016.

\bibitem{chevallier_semantic_2023}
Marc Chevallier, Nicoleta Rogovschi, Faouzi Boufarès, and Nistor Grozavu.
\newblock Semantic {Type} {Detection} in {Tabular} {Data} via {Machine}
  {Learning} {Using} {Semi}-synthetic {Data}.
\newblock In Ajith Abraham, Thomas Hanne, Niketa Gandhi, Pooja
  Manghirmalani~Mishra, Anu Bajaj, and Patrick Siarry, editors, {\em
  Proceedings of the 14th {International} {Conference} on {Soft} {Computing}
  and {Pattern} {Recognition} ({SoCPaR} 2022)}, pages 110--119, Cham, 2023.
  Springer Nature Switzerland.

\bibitem{DBLP:conf/icml/ChiangZ0ALLZ0JG24}
Wei-Lin Chiang, Lianmin Zheng, Ying Sheng, Anastasios~Nikolas Angelopoulos,
  Tianle Li, Dacheng Li, Banghua Zhu, Hao Zhang, Michael~I. Jordan, Joseph~E.
  Gonzalez, and Ion Stoica.
\newblock Chatbot arena: An open platform for evaluating llms by human
  preference.
\newblock In {\em ICML}, 2024.
\newblock URL: \url{https://openreview.net/forum?id=3MW8GKNyzI}.

\bibitem{chicco2022tips}
Davide Chicco, Luca Oneto, and Erica Tavazzi.
\newblock Eleven quick tips for data cleaning and feature engineering.
\newblock {\em PLOS Computational Biology}, 18(12):1--21, 12 2022.
\newblock \href {https://doi.org/10.1371/journal.pcbi.1010718}
  {\path{doi:10.1371/journal.pcbi.1010718}}.

\bibitem{scipystats}
SciPy Community.
\newblock Statistical functions (scipy.stats) — scipy v1.13.0 manual.
\newblock URL:
  \url{https://docs.scipy.org/doc/scipy/reference/stats.html#module-scipy.stats}.

\bibitem{deming1940least}
W~Edwards Deming and Frederick~F Stephan.
\newblock On a least squares adjustment of a sampled frequency table when the
  expected marginal totals are known.
\newblock {\em The Annals of Mathematical Statistics}, 11(4):427--444, 1940.

\bibitem{deville1992calibration}
Jean-Claude Deville and Carl-Erik S{\"a}rndal.
\newblock Calibration estimators in survey sampling.
\newblock {\em Journal of the American statistical Association},
  87(418):376--382, 1992.

\bibitem{Devlin_Chang_Lee_Toutanova_2019}
Jacob Devlin, Ming-Wei Chang, Kenton Lee, and Kristina Toutanova.
\newblock Bert: Pre-training of deep bidirectional transformers for language
  understanding.
\newblock In Jill Burstein, Christy Doran, and Thamar Solorio, editors, {\em
  Proceedings of the 2019 Conference of the North American Chapter of the
  Association for Computational Linguistics: Human Language Technologies,
  Volume 1 (Long and Short Papers)}, page 4171–4186, Minneapolis, Minnesota,
  June 2019. Association for Computational Linguistics.
\newblock URL: \url{https://aclanthology.org/N19-1423}, \href
  {https://doi.org/10.18653/v1/N19-1423} {\path{doi:10.18653/v1/N19-1423}}.

\bibitem{dinan-etal-2020-multi}
Emily Dinan, Angela Fan, Ledell Wu, Jason Weston, Douwe Kiela, and Adina
  Williams.
\newblock Multi-dimensional gender bias classification.
\newblock In {\em Proceedings of the 2020 Conference on Empirical Methods in
  Natural Language Processing (EMNLP)}, pages 314--331, Online, November 2020.
  Association for Computational Linguistics.
\newblock URL: \url{https://www.aclweb.org/anthology/2020.emnlp-main.23}, \href
  {https://doi.org/10.18653/v1/2020.emnlp-main.23}
  {\path{doi:10.18653/v1/2020.emnlp-main.23}}.

\bibitem{ICL}
Qingxiu Dong, Lei Li, Damai Dai, Ce~Zheng, Jingyuan Ma, Rui Li, Heming Xia,
  Jingjing Xu, Zhiyong Wu, Baobao Chang, Xu~Sun, Lei Li, and Zhifang Sui.
\newblock A survey on in-context learning.
\newblock In {\em Proceedings of the 2024 Conference on Empirical Methods in
  Natural Language Processing}, page 1107–1128, Miami, Florida, USA, 2024.
  Association for Computational Linguistics.
\newblock URL: \url{https://aclanthology.org/2024.emnlp-main.64}, \href
  {https://doi.org/10.18653/v1/2024.emnlp-main.64}
  {\path{doi:10.18653/v1/2024.emnlp-main.64}}.

\bibitem{prompt_generation_fairness}
Satyam Dwivedi, Sanjukta Ghosh, and Shivam Dwivedi.
\newblock Breaking the bias: Gender fairness in llms using prompt engineering
  and in-context learning.
\newblock {\em Rupkatha Journal on Interdisciplinary Studies in Humanities},
  15(4), December 2023.
\newblock URL: \url{https://rupkatha.com/v15n410}, \href
  {https://doi.org/10.21659/rupkatha.v15n4.10}
  {\path{doi:10.21659/rupkatha.v15n4.10}}.

\bibitem{elor2022smote}
Yotam Elor and Hadar Averbuch-Elor.
\newblock To smote, or not to smote?
\newblock {\em arXiv preprint arXiv:2201.08528}, 2022.

\bibitem{Farahani_Voghoei_Rasheed_Arabnia_2021}
Abolfazl Farahani, Sahar Voghoei, Khaled Rasheed, and Hamid~R. Arabnia.
\newblock {\em A Brief Review of Domain Adaptation}, page 877–894.
\newblock Springer International Publishing, Cham, 2021.
\newblock URL: \url{https://link.springer.com/10.1007/978-3-030-71704-9_65},
  \href {https://doi.org/10.1007/978-3-030-71704-9_65}
  {\path{doi:10.1007/978-3-030-71704-9_65}}.

\bibitem{finn2017model}
Chelsea Finn, Pieter Abbeel, and Sergey Levine.
\newblock Model-agnostic meta-learning for fast adaptation of deep networks.
\newblock In {\em International conference on machine learning}, pages
  1126--1135. PMLR, 2017.

\bibitem{frenay2013classification}
Beno{\^\i}t Fr{\'e}nay and Michel Verleysen.
\newblock Classification in the presence of label noise: a survey.
\newblock {\em IEEE transactions on neural networks and learning systems},
  25(5):845--869, 2013.

\bibitem{Friedrich_Lauscher_Ponzetto_Glavas_2021}
Niklas Friedrich, Anne Lauscher, Simone~Paolo Ponzetto, and Goran Glavaš.
\newblock Debie: A platform for implicit and explicit debiasing of word
  embedding spaces.
\newblock (arXiv:2103.06598), March 2021.
\newblock arXiv:2103.06598 [cs].
\newblock URL: \url{http://arxiv.org/abs/2103.06598}.

\bibitem{Gale_Marian_2024}
Abraham Gale and Amélie Marian.
\newblock Explainable disparity compensation for efficient fair ranking.
\newblock In {\em 2024 IEEE 40th International Conference on Data Engineering
  (ICDE)}, page 2192–2204, Utrecht, Netherlands, May 2024. IEEE.
\newblock URL: \url{https://ieeexplore.ieee.org/document/10597822/}, \href
  {https://doi.org/10.1109/ICDE60146.2024.00174}
  {\path{doi:10.1109/ICDE60146.2024.00174}}.

\bibitem{Gao_Xiong_Gao_Jia_Pan_Bi_Dai_Sun_Wang_Wang_2024}
Yunfan Gao, Yun Xiong, Xinyu Gao, Kangxiang Jia, Jinliu Pan, Yuxi Bi, Yi~Dai,
  Jiawei Sun, Meng Wang, and Haofen Wang.
\newblock Retrieval-augmented generation for large language models: A survey.
\newblock (arXiv:2312.10997), March 2024.
\newblock arXiv:2312.10997.
\newblock URL: \url{http://arxiv.org/abs/2312.10997}, \href
  {https://doi.org/10.48550/arXiv.2312.10997}
  {\path{doi:10.48550/arXiv.2312.10997}}.

\bibitem{gazzah_new_2008}
Sami Gazzah and Najoua Essoukri~Ben Amara.
\newblock New {Oversampling} {Approaches} {Based} on {Polynomial} {Fitting} for
  {Imbalanced} {Data} {Sets}.
\newblock In {\em 2008 {The} {Eighth} {IAPR} {International} {Workshop} on
  {Document} {Analysis} {Systems}}, pages 677--684, September 2008.
\newblock URL: \url{https://ieeexplore.ieee.org/abstract/document/4670021},
  \href {https://doi.org/10.1109/DAS.2008.74} {\path{doi:10.1109/DAS.2008.74}}.

\bibitem{geng2023grammarconstrained}
Saibo Geng, Martin Josifoski, Maxime Peyrard, and Robert West.
\newblock Grammar-constrained decoding for structured {NLP} tasks without
  finetuning.
\newblock In {\em The 2023 Conference on Empirical Methods in Natural Language
  Processing}, 2023.
\newblock URL: \url{https://openreview.net/forum?id=KkHY1WGDII}.

\bibitem{ghorbani_distributional_2020}
Amirata Ghorbani, Michael Kim, and James Zou.
\newblock A {Distributional} {Framework} {For} {Data} {Valuation}.
\newblock In {\em Proceedings of the 37th {International} {Conference} on
  {Machine} {Learning}}, pages 3535--3544. PMLR, November 2020.
\newblock URL: \url{https://proceedings.mlr.press/v119/ghorbani20a.html}.

\bibitem{ghorbani_data_2019}
Amirata Ghorbani and James Zou.
\newblock Data {Shapley}: {Equitable} {Valuation} of {Data} for {Machine}
  {Learning}, June 2019.
\newblock arXiv:1904.02868 [cs, stat].
\newblock URL: \url{http://arxiv.org/abs/1904.02868}.

\bibitem{goldberg2016primer}
Yoav Goldberg.
\newblock A primer on neural network models for natural language processing.
\newblock {\em Journal of Artificial Intelligence Research}, 57:345--420, 2016.

\bibitem{goldberg2022neural}
Yoav Goldberg.
\newblock {\em Neural network methods for natural language processing}.
\newblock Springer Nature, 2022.

\bibitem{Gonen_Goldberg_2019}
Hila Gonen and Yoav Goldberg.
\newblock Lipstick on a pig: Debiasing methods cover up systematic gender
  biases in word embeddings but do not remove them.
\newblock page 60–63, August 2019.
\newblock URL: \url{https://aclanthology.org/W19-3621}.

\bibitem{Guha_Khan_Stoyanovich_Schelter_2024}
Shubha Guha, Falaah~Arif Khan, Julia Stoyanovich, and Sebastian Schelter.
\newblock Automated data cleaning can hurt fairness in machine learning-based
  decision making.
\newblock {\em IEEE Transactions on Knowledge and Data Engineering}, page
  1–12, 2024.
\newblock URL: \url{https://ieeexplore.ieee.org/abstract/document/10433778},
  \href {https://doi.org/10.1109/TKDE.2024.3365524}
  {\path{doi:10.1109/TKDE.2024.3365524}}.

\bibitem{Gururangan_Marasovic_Swayamdipta_Lo_Beltagy_Downey_Smith_2020}
Suchin Gururangan, Ana Marasović, Swabha Swayamdipta, Kyle Lo, Iz~Beltagy,
  Doug Downey, and Noah~A. Smith.
\newblock Don’t stop pretraining: Adapt language models to domains and tasks.
\newblock In {\em Proceedings of the 58th Annual Meeting of the Association for
  Computational Linguistics}, page 8342–8360, Online, 2020. Association for
  Computational Linguistics.
\newblock URL: \url{https://www.aclweb.org/anthology/2020.acl-main.740}, \href
  {https://doi.org/10.18653/v1/2020.acl-main.740}
  {\path{doi:10.18653/v1/2020.acl-main.740}}.

\bibitem{Han_Zhang_Wu_Yuan_2023}
Xiao Han, Lu~Zhang, Yongkai Wu, and Shuhan Yuan.
\newblock {\em Achieving Counterfactual Fairness for Anomaly Detection}, volume
  13935, page 55–66.
\newblock Springer Nature Switzerland, Cham, 2023.
\newblock Series Title: Lecture Notes in Computer Science.
\newblock URL: \url{https://link.springer.com/10.1007/978-3-031-33374-3_5},
  \href {https://doi.org/10.1007/978-3-031-33374-3_5}
  {\path{doi:10.1007/978-3-031-33374-3_5}}.

\bibitem{harris54}
Zellig Harris.
\newblock Distributional structure.
\newblock {\em Word}, 10(2-3):146--162, 1954.
\newblock URL:
  \url{https://link.springer.com/chapter/10.1007/978-94-009-8467-7_1}, \href
  {https://doi.org/10.1007/978-94-009-8467-7_1}
  {\path{doi:10.1007/978-94-009-8467-7_1}}.

\bibitem{maximalrepsubsampling}
Tony Hauptmann, Sophie Fellenz, Laksan Nathan, Oliver Tüscher, and Stefan
  Kramer.
\newblock Discriminative machine learning for maximal representative
  subsampling.
\newblock {\em Scientific Reports}, 13(1):20925, November 2023.
\newblock \href {https://doi.org/10.1038/s41598-023-48177-3}
  {\path{doi:10.1038/s41598-023-48177-3}}.

\bibitem{hawkins1980identification}
Douglas~M Hawkins.
\newblock {\em Identification of outliers}, volume~11.
\newblock Springer, 1980.

\bibitem{he_adasyn_2008}
Haibo He, Yang Bai, Edwardo~A. Garcia, and Shutao Li.
\newblock {ADASYN}: {Adaptive} synthetic sampling approach for imbalanced
  learning.
\newblock In {\em 2008 {IEEE} {International} {Joint} {Conference} on {Neural}
  {Networks} ({IEEE} {World} {Congress} on {Computational} {Intelligence})},
  pages 1322--1328, June 2008.
\newblock ISSN: 2161-4407.
\newblock URL: \url{https://ieeexplore.ieee.org/abstract/document/4633969},
  \href {https://doi.org/10.1109/IJCNN.2008.4633969}
  {\path{doi:10.1109/IJCNN.2008.4633969}}.

\bibitem{he2016deep}
Kaiming He, Xiangyu Zhang, Shaoqing Ren, and Jian Sun.
\newblock Deep residual learning for image recognition.
\newblock In {\em Proceedings of the IEEE conference on computer vision and
  pattern recognition}, pages 770--778, 2016.

\bibitem{heger2022understanding}
Amy~K Heger, Liz~B Marquis, Mihaela Vorvoreanu, Hanna Wallach, and Jennifer
  Wortman~Vaughan.
\newblock Understanding machine learning practitioners' data documentation
  perceptions, needs, challenges, and desiderata.
\newblock {\em Proceedings of the ACM on Human-Computer Interaction},
  6(CSCW2):1--29, 2022.

\bibitem{heilmeier1992some}
George~H Heilmeier.
\newblock Some reflections on innovation and invention.
\newblock {\em The Bridge}, 1992.

\bibitem{Hossain2024}
Eklas Hossain.
\newblock {\em Machine Learning Algorithms}, pages 117--259.
\newblock Springer International Publishing, Cham, 2024.
\newblock \href {https://doi.org/10.1007/978-3-031-46990-9_3}
  {\path{doi:10.1007/978-3-031-46990-9_3}}.

\bibitem{hou2017hacking}
Youyang Hou and Dakuo Wang.
\newblock Hacking with npos: collaborative analytics and broker roles in civic
  data hackathons.
\newblock {\em Proceedings of the ACM on Human-Computer Interaction},
  1(CSCW):1--16, 2017.

\bibitem{houlsby2019parameter}
Neil Houlsby, Andrei Giurgiu, Stanislaw Jastrzebski, Bruna Morrone, Quentin
  De~Laroussilhe, Andrea Gesmundo, Mona Attariyan, and Sylvain Gelly.
\newblock Parameter-efficient transfer learning for {NLP}.
\newblock In {\em Proceedings of the 36th International Conference on Machine
  Learning}, 2019.

\bibitem{Hutiri_Patel_Ding_Scharenborg_2024}
Wiebke Hutiri, Tanvina Patel, Aaron~Yi Ding, and Odette Scharenborg.
\newblock As biased as you measure: Methodological pitfalls of bias evaluations
  in speaker verification research.
\newblock In {\em Interspeech 2024}, page 4268–4272. ISCA, September 2024.
\newblock URL:
  \url{https://www.isca-archive.org/interspeech_2024/hutiri24_interspeech.html},
  \href {https://doi.org/10.21437/Interspeech.2024-1158}
  {\path{doi:10.21437/Interspeech.2024-1158}}.

\bibitem{iso_8601}
{International Organization for Standards}.
\newblock Date and time format—iso 8601.
\newblock URL: \url{www.iso.org/iso-8601-date-and-time-format.html}.

\bibitem{jiang_identifying_2019}
Heinrich Jiang and Ofir Nachum.
\newblock Identifying and {Correcting} {Label} {Bias} in {Machine} {Learning},
  January 2019.
\newblock arXiv:1901.04966 [cs, stat].
\newblock URL: \url{http://arxiv.org/abs/1901.04966}.

\bibitem{johnson_survey_2019}
Justin~M. Johnson and Taghi~M. Khoshgoftaar.
\newblock Survey on deep learning with class imbalance.
\newblock {\em Journal of Big Data}, 6(1):27, March 2019.
\newblock \href {https://doi.org/10.1186/s40537-019-0192-5}
  {\path{doi:10.1186/s40537-019-0192-5}}.

\bibitem{Joseph_Vakayil_2022}
V.~Roshan Joseph and Akhil Vakayil.
\newblock Split: An optimal method for data splitting.
\newblock {\em Technometrics}, 64(2):166–176, April 2022.
\newblock arXiv:2012.10945 [cs, stat].
\newblock URL: \url{http://arxiv.org/abs/2012.10945}, \href
  {https://doi.org/10.1080/00401706.2021.1921037}
  {\path{doi:10.1080/00401706.2021.1921037}}.

\bibitem{Jung_Chun_Moon_2022}
Sangwon Jung, Sanghyuk Chun, and Taesup Moon.
\newblock Learning fair classifiers with partially annotated group labels.
\newblock In {\em Proceedings of the IEEE/CVF Conference on Computer Vision and
  Pattern Recognition}, pages 10348--10357, 2022.

\bibitem{Kaneko_Bollegala_2021}
Masahiro Kaneko and Danushka Bollegala.
\newblock Debiasing pre-trained contextualised embeddings.
\newblock In {\em Proceedings of the 16th Conference of the European Chapter of
  the Association for Computational Linguistics: Main Volume}, page
  1256–1266, Online, 2021. Association for Computational Linguistics.
\newblock URL: \url{https://aclanthology.org/2021.eacl-main.107}, \href
  {https://doi.org/10.18653/v1/2021.eacl-main.107}
  {\path{doi:10.18653/v1/2021.eacl-main.107}}.

\bibitem{Kang_2019_CVPR}
Guoliang Kang, Lu~Jiang, Yi~Yang, and Alexander~G. Hauptmann.
\newblock Contrastive adaptation network for unsupervised domain adaptation.
\newblock In {\em Proceedings of the IEEE/CVF Conference on Computer Vision and
  Pattern Recognition (CVPR)}, June 2019.

\bibitem{Kang_Roy_2024}
Wonjune Kang and Deb Roy.
\newblock Prompting large language models with audio for general-purpose speech
  summarization.
\newblock In {\em Interspeech 2024}, page 1955–1959. ISCA, September 2024.
\newblock URL:
  \url{https://www.isca-archive.org/interspeech_2024/kang24d_interspeech.html},
  \href {https://doi.org/10.21437/Interspeech.2024-2213}
  {\path{doi:10.21437/Interspeech.2024-2213}}.

\bibitem{Kar_Castellucci_Filice_Malmasi_Rokhlenko_2022}
Sudipta Kar, Giuseppe Castellucci, Simone Filice, Shervin Malmasi, and Oleg
  Rokhlenko.
\newblock Preventing catastrophic forgetting in continual learning of new
  natural language tasks.
\newblock In {\em Proceedings of the 28th ACM SIGKDD Conference on Knowledge
  Discovery and Data Mining}, page 3137–3145, Washington DC USA, August 2022.
  ACM.
\newblock URL: \url{https://dl.acm.org/doi/10.1145/3534678.3539169}, \href
  {https://doi.org/10.1145/3534678.3539169}
  {\path{doi:10.1145/3534678.3539169}}.

\bibitem{kelton}
Kari Kelton, Kenneth~R. Fleischmann, and William~A. Wallace.
\newblock Trust in digital information.
\newblock {\em Journal of the American Society for Information Science and
  Technology}, 59(3):363--374, 2008.
\newblock URL: \url{https://onlinelibrary.wiley.com/doi/abs/10.1002/asi.20722},
  \href
  {https://arxiv.org/abs/https://onlinelibrary.wiley.com/doi/pdf/10.1002/asi.20722}
  {\path{arXiv:https://onlinelibrary.wiley.com/doi/pdf/10.1002/asi.20722}},
  \href {https://doi.org/10.1002/asi.20722} {\path{doi:10.1002/asi.20722}}.

\bibitem{kerrigan_survey_2021}
Daniel Kerrigan, Jessica Hullman, and Enrico Bertini.
\newblock A {Survey} of {Domain} {Knowledge} {Elicitation} in {Applied}
  {Machine} {Learning}.
\newblock {\em Multimodal Technologies and Interaction}, 5(12):73, November
  2021.
\newblock URL: \url{https://www.mdpi.com/2414-4088/5/12/73}, \href
  {https://doi.org/10.3390/mti5120073} {\path{doi:10.3390/mti5120073}}.

\bibitem{Kim_Diaz_2024}
To~Eun Kim and Fernando Diaz.
\newblock Towards fair rag: On the impact of fair ranking in
  retrieval-augmented generation.
\newblock (arXiv:2409.11598), September 2024.
\newblock arXiv:2409.11598 [cs].
\newblock URL: \url{http://arxiv.org/abs/2409.11598}.

\bibitem{kimura2024short}
Masanari Kimura and Hideitsu Hino.
\newblock A short survey on importance weighting for machine learning, 2024.
\newblock \href {https://arxiv.org/abs/2403.10175} {\path{arXiv:2403.10175}}.

\bibitem{koh_wilds_2021}
Pang~Wei Koh, Shiori Sagawa, Henrik Marklund, Sang~Michael Xie, Marvin Zhang,
  Akshay Balsubramani, Weihua Hu, Michihiro Yasunaga, Richard~Lanas Phillips,
  Irena Gao, Tony Lee, Etienne David, Ian Stavness, Wei Guo, Berton~A.
  Earnshaw, Imran~S. Haque, Sara Beery, Jure Leskovec, Anshul Kundaje, Emma
  Pierson, Sergey Levine, Chelsea Finn, and Percy Liang.
\newblock {WILDS}: {A} {Benchmark} of in-the-{Wild} {Distribution} {Shifts},
  july 2021.
\newblock arXiv:2012.07421 [cs].
\newblock URL: \url{http://arxiv.org/abs/2012.07421}, \href
  {https://doi.org/10.48550/arXiv.2012.07421}
  {\path{doi:10.48550/arXiv.2012.07421}}.

\bibitem{korini_column_2023}
Keti Korini and Christian Bizer.
\newblock Column {Type} {Annotation} using {ChatGPT}, July 2023.
\newblock arXiv:2306.00745 [cs].
\newblock URL: \url{http://arxiv.org/abs/2306.00745}, \href
  {https://doi.org/10.48550/arXiv.2306.00745}
  {\path{doi:10.48550/arXiv.2306.00745}}.

\bibitem{kornowicz2023aggregating}
Jaroslaw Kornowicz and Kirsten Thommes.
\newblock Aggregating human domain knowledge for feature ranking.
\newblock In {\em International Conference on Human-Computer Interaction},
  pages 98--114. Springer, 2023.

\bibitem{krasanakis_adaptive_2018}
Emmanouil Krasanakis, Eleftherios Spyromitros-Xioufis, Symeon Papadopoulos, and
  Yiannis Kompatsiaris.
\newblock Adaptive {Sensitive} {Reweighting} to {Mitigate} {Bias} in
  {Fairness}-aware {Classification}.
\newblock In {\em Proceedings of the 2018 {World} {Wide} {Web} {Conference}},
  {WWW} '18, pages 853--862, Republic and Canton of Geneva, CHE, April 2018.
  International World Wide Web Conferences Steering Committee.
\newblock URL: \url{https://dl.acm.org/doi/10.1145/3178876.3186133}, \href
  {https://doi.org/10.1145/3178876.3186133}
  {\path{doi:10.1145/3178876.3186133}}.

\bibitem{Kubat_Matwin_1997}
M.~Kubát and S.~Matwin.
\newblock Addressing the curse of imbalanced training sets: One-sided
  selection.
\newblock 1997.
\newblock URL:
  \url{https://sci2s.ugr.es/keel/pdf/algorithm/congreso/kubat97addressing.pdf}.

\bibitem{Lahoti_Gummadi_Weikum_2019}
Preethi Lahoti, Krishna~P. Gummadi, and Gerhard Weikum.
\newblock ifair: Learning individually fair data representations for
  algorithmic decision making.
\newblock In {\em 2019 IEEE 35th International Conference on Data Engineering
  (ICDE)}, page 1334–1345, Macao, Macao, 2019. IEEE.
\newblock URL: \url{https://ieeexplore.ieee.org/document/8731591/}, \href
  {https://doi.org/10.1109/ICDE.2019.00121}
  {\path{doi:10.1109/ICDE.2019.00121}}.

\bibitem{Lester_Al-Rfou_Constant_2021}
Brian Lester, Rami Al-Rfou, and Noah Constant.
\newblock The power of scale for parameter-efficient prompt tuning.
\newblock In {\em Proceedings of the 2021 Conference on Empirical Methods in
  Natural Language Processing}, page 3045–3059, Online and Punta Cana,
  Dominican Republic, 2021. Association for Computational Linguistics.
\newblock URL: \url{https://aclanthology.org/2021.emnlp-main.243}, \href
  {https://doi.org/10.18653/v1/2021.emnlp-main.243}
  {\path{doi:10.18653/v1/2021.emnlp-main.243}}.

\bibitem{li2016biocreative}
Jiao Li, Yueping Sun, Robin~J Johnson, Daniela Sciaky, Chih-Hsuan Wei, Robert
  Leaman, Allan~Peter Davis, Carolyn~J Mattingly, Thomas~C Wiegers, and Zhiyong
  Lu.
\newblock Biocreative v cdr task corpus: a resource for chemical disease
  relation extraction.
\newblock {\em Database}, 2016, 2016.

\bibitem{li_table-gpt_2024}
Peng Li, Yeye He, Dror Yashar, Weiwei Cui, Song Ge, Haidong Zhang, Danielle
  Rifinski~Fainman, Dongmei Zhang, and Surajit Chaudhuri.
\newblock Table-{GPT}: {Table} {Fine}-tuned {GPT} for {Diverse} {Table}
  {Tasks}.
\newblock {\em Proceedings of the ACM on Management of Data},
  2(3):176:1--176:28, May 2024.
\newblock URL: \url{https://dl.acm.org/doi/10.1145/3654979}, \href
  {https://doi.org/10.1145/3654979} {\path{doi:10.1145/3654979}}.

\bibitem{limaye2016ecg}
Hrishikesh Limaye and VV~Deshmukh.
\newblock Ecg noise sources and various noise removal techniques: A survey.
\newblock {\em International Journal of Application or Innovation in
  Engineering \& Management}, 5(2):86--92, 2016.

\bibitem{Liu_Hsieh_2020}
Chien-Liang Liu and Po-Yen Hsieh.
\newblock Model-based synthetic sampling for imbalanced data.
\newblock {\em IEEE Transactions on Knowledge and Data Engineering},
  32(8):1543–1556, August 2020.
\newblock URL:
  \url{https://ieeexplore.ieee.org/document/8668459/;jsessionid=A4DAB03AE1C5BF2DE06E6F25EF2BC7F0},
  \href {https://doi.org/10.1109/TKDE.2019.2905559}
  {\path{doi:10.1109/TKDE.2019.2905559}}.

\bibitem{liu2023is}
Jiawei Liu, Chunqiu~Steven Xia, Yuyao Wang, and LINGMING ZHANG.
\newblock Is your code generated by chat{GPT} really correct? rigorous
  evaluation of large language models for code generation.
\newblock In {\em Thirty-seventh Conference on Neural Information Processing
  Systems}, 2023.
\newblock URL: \url{https://openreview.net/forum?id=1qvx610Cu7}.

\bibitem{Liu_Just_Chang_Chen_Jia_2023}
Zhihong Liu, Hoang~Anh Just, Xiangyu Chang, Xi~Chen, and Ruoxi Jia.
\newblock 2d-shapley: A framework for fragmented data valuation.
\newblock (arXiv:2306.10473), July 2023.
\newblock arXiv:2306.10473 [cs].
\newblock URL: \url{http://arxiv.org/abs/2306.10473}.

\bibitem{longpre2024dataauthenticityconsent}
Shayne Longpre, Robert Mahari, Naana Obeng-Marnu, William Brannon, Tobin South,
  Katy Gero, Sandy Pentland, and Jad Kabbara.
\newblock Data authenticity, consent, \& provenance for ai are all broken: what
  will it take to fix them?, 2024.
\newblock URL: \url{https://arxiv.org/abs/2404.12691}, \href
  {https://arxiv.org/abs/2404.12691} {\path{arXiv:2404.12691}}.

\bibitem{Lum_Zhang_Bower_2022}
Kristian Lum, Yunfeng Zhang, and Amanda Bower.
\newblock De-biasing “bias” measurement.
\newblock In {\em 2022 ACM Conference on Fairness, Accountability, and
  Transparency}, page 379–389, Seoul Republic of Korea, June 2022. ACM.
\newblock URL: \url{https://dl.acm.org/doi/10.1145/3531146.3533105}, \href
  {https://doi.org/10.1145/3531146.3533105}
  {\path{doi:10.1145/3531146.3533105}}.

\bibitem{m_teng_correcting_1999}
C~M.~Teng.
\newblock Correcting noisy data.
\newblock In {\em Proceedings of the sixteenth international conference on
  machine learning}, pages 239--248, 1999.
\newblock URL:
  \url{https://citeseerx.ist.psu.edu/document?repid=rep1&type=pdf&doi=ac3f433755c4e90cd97fb12ae49f17c232cae85c}.

\bibitem{Manzini_Lim_Tsvetkov_Black_2019}
Thomas Manzini, Yao~Chong Lim, Yulia Tsvetkov, and Alan~W. Black.
\newblock Black is to criminal as caucasian is to police: Detecting and
  removing multiclass bias in word embeddings.
\newblock (arXiv:1904.04047), July 2019.
\newblock arXiv:1904.04047 [cs, stat].
\newblock URL: \url{http://arxiv.org/abs/1904.04047}.

\bibitem{Mao_2020}
Huanru~Henry Mao.
\newblock A survey on self-supervised pre-training for sequential transfer
  learning in neural networks.
\newblock (arXiv:2007.00800), July 2020.
\newblock arXiv:2007.00800 [cs, stat].
\newblock URL: \url{http://arxiv.org/abs/2007.00800}.

\bibitem{mao2019data}
Yaoli Mao, Dakuo Wang, Michael Muller, Kush~R Varshney, Ioana Baldini, Casey
  Dugan, and Aleksandra Mojsilovi{\'c}.
\newblock How data scientists work together with domain experts in scientific
  collaborations: To find the right answer or to ask the right question?
\newblock {\em Proceedings of the ACM on Human-Computer Interaction},
  3(GROUP):1--23, 2019.

\bibitem{Marvin_Hellen_Jjingo_Nakatumba-Nabende_2024}
Ggaliwango Marvin, Nakayiza Hellen, Daudi Jjingo, and Joyce Nakatumba-Nabende.
\newblock {\em Prompt Engineering in Large Language Models}, page 387–402.
\newblock Springer Nature Singapore, Singapore, 2024.
\newblock URL: \url{https://link.springer.com/10.1007/978-981-99-7962-2_30},
  \href {https://doi.org/10.1007/978-981-99-7962-2_30}
  {\path{doi:10.1007/978-981-99-7962-2_30}}.

\bibitem{May_Wang_Bordia_Bowman_Rudinger_2019}
Chandler May, Alex Wang, Shikha Bordia, Samuel~R. Bowman, and Rachel Rudinger.
\newblock On measuring social biases in sentence encoders.
\newblock In {\em Proceedings of the 2019 Conference of the North}, page
  622–628, Minneapolis, Minnesota, 2019. Association for Computational
  Linguistics.
\newblock URL: \url{http://aclweb.org/anthology/N19-1063}, \href
  {https://doi.org/10.18653/v1/N19-1063} {\path{doi:10.18653/v1/N19-1063}}.

\bibitem{mcinnes2018umap}
Leland McInnes, John Healy, and James Melville.
\newblock Umap: Uniform manifold approximation and projection for dimension
  reduction.
\newblock {\em arXiv preprint arXiv:1802.03426}, 2018.

\bibitem{mehrabi2021survey}
Ninareh Mehrabi, Fred Morstatter, Nripsuta Saxena, Kristina Lerman, and Aram
  Galstyan.
\newblock A survey on bias and fairness in machine learning.
\newblock {\em ACM computing surveys (CSUR)}, 54(6):1--35, 2021.

\bibitem{Meng_McCreadie_Macdonald_Ounis_2020}
Zaiqiao Meng, Richard McCreadie, Craig Macdonald, and Iadh Ounis.
\newblock Exploring data splitting strategies for the evaluation of
  recommendation models.
\newblock In {\em Proceedings of the 14th ACM Conference on Recommender
  Systems}, RecSys ’20, page 681–686, New York, NY, USA, September 2020.
  Association for Computing Machinery.
\newblock \href {https://doi.org/10.1145/3383313.3418479}
  {\path{doi:10.1145/3383313.3418479}}.

\bibitem{Mikolov_Chen_Corrado_Dean_2013}
Tomas Mikolov, Kai Chen, Greg Corrado, and Jeffrey Dean.
\newblock Efficient estimation of word representations in vector space.
\newblock 2013.
\newblock URL: \url{https://arxiv.org/abs/1301.3781}, \href
  {https://doi.org/10.48550/ARXIV.1301.3781}
  {\path{doi:10.48550/ARXIV.1301.3781}}.

\bibitem{Muennighoff_Tazi_Magne_Reimers_2023}
Niklas Muennighoff, Nouamane Tazi, Loic Magne, and Nils Reimers.
\newblock Mteb: Massive text embedding benchmark.
\newblock In Andreas Vlachos and Isabelle Augenstein, editors, {\em Proceedings
  of the 17th Conference of the European Chapter of the Association for
  Computational Linguistics}, page 2014–2037, Dubrovnik, Croatia, May 2023.
  Association for Computational Linguistics.
\newblock URL: \url{https://aclanthology.org/2023.eacl-main.148}, \href
  {https://doi.org/10.18653/v1/2023.eacl-main.148}
  {\path{doi:10.18653/v1/2023.eacl-main.148}}.

\bibitem{muller2019data}
Michael Muller, Ingrid Lange, Dakuo Wang, David Piorkowski, Jason Tsay, Q~Vera
  Liao, Casey Dugan, and Thomas Erickson.
\newblock How data science workers work with data: Discovery, capture,
  curation, design, creation.
\newblock In {\em Proceedings of the 2019 CHI conference on human factors in
  computing systems}, pages 1--15, 2019.

\bibitem{nguyen2022kl}
A.~Tuan Nguyen, Toan Tran, Yarin Gal, Philip Torr, and Atilim~Gunes Baydin.
\newblock {KL} guided domain adaptation.
\newblock In {\em International Conference on Learning Representations}, 2022.
\newblock URL: \url{https://openreview.net/forum?id=0JzqUlIVVDd}.

\bibitem{nicholson_label_2015}
Bryce Nicholson, Jing Zhang, Victor~S. Sheng, and Zhiheng Wang.
\newblock Label noise correction methods.
\newblock In {\em 2015 {IEEE} {International} {Conference} on {Data} {Science}
  and {Advanced} {Analytics} ({DSAA})}, pages 1--9, October 2015.
\newblock URL:
  \url{https://ieeexplore.ieee.org/abstract/document/7344791?casa_token=1sYueJXKBJIAAAAA:80cb-60uhm44tH31jhZeunzrkvM6x5xZMjtlPFYbduDPF0pS_eMLPWjucy-roGpTN8Odlt4Hg9U},
  \href {https://doi.org/10.1109/DSAA.2015.7344791}
  {\path{doi:10.1109/DSAA.2015.7344791}}.

\bibitem{northcutt2021confidentlearning}
Curtis~G. Northcutt, Lu~Jiang, and Isaac~L. Chuang.
\newblock Confident learning: Estimating uncertainty in dataset labels.
\newblock {\em Journal of Artificial Intelligence Research (JAIR)},
  70:1373--1411, 2021.

\bibitem{dodaiethicalprinciples}
United States~Department of~Defense.
\newblock Dod adopts ethical principles for artificial intelligence [press
  release], 2020.
\newblock URL:
  \url{https://www.defense.gov/News/Releases/Release/Article/2091996/dod-adopts-ethical-principles-for-artificial-intelligence/}.

\bibitem{o2006uncertain}
A.~O'Hagan, C.E. Buck, A.~Daneshkhah, J.R. Eiser, P.H. Garthwaite, D.J.
  Jenkinson, J.E. Oakley, and T.~Rakow.
\newblock {\em Uncertain Judgements: Eliciting Experts' Probabilities}.
\newblock Statistics in Practice. Wiley, 2006.
\newblock URL: \url{https://books.google.com/books?id=tl7CDpmlgFwC}.

\bibitem{p_fair_2020}
Deepak P and Savitha~Sam Abraham.
\newblock Fair {Outlier} {Detection}, August 2020.
\newblock arXiv:2005.09900 [cs].
\newblock URL: \url{http://arxiv.org/abs/2005.09900}, \href
  {https://doi.org/10.48550/arXiv.2005.09900}
  {\path{doi:10.48550/arXiv.2005.09900}}.

\bibitem{park2021facilitating}
Soya Park, April~Yi Wang, Ban Kawas, Q~Vera Liao, David Piorkowski, and Marina
  Danilevsky.
\newblock Facilitating knowledge sharing from domain experts to data scientists
  for building nlp models.
\newblock In {\em 26th International Conference on Intelligent User
  Interfaces}, pages 585--596, 2021.

\bibitem{passi2018trust}
Samir Passi and Steven~J Jackson.
\newblock Trust in data science: Collaboration, translation, and accountability
  in corporate data science projects.
\newblock {\em Proceedings of the ACM on Human-Computer Interaction},
  2(CSCW):1--28, 2018.

\bibitem{Peng}
Roger~D. Peng.
\newblock {\em 6.3 Rejection Sampling | Advanced Statistical Computing}.
\newblock URL:
  \url{https://bookdown.org/rdpeng/advstatcomp/rejection-sampling.html}.

\bibitem{Pennington_Socher_Manning_2014}
Jeffrey Pennington, Richard Socher, and Christopher Manning.
\newblock Glove: Global vectors for word representation.
\newblock In Alessandro Moschitti, Bo~Pang, and Walter Daelemans, editors, {\em
  Proceedings of the 2014 Conference on Empirical Methods in Natural Language
  Processing (EMNLP)}, page 1532–1543, Doha, Qatar, October 2014. Association
  for Computational Linguistics.
\newblock URL: \url{https://aclanthology.org/D14-1162}, \href
  {https://doi.org/10.3115/v1/D14-1162} {\path{doi:10.3115/v1/D14-1162}}.

\bibitem{piorkowski2021ai}
David Piorkowski, Soya Park, April~Yi Wang, Dakuo Wang, Michael Muller, and
  Felix Portnoy.
\newblock How ai developers overcome communication challenges in a
  multidisciplinary team: A case study.
\newblock {\em Proceedings of the ACM on Human-Computer Interaction},
  5(CSCW1):1--25, 2021.

\bibitem{pombal_fairness-aware_2023}
José Pombal, Pedro Saleiro, Mário A.~T. Figueiredo, and Pedro Bizarro.
\newblock Fairness-{Aware} {Data} {Valuation} for {Supervised} {Learning},
  March 2023.
\newblock arXiv:2303.16963 [cs].
\newblock URL: \url{http://arxiv.org/abs/2303.16963}.

\bibitem{pukelsheim2014biproportional}
Friedrich Pukelsheim.
\newblock Biproportional scaling of matrices and the iterative proportional
  fitting procedure.
\newblock {\em Annals of Operations Research}, 215:269--283, 2014.

\bibitem{Quaresmini_Primiero_2023}
Camilla Quaresmini and Giuseppe Primiero.
\newblock Data quality dimensions for fair ai.
\newblock (arXiv:2305.06967), May 2023.
\newblock arXiv:2305.06967 [cs].
\newblock URL: \url{http://arxiv.org/abs/2305.06967}, \href
  {https://doi.org/10.48550/arXiv.2305.06967}
  {\path{doi:10.48550/arXiv.2305.06967}}.

\bibitem{Quinonero-CandelaJoaquin2008DSiM}
Joaquin Quiñonero-Candela, Masashi Sugiyama, Anton Schwaighofer, Neil~D
  Lawrence, Amos Storkey, David Corfield, Matthias Hein, Lars~Kai Hansen, Shai
  Ben-David, and Takafumi Kanamori.
\newblock {\em Dataset Shift in Machine Learning}.
\newblock Neural Information Processing series. MIT Press, Cambridge, 1
  edition, 2008.

\bibitem{Rakshit_Singh_Keshari_Chowdhury_Jain_Chadha_2024}
Aishik Rakshit, Smriti Singh, Shuvam Keshari, Arijit~Ghosh Chowdhury, Vinija
  Jain, and Aman Chadha.
\newblock From prejudice to parity: A new approach to debiasing large language
  model word embeddings.
\newblock (arXiv:2402.11512), April 2024.
\newblock arXiv:2402.11512 [cs].
\newblock URL: \url{http://arxiv.org/abs/2402.11512}.

\bibitem{Reitermanova_2010}
Z.~Reitermanová.
\newblock Data splitting.
\newblock 2010.
\newblock URL:
  \url{https://physics.mff.cuni.cz/wds/proc/pdf10/WDS10_105_i1_Reitermanova.pdf}.

\bibitem{ren_learning_2019}
Mengye Ren, Wenyuan Zeng, Bin Yang, and Raquel Urtasun.
\newblock Learning to {Reweight} {Examples} for {Robust} {Deep} {Learning}, May
  2019.
\newblock arXiv:1803.09050 [cs, stat].
\newblock URL: \url{http://arxiv.org/abs/1803.09050}, \href
  {https://doi.org/10.48550/arXiv.1803.09050}
  {\path{doi:10.48550/arXiv.1803.09050}}.

\bibitem{rote2007matrix}
G{\"u}nter Rote and Martin Zachariasen.
\newblock Matrix scaling by network flow.
\newblock In {\em Symposium on Discrete Algorithms: Proceedings of the
  eighteenth annual ACM-SIAM symposium on Discrete algorithms}, volume~7, pages
  848--854, 2007.

\bibitem{roy_multi-fairness_2022}
Arjun Roy, Vasileios Iosifidis, and Eirini Ntoutsi.
\newblock Multi-fairness {Under} {Class}-{Imbalance}.
\newblock In Poncelet Pascal and Dino Ienco, editors, {\em Discovery
  {Science}}, Lecture {Notes} in {Computer} {Science}, pages 286--301, Cham,
  2022. Springer Nature Switzerland.
\newblock \href {https://doi.org/10.1007/978-3-031-18840-4_21}
  {\path{doi:10.1007/978-3-031-18840-4_21}}.

\bibitem{pmlr-v80-ruff18a}
Lukas Ruff, Robert Vandermeulen, Nico Goernitz, Lucas Deecke, Shoaib~Ahmed
  Siddiqui, Alexander Binder, Emmanuel M{\"u}ller, and Marius Kloft.
\newblock Deep one-class classification.
\newblock In Jennifer Dy and Andreas Krause, editors, {\em Proceedings of the
  35th International Conference on Machine Learning}, volume~80 of {\em
  Proceedings of Machine Learning Research}, pages 4393--4402. PMLR, 10--15 Jul
  2018.
\newblock URL: \url{https://proceedings.mlr.press/v80/ruff18a.html}.

\bibitem{russell2016artificial}
Stuart~J Russell and Peter Norvig.
\newblock {\em Artificial Intelligence: A Modern Approach}.
\newblock Pearson, 2010.

\bibitem{Sadeghi_Boddeti_2020}
Bashir Sadeghi and Vishnu~Naresh Boddeti.
\newblock Imparting fairness to pre-trained biased representations.
\newblock In {\em 2020 IEEE/CVF Conference on Computer Vision and Pattern
  Recognition Workshops (CVPRW)}, page 75–82, Seattle, WA, USA, 2020. IEEE.
\newblock URL: \url{https://ieeexplore.ieee.org/document/9150917/}, \href
  {https://doi.org/10.1109/CVPRW50498.2020.00016}
  {\path{doi:10.1109/CVPRW50498.2020.00016}}.

\bibitem{saleiro2019}
Pedro Saleiro, Benedict Kuester, Loren Hinkson, Jesse London, Abby Stevens, Ari
  Anisfeld, Kit~T. Rodolfa, and Rayid Ghani.
\newblock Aequitas: A bias and fairness audit toolkit.
\newblock (arXiv:1811.05577), April 2019.
\newblock arXiv:1811.05577 [cs].
\newblock URL: \url{http://arxiv.org/abs/1811.05577}.

\bibitem{Schroder_Schulz_Kenneweg_Feldhans_Hinder_Hammer_2024}
Sarah Schröder, Alexander Schulz, Philip Kenneweg, Robert Feldhans, Fabian
  Hinder, and Barbara Hammer.
\newblock Evaluating metrics for bias in word embeddings.
\newblock (arXiv:2111.07864), September 2024.
\newblock arXiv:2111.07864.
\newblock URL: \url{http://arxiv.org/abs/2111.07864}, \href
  {https://doi.org/10.48550/arXiv.2111.07864}
  {\path{doi:10.48550/arXiv.2111.07864}}.

\bibitem{DBLP:journals/corr/abs-2406-06608}
Sander Schulhoff, Michael Ilie, Nishant Balepur, Konstantine Kahadze, Amanda
  Liu, Chenglei Si, Yinheng Li, Aayush Gupta, HyoJung Han, Sevien Schulhoff,
  Pranav~Sandeep Dulepet, Saurav Vidyadhara, Dayeon Ki, Sweta Agrawal, Chau
  Pham, Gerson~C. Kroiz, Feileen Li, Hudson Tao, Ashay Srivastava, Hevander~Da
  Costa, Saloni Gupta, Megan~L. Rogers, Inna Goncearenco, Giuseppe Sarli, Igor
  Galynker, Denis Peskoff, Marine Carpuat, Jules White, Shyamal Anadkat,
  Alexander~Miserlis Hoyle, and Philip Resnik.
\newblock The prompt report: A systematic survey of prompting techniques.
\newblock {\em CoRR}, abs/2406.06608, 2024.
\newblock URL: \url{https://doi.org/10.48550/arXiv.2406.06608}.

\bibitem{seiffert2008resampling}
Chris Seiffert, Taghi~M Khoshgoftaar, Jason Van~Hulse, and Amri Napolitano.
\newblock Resampling or reweighting: A comparison of boosting implementations.
\newblock In {\em 2008 20th IEEE International Conference on Tools with
  Artificial Intelligence}, volume~1, pages 445--451. IEEE, 2008.

\bibitem{Shen_Qu_Zhang_Yu_2018}
Jian Shen, Yanru Qu, Weinan Zhang, and Yong Yu.
\newblock Wasserstein distance guided representation learning for domain
  adaptation.
\newblock {\em Proceedings of the AAAI Conference on Artificial Intelligence},
  32(1), April 2018.
\newblock URL: \url{https://ojs.aaai.org/index.php/AAAI/article/view/11784},
  \href {https://doi.org/10.1609/aaai.v32i1.11784}
  {\path{doi:10.1609/aaai.v32i1.11784}}.

\bibitem{sim_data_2022}
Rachael Hwee~Ling Sim, Xinyi Xu, and Bryan Kian~Hsiang Low.
\newblock Data {Valuation} in {Machine} {Learning}: "{Ingredients}",
  {Strategies}, and {Open} {Challenges}.
\newblock In {\em Proceedings of the {Thirty}-{First} {International} {Joint}
  {Conference} on {Artificial} {Intelligence}}, pages 5607--5614, Vienna,
  Austria, July 2022. International Joint Conferences on Artificial
  Intelligence Organization.
\newblock URL: \url{https://www.ijcai.org/proceedings/2022/782}, \href
  {https://doi.org/10.24963/ijcai.2022/782}
  {\path{doi:10.24963/ijcai.2022/782}}.

\bibitem{Singh_Joachims_2018}
Ashudeep Singh and Thorsten Joachims.
\newblock Fairness of exposure in rankings.
\newblock In {\em Proceedings of the 24th ACM SIGKDD International Conference
  on Knowledge Discovery \& Data Mining}, page 2219–2228, London United
  Kingdom, July 2018. ACM.
\newblock URL: \url{https://dl.acm.org/doi/10.1145/3219819.3220088}, \href
  {https://doi.org/10.1145/3219819.3220088}
  {\path{doi:10.1145/3219819.3220088}}.

\bibitem{coral}
Baochen Sun, Jiashi Feng, and Kate Saenko.
\newblock Return of frustratingly easy domain adaptation.
\newblock In {\em AAAI}, 2016.

\bibitem{coral-lda}
Baochen Sun and Kate Saenko.
\newblock From virtual to reality: Fast adaptation of virtual object detectors
  to real domains.
\newblock In {\em BMVC}, 2014.

\bibitem{Tiwald_Ebert_Soukup_2021}
Paul Tiwald, Alexandra Ebert, and Daniel~T. Soukup.
\newblock Representative \& fair synthetic data.
\newblock (arXiv:2104.03007), April 2021.
\newblock arXiv:2104.03007 [cs, stat].
\newblock URL: \url{http://arxiv.org/abs/2104.03007}, \href
  {https://doi.org/10.48550/arXiv.2104.03007}
  {\path{doi:10.48550/arXiv.2104.03007}}.

\bibitem{valliant2013practical}
Richard Valliant, Jill~A Dever, and Frauke Kreuter.
\newblock {\em Practical tools for designing and weighting survey samples},
  volume~1.
\newblock Springer, 2013.

\bibitem{Verma_Ernst_Just_2021}
Sahil Verma, Michael Ernst, and Rene Just.
\newblock Removing biased data to improve fairness and accuracy.
\newblock (arXiv:2102.03054), February 2021.
\newblock arXiv:2102.03054 [cs].
\newblock URL: \url{http://arxiv.org/abs/2102.03054}, \href
  {https://doi.org/10.48550/arXiv.2102.03054}
  {\path{doi:10.48550/arXiv.2102.03054}}.

\bibitem{Vrbancic_Podgorelec_2020}
Grega Vrbancic and Vili Podgorelec.
\newblock Transfer learning with adaptive fine-tuning.
\newblock {\em IEEE Access}, 8:196197–196211, 2020.
\newblock URL: \url{https://ieeexplore.ieee.org/document/9241777/}, \href
  {https://doi.org/10.1109/ACCESS.2020.3034343}
  {\path{doi:10.1109/ACCESS.2020.3034343}}.

\bibitem{Wang2021IsIW}
Ke~Alexander Wang, Niladri~S. Chatterji, Saminul Haque, and Tatsunori
  Hashimoto.
\newblock Is importance weighting incompatible with interpolating classifiers?
\newblock {\em ArXiv}, abs/2112.12986, 2021.
\newblock URL: \url{https://api.semanticscholar.org/CorpusID:245502568}.

\bibitem{Wang_Chen_Zhu_2021}
Xin Wang, Yudong Chen, and Wenwu Zhu.
\newblock A survey on curriculum learning.
\newblock {\em IEEE Transactions on Pattern Analysis and Machine Intelligence},
  page 1–1, 2021.
\newblock URL: \url{https://ieeexplore.ieee.org/document/9392296/}, \href
  {https://doi.org/10.1109/TPAMI.2021.3069908}
  {\path{doi:10.1109/TPAMI.2021.3069908}}.

\bibitem{Wang_Gan_Yang_Wu_Yan_2019}
Yiru Wang, Weihao Gan, Jie Yang, Wei Wu, and Junjie Yan.
\newblock Dynamic curriculum learning for imbalanced data classification.
\newblock In {\em 2019 IEEE/CVF International Conference on Computer Vision
  (ICCV)}, page 5016–5025, Seoul, Korea (South), 2019. IEEE.
\newblock URL: \url{https://ieeexplore.ieee.org/document/9008373/}, \href
  {https://doi.org/10.1109/ICCV.2019.00512}
  {\path{doi:10.1109/ICCV.2019.00512}}.

\bibitem{Webson_Pavlick_2022}
Albert Webson and Ellie Pavlick.
\newblock Do prompt-based models really understand the meaning of their
  prompts?
\newblock In {\em Proceedings of the 2022 Conference of the North American
  Chapter of the Association for Computational Linguistics: Human Language
  Technologies}, page 2300–2344, Seattle, United States, 2022. Association
  for Computational Linguistics.
\newblock URL: \url{https://aclanthology.org/2022.naacl-main.167}, \href
  {https://doi.org/10.18653/v1/2022.naacl-main.167}
  {\path{doi:10.18653/v1/2022.naacl-main.167}}.

\bibitem{shelf}
Cameron~J. Williams, Kevin~J. Wilson, and Nina Wilson.
\newblock {A Comparison of Prior Elicitation Aggregation Using the Classical
  Method and SHELF}.
\newblock {\em Journal of the Royal Statistical Society Series A: Statistics in
  Society}, 184(3):920--940, 05 2021.
\newblock \href
  {https://arxiv.org/abs/https://academic.oup.com/jrsssa/article-pdf/184/3/920/49411844/jrsssa\_184\_3\_920.pdf}
  {\path{arXiv:https://academic.oup.com/jrsssa/article-pdf/184/3/920/49411844/jrsssa\_184\_3\_920.pdf}},
  \href {https://doi.org/10.1111/rssa.12691} {\path{doi:10.1111/rssa.12691}}.

\bibitem{Wilson_1972}
Dennis~L. Wilson.
\newblock Asymptotic properties of nearest neighbor rules using edited data.
\newblock {\em IEEE Transactions on Systems, Man, and Cybernetics},
  SMC-2(3):408–421, July 1972.
\newblock URL: \url{https://ieeexplore.ieee.org/document/4309137}, \href
  {https://doi.org/10.1109/TSMC.1972.4309137}
  {\path{doi:10.1109/TSMC.1972.4309137}}.

\bibitem{Xia_Kong_Yu_Guo_Rossi_Kim_Li_2024}
Yu~Xia, Fang Kong, Tong Yu, Liya Guo, Ryan~A. Rossi, Sungchul Kim, and Shuai
  Li.
\newblock Which llm to play? convergence-aware online model selection with
  time-increasing bandits.
\newblock In {\em Proceedings of the ACM Web Conference 2024}, page
  4059–4070, Singapore Singapore, May 2024. ACM.
\newblock URL: \url{https://dl.acm.org/doi/10.1145/3589334.3645420}, \href
  {https://doi.org/10.1145/3589334.3645420}
  {\path{doi:10.1145/3589334.3645420}}.

\bibitem{xiong_towards_2024}
Haoyi Xiong, Xuhong Li, Xiaofei Zhang, Jiamin Chen, Xinhao Sun, Yuchen Li, Zeyi
  Sun, and Mengnan Du.
\newblock Towards {Explainable} {Artificial} {Intelligence} ({XAI}): {A} {Data}
  {Mining} {Perspective}, January 2024.
\newblock arXiv:2401.04374 [cs].
\newblock URL: \url{http://arxiv.org/abs/2401.04374}.

\bibitem{Xu2021UnderstandingTR}
Da~Xu, Yuting Ye, and Chuanwei Ruan.
\newblock Understanding the role of importance weighting for deep learning.
\newblock {\em ArXiv}, abs/2103.15209, 2021.
\newblock URL: \url{https://api.semanticscholar.org/CorpusID:231807280}.

\bibitem{yan2022forml}
Bobby Yan, Skyler Seto, and Nicholas Apostoloff.
\newblock Forml: Learning to reweight data for fairness.
\newblock {\em arXiv preprint arXiv:2202.01719}, 2022.

\bibitem{yan_synthesizing_2018}
Cong Yan and Yeye He.
\newblock Synthesizing {Type}-{Detection} {Logic} for {Rich} {Semantic} {Data}
  {Types} using {Open}-source {Code}.
\newblock In {\em Proceedings of the 2018 {International} {Conference} on
  {Management} of {Data}}, {SIGMOD} '18, pages 35--50, New York, NY, USA, May
  2018. Association for Computing Machinery.
\newblock URL: \url{https://dl.acm.org/doi/10.1145/3183713.3196888}, \href
  {https://doi.org/10.1145/3183713.3196888}
  {\path{doi:10.1145/3183713.3196888}}.

\bibitem{Yan_2017_CVPR}
Hongliang Yan, Yukang Ding, Peihua Li, Qilong Wang, Yong Xu, and Wangmeng Zuo.
\newblock Mind the class weight bias: Weighted maximum mean discrepancy for
  unsupervised domain adaptation.
\newblock In {\em Proceedings of the IEEE Conference on Computer Vision and
  Pattern Recognition (CVPR)}, July 2017.

\bibitem{Yang_Song_King_Xu_2023}
Xiangli Yang, Zixing Song, Irwin King, and Zenglin Xu.
\newblock A survey on deep semi-supervised learning.
\newblock {\em IEEE Transactions on Knowledge and Data Engineering},
  35(9):8934–8954, September 2023.
\newblock URL: \url{https://ieeexplore.ieee.org/document/9941371/}, \href
  {https://doi.org/10.1109/TKDE.2022.3220219}
  {\path{doi:10.1109/TKDE.2022.3220219}}.

\bibitem{yarovoy_assessing_2020}
Alex Yarovoy, Yiftach Nagar, Einat Minkov, and Ofer Arazy.
\newblock Assessing the {Contribution} of {Subject}-matter {Experts} to
  {Wikipedia}.
\newblock {\em ACM Transactions on Social Computing}, 3(4):1--36, December
  2020.
\newblock URL: \url{https://dl.acm.org/doi/10.1145/3416853}, \href
  {https://doi.org/10.1145/3416853} {\path{doi:10.1145/3416853}}.

\bibitem{yoon_data_2020}
Jinsung Yoon, Sercan Arik, and Tomas Pfister.
\newblock Data {Valuation} using {Reinforcement} {Learning}.
\newblock In {\em Proceedings of the 37th {International} {Conference} on
  {Machine} {Learning}}, pages 10842--10851. PMLR, November 2020.
\newblock URL: \url{https://proceedings.mlr.press/v119/yoon20a.html}.

\bibitem{Zehlike_Castillo_2020}
Meike Zehlike and Carlos Castillo.
\newblock Reducing disparate exposure in ranking: A learning to rank approach.
\newblock In {\em Proceedings of The Web Conference 2020}, page 2849–2855,
  Taipei Taiwan, April 2020. ACM.
\newblock URL: \url{https://dl.acm.org/doi/10.1145/3366424.3380048}, \href
  {https://doi.org/10.1145/3366424.3380048}
  {\path{doi:10.1145/3366424.3380048}}.

\bibitem{zhang2020data}
Amy~X Zhang, Michael Muller, and Dakuo Wang.
\newblock How do data science workers collaborate? roles, workflows, and tools.
\newblock {\em Proceedings of the ACM on Human-Computer Interaction},
  4(CSCW1):1--23, 2020.

\bibitem{Zhang_Lemoine_Mitchell_2018}
Brian~Hu Zhang, Blake Lemoine, and Margaret Mitchell.
\newblock Mitigating unwanted biases with adversarial learning.
\newblock In {\em Proceedings of the 2018 AAAI/ACM Conference on AI, Ethics,
  and Society}, page 335–340, New Orleans LA USA, December 2018. ACM.
\newblock URL: \url{https://dl.acm.org/doi/10.1145/3278721.3278779}, \href
  {https://doi.org/10.1145/3278721.3278779}
  {\path{doi:10.1145/3278721.3278779}}.

\bibitem{Zhang_Lu_Abdalla_McDermott_Ghassemi_2020}
Haoran Zhang, Amy~X. Lu, Mohamed Abdalla, Matthew McDermott, and Marzyeh
  Ghassemi.
\newblock Hurtful words: quantifying biases in clinical contextual word
  embeddings.
\newblock In {\em Proceedings of the ACM Conference on Health, Inference, and
  Learning}, page 110–120, Toronto Ontario Canada, April 2020. ACM.
\newblock URL: \url{https://dl.acm.org/doi/10.1145/3368555.3384448}, \href
  {https://doi.org/10.1145/3368555.3384448}
  {\path{doi:10.1145/3368555.3384448}}.

\bibitem{zhang_towards_2021}
Hongjing Zhang and Ian Davidson.
\newblock Towards {Fair} {Deep} {Anomaly} {Detection}.
\newblock In {\em Proceedings of the 2021 {ACM} {Conference} on {Fairness},
  {Accountability}, and {Transparency}}, pages 138--148, Virtual Event Canada,
  March 2021. ACM.
\newblock URL: \url{https://dl.acm.org/doi/10.1145/3442188.3445878}, \href
  {https://doi.org/10.1145/3442188.3445878}
  {\path{doi:10.1145/3442188.3445878}}.

\bibitem{zhang2023automatic}
Zhuosheng Zhang, Aston Zhang, Mu~Li, and Alex Smola.
\newblock Automatic chain of thought prompting in large language models.
\newblock In {\em The Eleventh International Conference on Learning
  Representations}, 2023.
\newblock URL: \url{https://openreview.net/forum?id=5NTt8GFjUHkr}.

\bibitem{Zhao_Wang_Yatskar_Ordonez_Chang_2018}
Jieyu Zhao, Tianlu Wang, Mark Yatskar, Vicente Ordonez, and Kai-Wei Chang.
\newblock Gender bias in coreference resolution: Evaluation and debiasing
  methods.
\newblock In {\em Proceedings of the 2018 Conference of the North American
  Chapter of            the Association for Computational Linguistics:
  Human Language            Technologies, Volume 2 (Short Papers)}, page
  15–20, New Orleans, Louisiana, 2018. Association for Computational
  Linguistics.
\newblock URL: \url{http://aclweb.org/anthology/N18-2003}, \href
  {https://doi.org/10.18653/v1/N18-2003} {\path{doi:10.18653/v1/N18-2003}}.

\bibitem{Zhao_Wallace_Feng_Klein_Singh_2021}
Zihao Zhao, Eric Wallace, Shi Feng, Dan Klein, and Sameer Singh.
\newblock Calibrate before use: Improving few-shot performance of language
  models.
\newblock In {\em International Conference on Machine Learning}, page
  12697–12706, 2021.
\newblock URL: \url{http://proceedings.mlr.press/v139/zhao21c/ zhao21c.pdf}.

\bibitem{Zhou_Liu_Li_Jin_Qian_Liu_Li_Dou_Ho_Yu_2024}
Yujia Zhou, Yan Liu, Xiaoxi Li, Jiajie Jin, Hongjin Qian, Zheng Liu, Chaozhuo
  Li, Zhicheng Dou, Tsung-Yi Ho, and Philip~S. Yu.
\newblock Trustworthiness in retrieval-augmented generation systems: A survey.
\newblock (arXiv:2409.10102), September 2024.
\newblock arXiv:2409.10102 [cs].
\newblock URL: \url{http://arxiv.org/abs/2409.10102}.

\end{thebibliography}

\appendix
\chapter{Glossary}
\label{sec:glossary}
\textbf{Adversarial Representation Learning (ARL)} - an approach used for debiasing embeddings across modalities. Adversarial learning is characterized by using an ``adversarial" or ``malicious" model to attempt to ``trick" the model being trained in order to encourage the model to learn more robust representations. 

\textbf{AI-enabled system} - a system with an AI component.

\textbf{AI product lifecycle} - the multi-phase process of creating an AI-enabled system. Data curation occupies a phase in the AI product lifecycle after available data have been identified, gathered, and annotated but before a model that will eventually be part of the AI-enabled system has been fully trained.

\textbf{Application Programming Interface (API)} - interface between pieces of software.

\textbf{Artificial Intelligence (AI)} - ``AI refers to the ability of machines to perform tasks that normally require human intelligence – for example, recognizing patterns, learning from experience, drawing conclusions, making predictions, or taking action – whether digitally or as the smart software behind autonomous physical systems." \cite{ai_strategy_summary_2018}

\textbf{Autoregressive models} - statistical models that use past data to predict future data. In NLP, autoregressive models predict the next word (or word piece) given the preceding text, and then that token is added to the input for the next pass through the model. An example of an autoregressive model is GPT. 

\textbf{Biased embedding} - vector-based representation of an entity that leads to \emph{unequal} performance along the dimension of a given selected characteristic where performance is expected to remain equal.

\textbf{Bias meta metric} - a metric to measure the amount of bias exhibited in the model across selected characteristics and tasks.

\textbf{Bias metric} - a metric that measures the degree to which a model exhibits equal performance along a certain selected characteristic for a predetermined task.

\textbf{Bidirectional Encoder Representations from Transformers (BERT)} - A language model which learns representations for text using a neural network architecture (a transformer) and an MLM training scheme.

\textbf{Catastrophic forgetting} - the phenomenon of a neural network performing significantly worse on tasks that it was previously successful at after being fine-tuned to a different task. 

\textbf{Categoricity} - features which fit perfectly into categories. This can be unrealistic in practice, as when a model uses color to distinguish civilian from military ships, but more colors are seen in the deployment environment than in training (insufficient coverage), or colors in the training set are mistakenly recorded as either “blue” or “yellow” and nothing else (mislabeled cases).

\textbf{Causal Language Modeling (CLM)} - a training procedure in which some the model must produce the next word token in a sequence of tokens. 

\textbf{Chain of Thought (CoT) prompt engineering} - a prompt engineering procedure that encourages an LLM to output intermediate reasoning steps when performing logical reasoning tasks. 

\textbf{Cluster-based correction} - a method to correct mischaracterized labels, by which data points are iteratively clustered based on features, and each data point is assigned the most common label of its clusters.

\textbf{Computer Vision (CV)} - A branch of artificial intelligence primarily focused on automatically processing, analyzing, and drawing inferences from visual data modalities (i.e., images and video). 

\textbf{Confident Learning} - data points with likely mischaracterized labels are identified according to a “confident joint” and dropped from the data set.

\textbf{Confirmation bias} - the phenomenon of models that are training using pseudo-labels to overfit to incorrect pseudo-labels.
 
\textbf{Constrained grammar} - a definition of allowable output tokens that can be passed to some LLMs to limit the possible output of the model. Constrained grammars do not prevent hallucinations but rather force the model to respond in an expected way that can allow a used to more easily perform post processing validation on the output. 

\textbf{Contextual embeddings} - word embeddings that return different representations depending on the context provided. An example of contextual embeddings are BERT embeddings. 

\textbf{Counterfactually Fair Anomaly Detection (CFAD)} - an anomaly detection method that makes the same prediction even if a data point’s selected characteristic changes.

\textbf{Counterfactual fairness} - fairness is dependent on the effect of the selected characteristics on model outcomes.

\textbf{Convenience sampling} - subsetting technique whereby data points are selected according to ease of access, which may introduce bias.

\textbf{Data available for AI-enabled system development (abbr. ``data")} - the aggregate of data that developers have access to while they are developing an AI-enabled system prior to deployment.

\textbf{Data cleaning} - process by which data is reformatted from raw or original presentation, often to correct errors, remove noise or standardize in accordance with semantic types.

\textbf{Data curation for an AI-enabled system (abbr. data curation)} - the act of consuming available data and creating one or more datasets from which model parameters will be derived.

\textbf{Data enrichment} - combining data from multiple sources for purpose of verification.

\textbf{Data interoperability} - alignment between systems (e.g., via API) with respect to expected semantic types.

\textbf{Data leakage} - the same data points landing in multiple splits. When a model is validated on the same data it is trained on, evaluation scores will be inflated and unlikely to reflect real world performance.

\textbf{Data mining} - process by which patterns are found in data.

\textbf{Data point} - a unit in the data that represents one entity or phenomenon.

\textbf{Data provenance} - how a data set originated, potentially including source and how it was generated.

\textbf{Datatype} - a set of distinct values, characterized by properties of those values, and by operations on those values. 

\textbf{Deep Fair SVDD} - an extension of Deep SVDD that decorrelates the learned representation of data points in hyperspace from selected characteristics.

\textbf{Deep Support Vector Data Description (Deep SVDD)} - learned hypersphere around the data, such that data points lying outside the hypersphere may be mischaracterized.

\textbf{Deep Learning (DL)} - a branch of artificial intelligence focused on using neural networks for ML tasks (e.g., classification, regression).

\textbf{Deep Neural Networks (DNN)} - a type of neural network in which there are many layers between the input and output representations. In many modern contexts, the distinction between a deep neural network and other neural network architectures is not meaningful.

\textbf{Demographic parity} - the likelihood of a given label should not vary on selected characteristics.

\textbf{Deployed model} - a trained model that is incorporated into the AI-enabled system that is deployed.

\textbf{Domain knowledge} - relevant information about a project's area of interest.

\textbf{Down-sampling} - the process of creating a new pool of data from existing data, whereby data points in selected group(s) (typically the majority class) in the existing data are reduced in number. As with up-sampling, down-sampling---or under-sampling---is usually performed to balance representation across groups.

\textbf{Entity} - the real-world phenomenon that a data point represents.

\textbf{Embedding} - a numerical representation of an entity such that it can later be passed into a downstream ML model. Embeddings are often (but not always) learned such that similar entities exist close to one another in the vector space. 

\textbf{Equalized odds} - the proportion of true positives and false positives should be independent of selected characteristics, given the true label.

\textbf{Equal opportunity} - the proportion of true positives should be independent of selected characteristics, given the true label. Suppose we are distinguishing between images containing military and civilian ships, and the selected characteristic is the image zoom. Among images that contain civilian ships, the classifier should not perform better when the ship is closer to the viewfinder. 

\textbf{Feature} - an aspect of entities or phenomena (that the data points represent) that is encoded in part of the feature vector.

\textbf{Feature-label mischaracterization} - errant features or labels that are correlated with the true underlying features and labels.

\textbf{Feature Mischaracterization} - errant features that are correlated with the true underlying feature.

\textbf{Feature vector} - the machine-interpretable input to an ML model, which applies its parameters to the values of the coordinates of the vectors, computing a prediction of the label or any other model output.

\textbf{Finetuning} - the process of adapting a pretrained model to a new task.

\textbf{Generative Pre-trained Transformer (GPT)} - A type of large language model that is trained on large text corpora and is intended to generate plausible text as output. GPT is typically used in reference to models produced by OpenAI (e.g., ChatGPT, GPT-4o).

\textbf{Hallucination} - an incorrect response from a language model. Hallucinations are often characterized by salient sounding output that contains factual inaccuracies. 

\textbf{Hard debiasing} - an embedding debiasing method that fully removes bias along the dimension of the selected characteristic. This method runs the risk of losing some of the semantic meaning in the embedding. Compare to soft debiasing.  

\textbf{Human-interpretable feature} - A feature with explicit semantic meaning; knowledge about these features can be elicited from subject matter experts.

\textbf{Importance weighting} - see sample weighting.

\textbf{In-Context Learning (ICL)} - the ability of autoregressive models to use data within the input when generating output to extract information that was not present during training. ICL is a concept often used in prompt engineering, to provide relevant information to a given question in the prompt rather than relying on the training process to encode the answer to the question in the LLM. 

\textbf{Individual fairness metric} - a fairness metric that assumes that similar individuals should be treated similarly by a model. 

\textbf{Information Retrieval (IR)} - The task of identifying (given some search criteria, like a query) and retrieving relevant information from an information system. IR may sometimes be used synonymously with document retrieval, in which relevant documents are found from a given query (i.e., search). 

\textbf{Internationalized Resource Identifier (IRI)} - Internet standard to capture a unique sequence of characters meant to identify an abstract or physical resource. For example, IRIs can serve as unique identifiers to distinguish between specific instances of an object type, references in media to those instances, and the class to which an object belongs. 

\textbf{Label Mischaracterization} - errant labels that are correlated with the true underlying label.

\textbf{Labels on data available for AI-enabled system development (abbr. “labels”)} - data field(s) recording the intended output of the AI-enabled system for the associated input. Labels may not be available for all data points. If no labels are available, then the model is generally limited to performing clustering (i.e. grouping inputs by their similarities).

\textbf{Language Model (LM)} - A probabilistic model of natural language, generally trained to predict the likelihood of a word occurring given context.

\textbf{Large Language Model (LLM)} - A type of language model (statistical representation of how words in a given language co-occur) trained on very large text corpora and typically used to generate new text.

\textbf{Local Outlier Factor} - a method to identify mischaracterized data points based on a data point’s distance from its nearest neighbors.

\textbf{Low-Rank Adaptation (LoRA)} - a PEFT method that involves freezing pretrained model weights and injecting trainable weights into each transformer layer of an LLM. 

\textbf{Machine Learning (ML)} - a collection of algorithms for learning parameters of a computational model from data.

\textbf{Masked Language Modeling (MLM)} - A training procedure in which some portion of the inputs to a language model are hidden, or masked, and the model must reproduce the masked portions. 

\textbf{Maximal representative subsampling (MRS)} - similar to down-sampling, a method that uses PU (positive unlabeled) learning to reduce existing data to a subset that is indistinguishable from a target data set.

\textbf{Meta-learning} - a form of modeling that, concurrent with model-training, iteratively re-fits sample weights on the data points.

\textbf{Miscalibration} - difference along some consequential axis between two data sets, as when training set has higher proportion of a selected class than validation set.

\textbf{Mischaracterized data} - data points sufficiently unlike the bulk of the data such that they appear to be drawn from a different distribution.

\textbf{Model architecture} - the logic by which the parameters are applied to a model input to compute the model's output. Selecting a model architecture is not a part of data curation, but it can happen before, during, or after data curation. Some data curation techniques assume that a model architecture has been selected, and some only apply to certain model architectures.

\textbf{Model impact analysis} - detection of mischaracterized data points by assessing a data point’s influence on model outputs

\textbf{Natural Language Processing (NLP)} - A branch of artificial intelligence primarily focused on automatically processing, analyzing, and drawing inferences from textual data modalities (i.e., language).

\textbf{Optimal representative sample weighting} - a form of sample weighting that finds values for data points such that weighting the data set by these values results in a close match to pre-defined target values, typically determined by those in a target population. See representative sample selection.

\textbf{Out of distribution (OOD)} - In machine learning context, OOD data refers to inputs that were not well represented in the training data. That is, if the training data represents a distribution of features and their relative frequencies and co-occurrences, new data that is very unlikely (or completely unrepresented) in this distribution would be referred to as OOD.

\textbf{Out-of-sample prediction} - the model’s estimate for a data point that the model was not trained on.

\textbf{Over-sampling} - see up-sampling.

\textbf{Parameter Efficient Fine Tuning (PEFT)} - methodology for adapting LLMs with relatively low resource costs. 

\textbf{Polishing Labels} - a method to correct mischaracterized labels, by which data is split into ten sets, a classifier is trained on each set, and labels are flipped to the consensus vote of the classifiers’ predictions.

\textbf{Pretraining} - the process of training a model on a large amount of data in order to learn general features from the dataset.

\textbf{Prompt engineering} - the process of iteratively changing the input to an LLM in order to achieve more consistent or accurate results. 

\textbf{Pseudo-labels} - data labels obtained from a model rather than a human being used for tasks for which there are few or no ground truth labels.

\textbf{Random Mischaracterization} - errant features or labels, where neither the incorrect values nor the fact of their occurrence are correlated with the true underlying values.

\textbf{Random over-sampling} - a type of up-sampling where data points in selected group(s) (typically the minority class) in the existing data are chosen with uniform likelihood and with replacement, and the rest automatically chosen without replacement.

\textbf{Representative sample selection} - similar to down-sampling, a process that reduces existing data to a subset with attributes (e.g., class representation, feature means) that closely match target values, typically determined by those in a target population. This is a special case of optimal representative sample weighting where weights are restricted to 0 or 1/$k$, for $k$ the size of the subset.

\textbf{Reproducing Kernel Hilbert Space (RKHS)} - given a real-valued domain, an implied set of functions that can define linear, non-linear and even higher-dimensional similarities between data points on the domain. 

\textbf{Resource Description Framework (RDF)} - a widely used World Wide Web Consortium (W3C) knowledge representation language that allows data to be represented as a graph.

\textbf{Retrieval-Augmented Generation (RAG)} - A technique for augmenting LLM text-generation outputs by (1) retrieving relevant text for a given query and (2) incorporating retrieved text to the input for a generative LLM. RAG is intended to augment LLM generation with more relevant information, ideally enabling the model to produce more coherent or relevant outputs.

\textbf{Sample weighting} - assigning values to data points---typically between 0 and 1, such that all values over all data points sum to 1---that balance the data in some way (e.g., between minority and majority classes, or to align with target data set), for the purpose of more accurate downstream estimator or modeling.

\textbf{Selected characteristic} - a data field recording a distinction among data points across which similar performance is normatively expected by users. Selected characteristics are used widely in the field of group fairness for ML, which refers to them as protected or sensitive attributes. We adopt the name ``selected characteristic'' to highlight that this characteristic need not be sensitive or associated with legal protections for the methods from the fairness literature to apply. Group fairness metrics and techniques can be used on any selected characteristic. We do not assume that a selected characteristic is a feature, meaning it may not be part of an input to the deployed AI-enabled system. However, group fairness methods often require at least some of the data to be annotated with the values of the selected characteristic.

\textbf{Simple Random Sampling (SRS)} - select subset(s) or split data, selecting data points with uniform probability.

\textbf{Self-training correction} - a method to correct mischaracterized labels, by which a noise-filtering algorithm divides the data into a noisy set and a clean set, a classifier is trained on the clean set and used to score the noisy set, and labels of the data points in the noisy set that are likely to be misclassified are flipped.

\textbf{Semantic types} - descriptors for some entities represented in data, such as date.

\textbf{Soft debiasing} - an embedding debiasing method that only partially removes bias along the dimension of the selected characteristic. This allows some of the characteristic to remain in the representation to maintain the semantic meaning. Compare to hard debiasing.  

\textbf{Splitting} - dividing data into distinct sets (e.g., training and validation).

\textbf{Static embedding} - word embeddings which map each word type to a single vector. Static embeddings do \emph{not} change depending on the context around them. Examples of static embeddings are Word2Vec and GLoVE.

\textbf{Stratified sampling} - select subset(s) or split data uniformly over subgroups (i.e., strata).

\textbf{Subject Matter Experts (SMEs)} - People with extensive knowledge and skills in a certain domain.

\textbf{Synthetic minority over-sampling technique (SMOTE)} - up-sampling technique that fabricates data points by interpolating between nearest neighbors in the minority class.

\textbf{Systematic sampling} - for data with a natural ordering, select subset(s) by choosing a random starting point and choosing every $k$th data point from the ordered dataset.

\textbf{Training dataset} - a set of curated data that is used to run the learning algorithm associated with the model architecture, thereby training a candidate model. We assume the training dataset is chosen for its size and may not be representative of the true distribution.

\textbf{Trial and Error} - iterative simple random sampling, so that incidental imbalance in one run is corrected over multiple runs.

\textbf{True distribution of inputs to the deployed AI-enabled system (abbr. true distribution)} - probability distribution over the set of possible inputs to the AI-enabled system once it is deployed. In general, this distribution is assumed to be over infinitely many possible inputs. The inputs that the system actually encounters in the future, once deployed, are assumed to be a uniformly random sample from the true distribution. The true distribution is contrasted with the distribution of the available data.

\textbf{Trustworthiness of an AI-enabled system} - the extent to which the system has been optimized for performance on the true distribution.

\textbf{Unbiased embedding} - vector-based representation of an entity that leads to \emph{equal} performance along the dimension of a given selected characteristic where performance is expected to remain equal. See biased embedding.

\textbf{Under-sampling} - see down-sampling. 

\textbf{Uniform Resource Identifier (URI)} - Analogous to semantic type, but more specific or localized to one’s data, a URI describes a type of entity in data.

\textbf{Up-sampling} - the process of creating a new pool of data from existing data for the purpose of boosting representation of selected group(s) (typically the minority class). As with down-sampling, up-sampling---or over-sampling---is usually performed to balance representation across groups, and can be conducted via random over-sampling or synthesizing new data points (see synthetic minority over-sampling technique).

\textbf{Validation dataset} - a set of curated data that is used to select from trained candidate models. We assume the validation dataset is chosen to most closely approximate a representative sample from the true distribution.

\textbf{Word Embedding Fairness Evaluation (WEFE)} - an open source Python library used to measure and mitigate bias in static word embeddings. 

\

\chapter{List of open-source data curation tools}
\label{sec:tools_list}
Throughout this report, we refer to many open-source tools that implement data curation approaches. These are explained and contextualized throughout the report. We list the tool names and URLs in Table~\ref{tab:tool_list} for quick access, and we refer to the associated sections of the report to describe what each tool does and when it is appropriate to use. We also exemplified the use of several of the tools in 

Appendices~\ref{sec:cv_tutorial} and \ref{sec:nlp_tutorial}; 

such tools are indicated with an asterisk, and the relevant appendix is referenced. When multiple tools are provided in the same repository or package, we group them together in Table~\ref{tab:tool_list}.  

\begin{table}[ht]
    \begin{tabular}{cc|c|c}
        & \textbf{Tool Name} & \textbf{Report Reference} & \textbf{URL} \\ \hline
        * & scikit-learn & \ref{sec:performance}, \ref{sec:split_normal}, \ref{subsec:practitioners_perspective_sss}, \ref{sec:cv_tutorial} & \url{https://scikit-learn.org/stable/} \\
        & AI Fairness 360 & \ref{sec:performance} & \url{https://github.com/Trusted-AI/AIF360} \\
        & Aequitas & \ref{knowledge/transformation} & \url{https://github.com/dssg/aequitas} \\
        & SciPy & \ref{knowledge/true/parameters} & \url{https://scipy.org/} \\
        & SHELF & \ref{knowledge/true/shelf} & \url{https://shelf.sites.sheffield.ac.uk/} \\
        & WILDS & \ref{subsubsec:split_distro_shift} & \url{https://wilds.stanford.edu/} \\
        & Weight Debiasing & \ref{subsubsec:split_distro_shift} & \url{https://github.com/kramerlab/WeightDebiasing} \\
        * & rsw & \ref{sec:optimal_rep_sample_weighting}, \ref{subsec:practioners_perspective_rsw_selection}, \ref{sec:reweight/raking}, \ref{subsec:practioners_perspective_rsw_weighting}, \ref{sec:nlp_tutorial} & \url{https://github.com/cvxgrp/rsw} \\
        * & imbalanced learn & \ref{sec:resampling}, \ref{sec:resample/practitioners}, \ref{sec:cv_tutorial} & \url{https://imbalanced-learn.org/stable/} \\
        & OpenDataVal & \ref{sec:reweight/initial_perf}, \ref{sec:reweight_dynamic} & \url{https://github.com/opendataval/opendataval} \\
        & Google Looker Studio & \ref{sec:semantic_types/recognize} & \url{https://lookerstudio.google.com/overview} \\
        & Tensorflow Evaluator & \ref{sec:semantic_types/verify} & \url{https://www.tensorflow.org/tfx/guide/evaluator} \\
        & OoD Analyzer & \ref{mischar/detect} & \url{https://github.com/thu-vis/OoDAnalyzer} \\
        * & CleanLab & \ref{mischar/correct/general}, \ref{mischar/practitioners}, \ref{sec:cv_tutorial} & \url{https://cleanlab.ai/} \\
        & CGL Fairness & \ref{mischar/correct/fair} & \url{https://github.com/naver-ai/cgl_fairness} \\
        * & WEFE & \ref{sec:measure_we_bias}, \ref{sec:correct_we_bias}, \ref{subsec:practitioners_perspective_wefe}, \ref{sec:nlp_tutorial} & \url{https://github.com/dccuchile/wefe} \\
        & DebIE & \ref{sec:correct_we_bias} & \url{https://github.com/anlausch/DEBIE}
    \end{tabular}
    \caption{Open-Source Data Curation Tools}
    \label{tab:tool_list}
\end{table}

\chapter{Data curation tutorial for a computer vision use case}
The Defense Advanced Research Projects Agency (DARPA) Triage Challenge Phase 1 manikin video data was collected in collaboration with the Army Telemedicine and Advanced Technology Research Center (TATRC).
\label{sec:cv_tutorial}
\includepdf[pages=-,pagecommand={},templatesize={7in}{9.5in}]{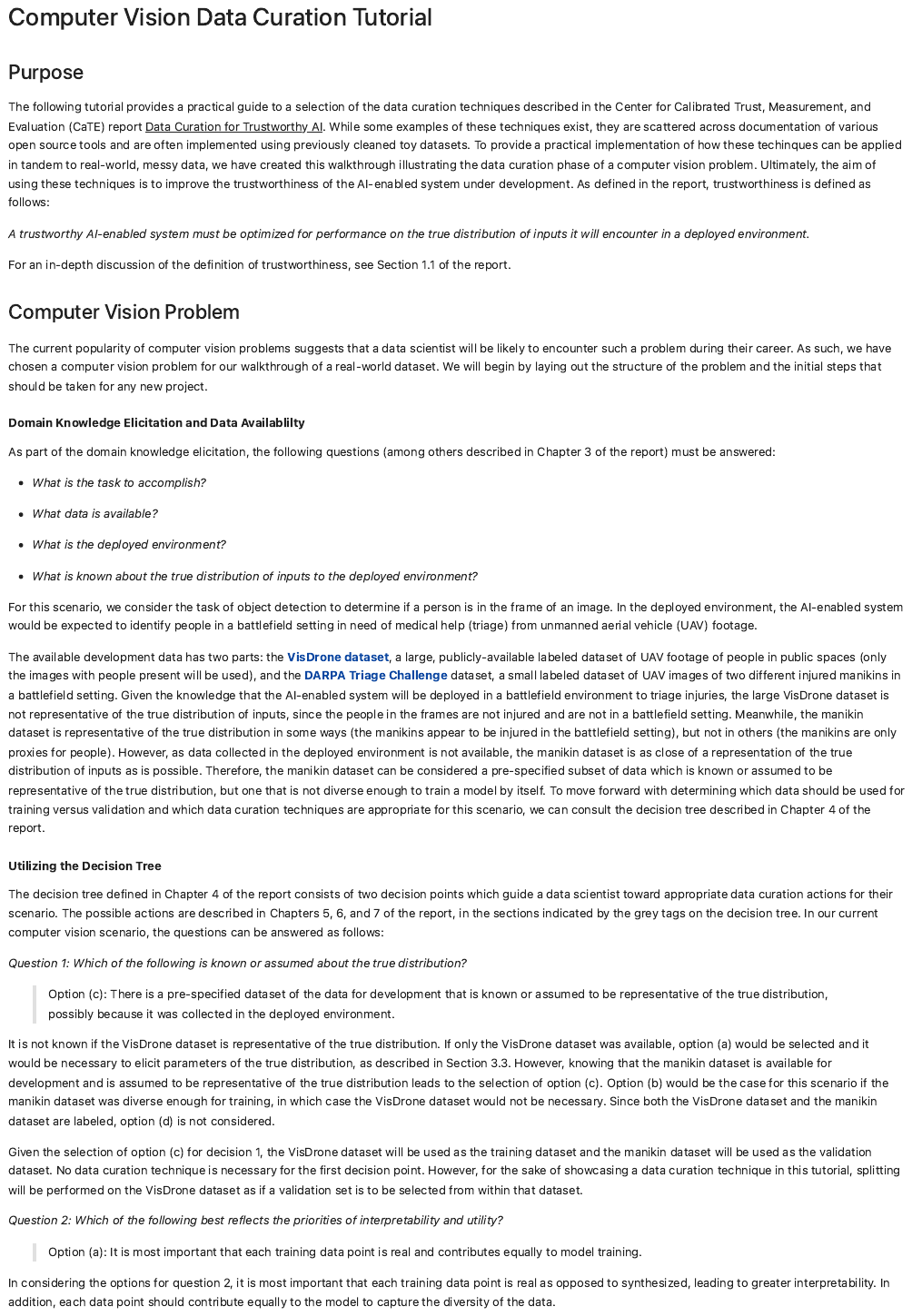}

\chapter{Data curation tutorial for a natural language processing use case}
\label{sec:nlp_tutorial}
\includepdf[pages=-,pagecommand={},templatesize={7in}{9.5in}]{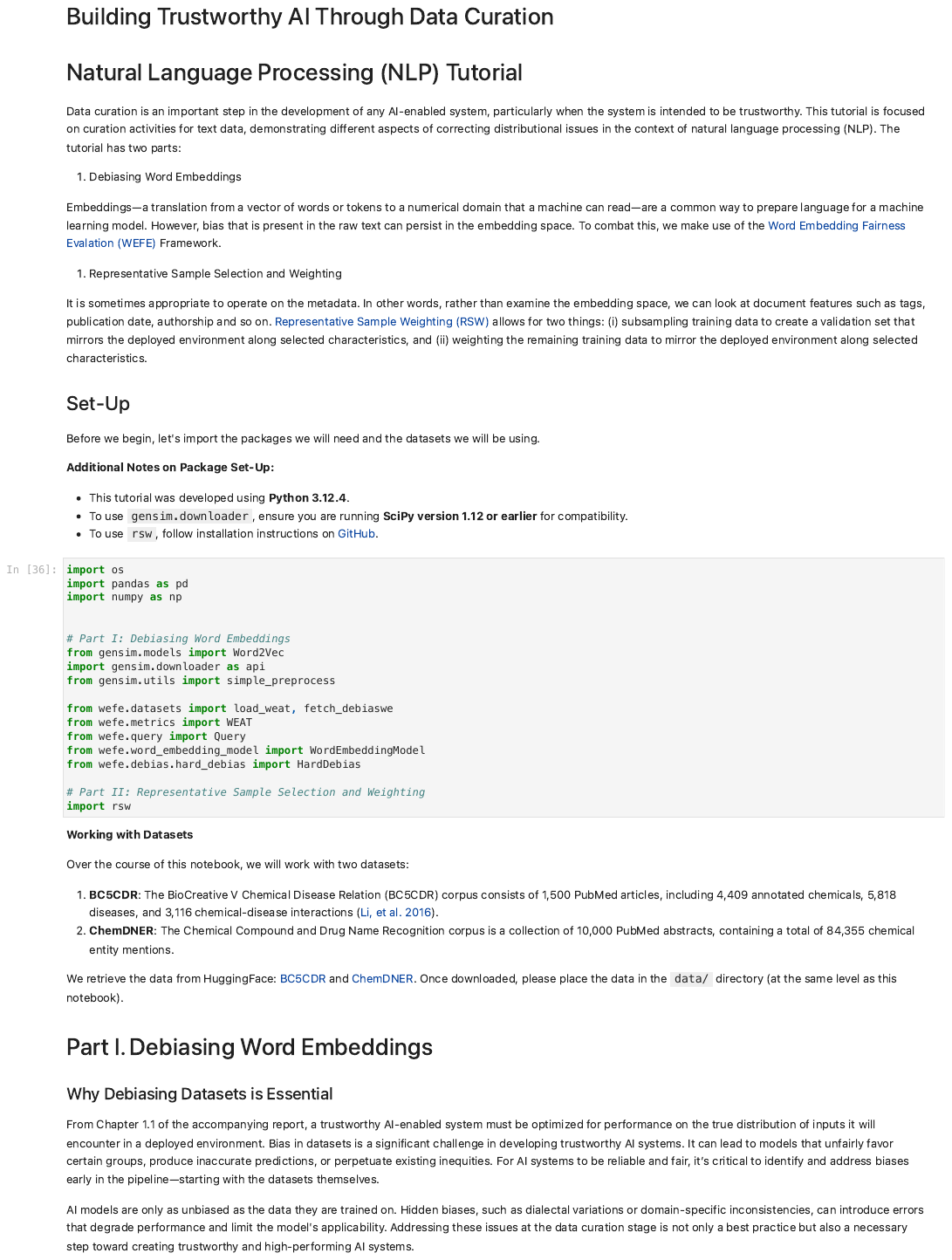}


\end{document}